\documentclass[lettersize,journal]{IEEEtran}
\IEEEoverridecommandlockouts
\usepackage{amsmath,graphicx}
\usepackage[utf8]{inputenc}
\usepackage{orcidlink}
\usepackage{amssymb}
\usepackage{stackengine}

\usepackage[T1]{fontenc}
\usepackage{gensymb}
\usepackage{multirow}

 \usepackage{subcaption}
\usepackage{tikz}
\usepackage{tcolorbox}
\usepackage{breqn}
\definecolor{arrowblue}{rgb}{0,0,0.8}
\usepackage{acronym}
\usepackage{array}
\usepackage{algorithm}
\usepackage{algpseudocode}
\usepackage{url}

\usepackage{cite} 
\usepackage[export]{adjustbox}
\acrodef{GSF}{glare spread function}
\acrodef{HDR}{high dynamic range}
\acrodef{CV}{computer vision}
\acrodef{TF}{transfer function}
\usepackage{hyperref}

\begin{document}

 \title{Depth-Aware Image and Video Orientation Estimation}
\author{
   Muhammad Z. Alam\orcidlink{https://orcid.org/0000-0002-0114-8248}, \thanks{Muhammad Z. Alam, Faculty of Computer science, University of NewBrunswick, (e-mail: muhammad.alam@unb.ca)}
  \and
  Larry Stetsiuk, \thanks{Larry Stetsiuk, Department of Computer Science, Brandon University,  (e-mail: stetsil37@brandonu.ca)}
   \and
    M. Umair Mukati\orcidlink{https://orcid.org/0000-0001-6797-2814},\thanks{M. Umair Mukati, Department of Electrical and Photonics Engineering, Technical University of Denmark, 2800 Kgs. Lyngby, Denmark,  (e-mail: mummu@dtu.dk)}
     \and  
  Zeeshan Kaleem\orcidlink{https://orcid.org/0000-0002-7163-0443}, \thanks{Zeeshan Kaleem, Department of Electrical and Computer Engineering, COMSATS University,  (e-mail: zeeshankaleem@ciitwah.edu.pk)}
}
\maketitle
\begin{abstract}
This paper introduces a novel approach for image and video orientation estimation by leveraging depth distribution in natural images. The proposed method estimates the orientation based on the depth distribution across different quadrants of the image, providing a robust framework for orientation estimation suited for applications such as virtual reality (VR), augmented reality (AR), autonomous navigation, and interactive surveillance systems. To further enhance fine-scale perceptual alignment, we incorporate depth gradient consistency (DGC) and horizontal symmetry analysis (HSA), enabling precise orientation correction. This hybrid strategy effectively exploits depth cues to support spatial coherence and perceptual stability in immersive visual content. Qualitative and quantitative evaluations demonstrate the robustness and accuracy of the proposed approach, outperforming existing techniques across diverse scenarios.

\end{abstract}
\begin{keywords}
Image orientation estimation, immersive visualization, depth cues, depth gradient consistency, horizontal symmetry analysis, automated spatial alignment.
\end{keywords}
\section{Introduction}
\begin{figure*}[!h]
\centering
\includegraphics[trim={0 105 0 110 },clip,width=1\textwidth]{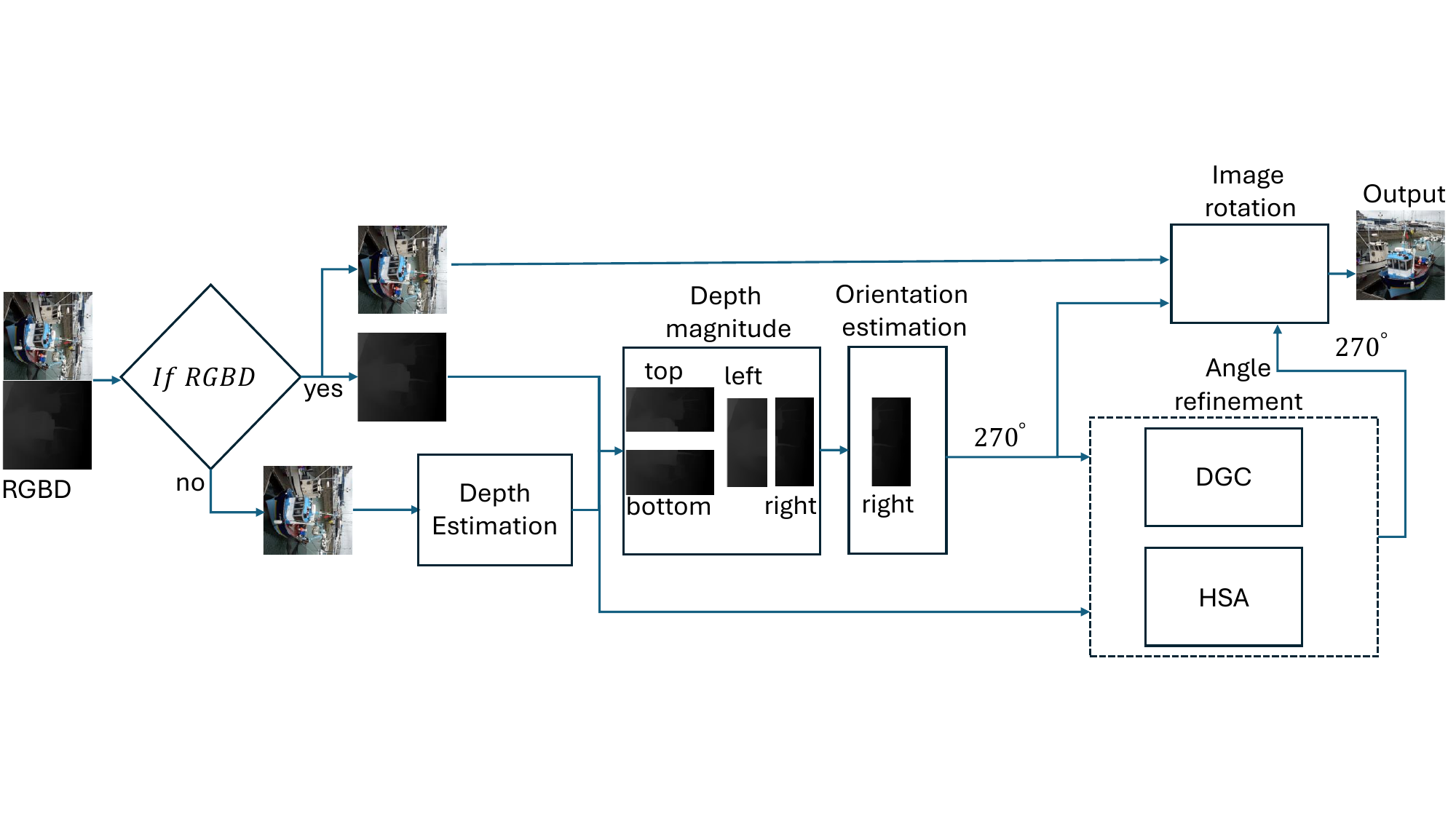}
 \caption{Illustration of the orientation estimation pipeline.}
\label{illustration}
\end{figure*}

Accurate orientation estimation of natural images is critical in immersive environments such as virtual reality (VR), augmented reality (AR), and 360° video systems. Misoriented content in these settings can cause discomfort and diminish the sense of presence, core pillars of immersive user experience. In head-mounted displays (HMDs) and VR-based telepresence, even minor misalignments in image orientation can lead to perceptual dissonance or spatial inconsistency between virtual and real-world cues, ultimately affecting interaction fidelity and user comfort \cite{narumi2016jackin,9133103}.\\ Applications in exploratory visual analytics, wearable capture systems, and drone-based visualization \cite{6634090} increasingly rely on the automatic correction of orientation to ensure visual realism and coherence across viewpoints. In such contexts, traditional orientation correction techniques that rely solely on scene layout heuristics or EXIF metadata often fall short. \\
These methods do not generalize well to unconstrained outdoor scenes or dynamic acquisition settings where metadata is missing or unreliable. Instead, robust, data-driven solutions are needed, ones that can infer orientation from visual cues directly, even in the absence of human annotations or auxiliary sensors. This paper introduces a depth-aware framework designed specifically to infer image orientation in unconstrained environments, using geometric features that align well with the requirements of immersive applications.\\
Given the vast and ever-growing volume of digital imagery generated every day, automated and precise image orientation correction has become imperative. Devices ranging from head-mounted displays to smartphones and aerial cameras generate high-resolution visual data that demands efficient orientation handling. Automated image orientation correction not only enhances the visual coherence and realism in immersive and interactive systems but also reduces the manual effort required in preparing large-scale visual datasets. This advancement is crucial for leveraging large datasets in training robust machine learning models, enabling more sophisticated and accurate analysis, and ultimately driving innovation in the aforementioned technical fields.\\
Currently, state-of-the-art image orientation methods employ deep learning models \cite{joshi2017automatic}, and \cite{maji2020deep} that necessitate extensive datasets for training, and their predictive accuracy is significantly influenced by the heterogeneity of the dataset. Typically, deep learning models extract pertinent features through training involving multiple convolutional layers. These layers are iteratively updated as the algorithm is trained with labeled data, converging on the optimal weights through numerical optimization techniques that best represent the training data \cite{Glare}. Our approach offers a generalizable solution for image orientation prediction. It's perhaps the only non-learning strategy capable of determining fine-scale angles, a feature unique to state-of-the-art deep-learning techniques, that lack generalization.\\
In our proposed method, illustrated in Fig. \ref{illustration}, we introduce a novel approach for estimating image/video orientation by harnessing the depth distribution in natural images. Depth which is crucial for understanding scene geometry and thereby instrumental in estimating image orientation, has not been previously considered in this context. We analyze the depth distribution across four primary quadrants of the image, corresponding to \(0^\circ\), \(90^\circ\), \(180^\circ\), and \(270^\circ\), to determine the broad orientation category (e.g., landscape, portrait, or flipped views). \\The depth magnitude within each quadrant is evaluated to identify the quadrant with the highest depth magnitude, providing an initial coarse orientation estimate. Once the primary quadrant is determined, the search is refined further by restricting the evaluation to a range of \(\pm 45^\circ\) around the identified quadrant, enabling a more precise prediction of the image’s orientation. For further refinement two techniques are utilized: Depth Gradient Consistency (DGC) and Horizontal Symmetry Analysis (HSA). DGC evaluates the consistency of vertical depth gradients within specific regions at various rotation angles. Meanwhile, HSA assesses the symmetry along the horizontal axis of the rotated depth map, by comparing the left and right halves of the depth distribution. By adding these refinement techniques, our method can determine the fine-scale orientation of the image.

Depth sensing technologies such as Microsoft Kinect \cite{han2013enhanced}, light field cameras \cite{alam2018hybrid},\cite{alam2018deconvolution} stereo vision systems \cite{o2018stereo}, and Time-of-Flight (ToF) sensors \cite{7360902} have become foundational to immersive environments and interactive visualization systems. These devices, often equipped with RGB-D (color and depth) capabilities, enable rich spatial understanding crucial for virtual and augmented reality (VR/AR), mixed-reality navigation, and 3D scene reconstruction. \\
In immersive applications, accurate depth information enhances real-time perception of spatial layout, improves user interaction with virtual objects, and supports consistent alignment between physical and virtual spaces. As depth acquisition becomes increasingly ubiquitous, leveraging this information for tasks such as orientation estimation plays a key role in stabilizing and enriching the immersive visual experience.
In scenarios where depth information is not inherently available, modern computational techniques such as, Depth-from-defocus, Depth-from-shading, and learning-based methods\cite{Ranftl2020},\cite{alam2021analysis} and \cite{gasperini2023robust} enable monocular depth estimation effectively, ensuring that depth data can be reliably generated and utilized.\\
Notably, in the proposed method, depth estimation occurs at a coarse level, only for inputs without existing depth maps, focusing not on precise measurements but rather on identifying broad depth patterns to inform our analysis. This approach provides essential contextual information for image orientation, focusing on broader scene understanding rather than detailed depth accuracy which is computationally intensive. 
The major contributions of this paper are summarized as follows:
\begin{enumerate}
  \item A novel depth cue is proposed for accurate image orientation estimation, suitable for immersive visualization environments such as virtual reality (VR), augmented reality (AR), and 360° content. 
  \item We introduce a defocus-based technique for orientation estimation that operates without requiring explicit depth maps, enabling seamless integration into real-time image and video pipelines, including lightweight AR/VR systems where depth data may be unavailable or noisy.
\end{enumerate}
% Our code is available at \href{https://github.com/ZeshanAlam/Image_orientation/tree/main}{\color{pink}Image_orientation}
\section{Related Work}
\color{black}
\label{literature}
Image orientation estimation has evolved significantly over the years. In  \cite{vailaya2002automatic} an automatic image orientation detection algorithm using a Bayesian learning framework is proposed for estimating class-conditional densities required by Bayesian methodology. In \cite{wang2004detecting} an automatic image orientation detection method focusing on low-level visual content, specifically luminance and chrominance  features is proposed. Their approach integrates outputs from multiple classifiers using both static and trainable combiners and incorporates rejection options to enhance accuracy by filtering out low-confidence images.\\
In \cite{ciocca2010image} a combination of a face detector and low-level features analysis concerning color and edge distributions is utilized. This method leverages the Viola and Jones algorithm for face detection. For images without detectable faces, orientation is assessed through an image classifier analyzing low-level features.\\
In more recent years, the advent of deep learning has further transformed the landscape. The work in \cite{joshi2017automatic} exemplifies this shift. This research harnesses the power of Convolutional Neural Networks (CNNs) with modified backpropagation which allows only positive gradients to flow back through the network. This focuses the visualization on the features that most activate a given neuron. In \cite{maji2020deep} a sophisticated learning model that uses a custom loss function is proposed. The loss function addresses the cyclic nature of angles, ensuring accurate predictions even when orientations are near the cyclic boundary. This approach not only detects the orientation but also estimates the orientation angle, offering a more granular understanding of the image's orientation. In \cite {lu2023orientation} a method for pedestrian attribute recognition by modeling relationships between attributes and image orientations using Graph Convolution Networks (GCNs) is proposed to improve the recognition accuracy in various orientations.
The aforementioned methods contribute to the evolving area of image orientation estimation, showcasing a trajectory from traditional machine learning techniques to the latest advancements in deep learning.\\
\textit{Depth Estimation:}
The proposed image orientation estimation technique requires depth estimation from a single image. In iDisc \cite{piccinelli2023idisc} an internal discretization module to improve monocular depth estimation is introduced. By learning high-level patterns through a continuous-discrete-continuous bottleneck it avoids explicit geometric constraints or priors. A trap attention mechanism to improve monocular depth estimation is presented in \cite{ning2023trap} . The trap attention sets traps in the extended space for each pixel, converting quadratic computational complexity to linear form. The method incorporates a vision transformer as the encoder and uses the trap attention in the decoder to estimate depth from a single image. \\
In \cite{su2020monocular} an Information Exchange Network (IEN) for monocular depth estimation is introduced. The network focuses on exchanging and integrating local and global information to improve depth prediction from a single RGB image. By leveraging both contextual and detailed feature information, the method achieves high accuracy in depth estimation. \\
In the non-learing based depth estimation approches, estimating depth from a single defocused image has been studied extensively in \cite{Aslantas},\cite{Lin},\cite{XZHU}. Single image captured with circular aperture usually assumes 
some prior information about either the PSF \cite{Aslantas}, texture \cite{Lin} or color information \cite{XZHU}. DFD techniques that use circular apertures have shown some promising results in the past \cite{alam2022light} but, there are some limitations of circular aperture in depth discrimination.\\
Coded aperture photography has been used for both, improving out of focus deblurring  \cite{VEER,CZHO,Masia} and better depth estimation \cite{levin,Sellent,MasoudifarImageAD,misbah2024msf}. \\
\textit{Inspired by \cite{alam2022light} that utilize a single image from conventional camera without any optical modification, We introduce a non-learning-based method to gather defocus information for image orientation estimation in scenarios where depth data is not intrinsically available.}
\section{Linear perspective and camera angle}
\label{LP}
\begin{figure}[h]
\centering
\includegraphics[width=0.45\textwidth]{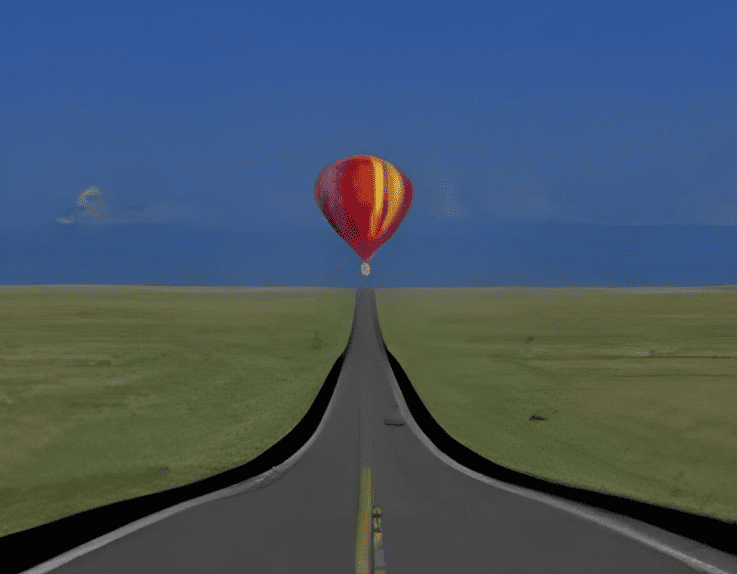}
\caption{Illustration of linear perspective in image acquisition.}
\label{illus}
\end{figure}

Linear perspective, a principle originating from the fields of art and geometry, is applied in the context of image acquisition to explain how objects appear higher in the field of view as they recede in distance, as shown in Fig. \ref{illus}. In the realm of photography and image acquisition, an intriguing phenomenon occurs where objects or scenes appear higher in an image as they move further away from the camera.\\
Parallel lines in the three-dimensional world appear to converge as they recede into the distance. This phenomenon causes distant objects to appear smaller and higher in the field of view. For example, if you look at a straight road stretching into the distance, the sides of the road seem to converge and rise towards the horizon.\\
The horizon line acts as a reference point in an image. Objects that are far away tend to align with this line, while objects closer to the camera appears in the lower portion of the image it. As an object moves away from the camera, its image shifts upwards towards the horizon line in the viewer’s field of vision.\\
According to perspective projection model a point in the world coordinate system \((x, y, z)\), where \(z\) represents the depth (distance from the camera), \(y\) is the vertical position, and \(x\) is the horizontal position, is projected onto the image plane involves the following transformation:
\begin{equation}
% \[
\begin{bmatrix} 
u \\ 
v \\ 
w 
\end{bmatrix} = \begin{bmatrix} 
fx + c_xz \\ 
fy + c_yz \\ 
z 
\end{bmatrix}
% \]
\end{equation}
where \(f\) is the focal length of the camera, and \((c_x, c_y)\) are the coordinates of the principal point in the image plane. 
To get the 2D image coordinates, we normalize the homogeneous coordinates:
\begin{equation}
u' = \frac{u}{w} = \frac{fx + c_xz}{z} = f \frac{x}{z} + c_x
\end{equation}
\begin{equation}
v' = \frac{v}{w} = \frac{fy + c_yz}{z} = f \frac{y}{z} + c_y
\end{equation}

Assuming the horizon line is at half the image height, we set \(c_y = \frac{H}{2}\), where \(H\) is the height of the image. This simplifies to:

\begin{equation}
v' = f \frac{y}{z} + \frac{H}{2}
\end{equation}
Here, \(v'\) is the vertical position of the object's projection on the image plane, \(f\) is the focal length, \(y\) is the vertical coordinate of the object in the world coordinate system, and \(z\) is the depth.
This equation shows that as \(z\) increases, the term \(f \frac{y}{z}\) decreases, causing \(v'\) to approach the horizon line \(\frac{H}{2}\). This means that objects closer to the camera (with smaller \(z\)) will appear lower in the image.
\begin{figure*}[!h]
\centering
\raisebox{.7\height}{ \rotatebox[origin=]{90}{High Level}}
\begin{subfigure}{2.0in}
{\includegraphics[width=2.0in,height=2.0in,keepaspectratio]{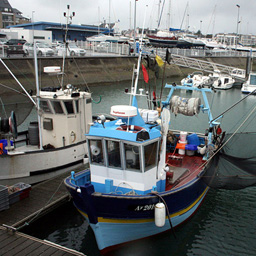}}
\end{subfigure}
\raisebox{.7\height}{ \rotatebox[origin=]{90}{Eye Level}}
\begin{subfigure}{2.0in}
{\includegraphics[width=2.0in,height=2.0in,keepaspectratio]{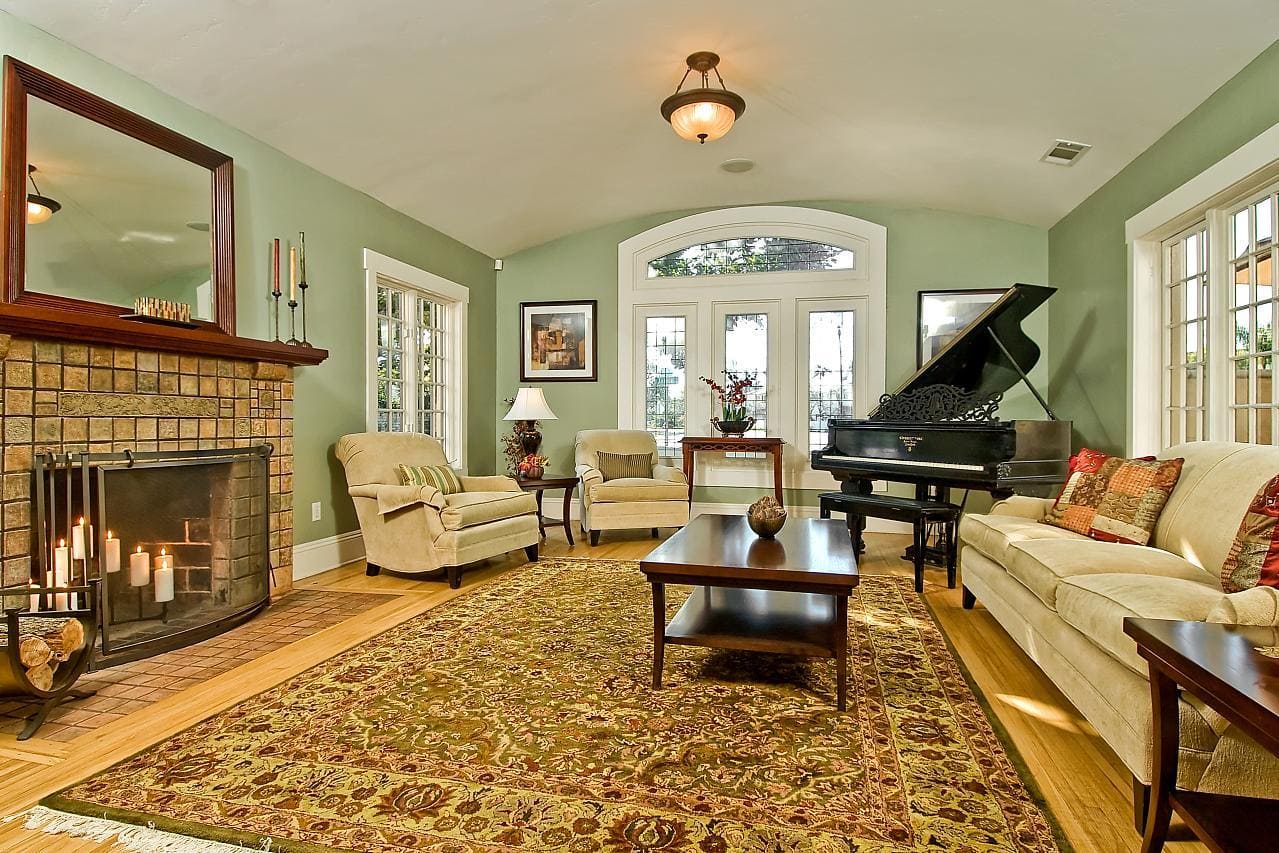}}
\end{subfigure}
\raisebox{.7\height}{ \rotatebox[origin=]{90}{Low Level}}
\begin{subfigure}{2.0in}
{\includegraphics[width=2.0in,height=2.0in,keepaspectratio]{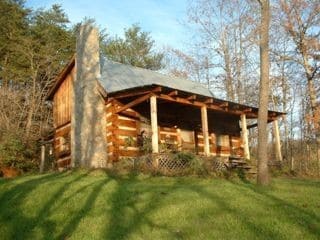}}
\end{subfigure}\\
\raisebox{.7\height}{ \rotatebox[origin=]{90}{High Level}}
\begin{subfigure}{2.0in}
{\includegraphics[width=2.0in,height=2.0in,keepaspectratio]{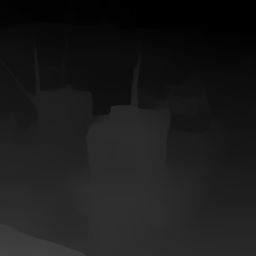}}
\end{subfigure}
\raisebox{.7\height}{ \rotatebox[origin=]{90}{Eye Level}}
\begin{subfigure}{2.0in}
{\includegraphics[width=2.0in,height=2.0in,keepaspectratio]{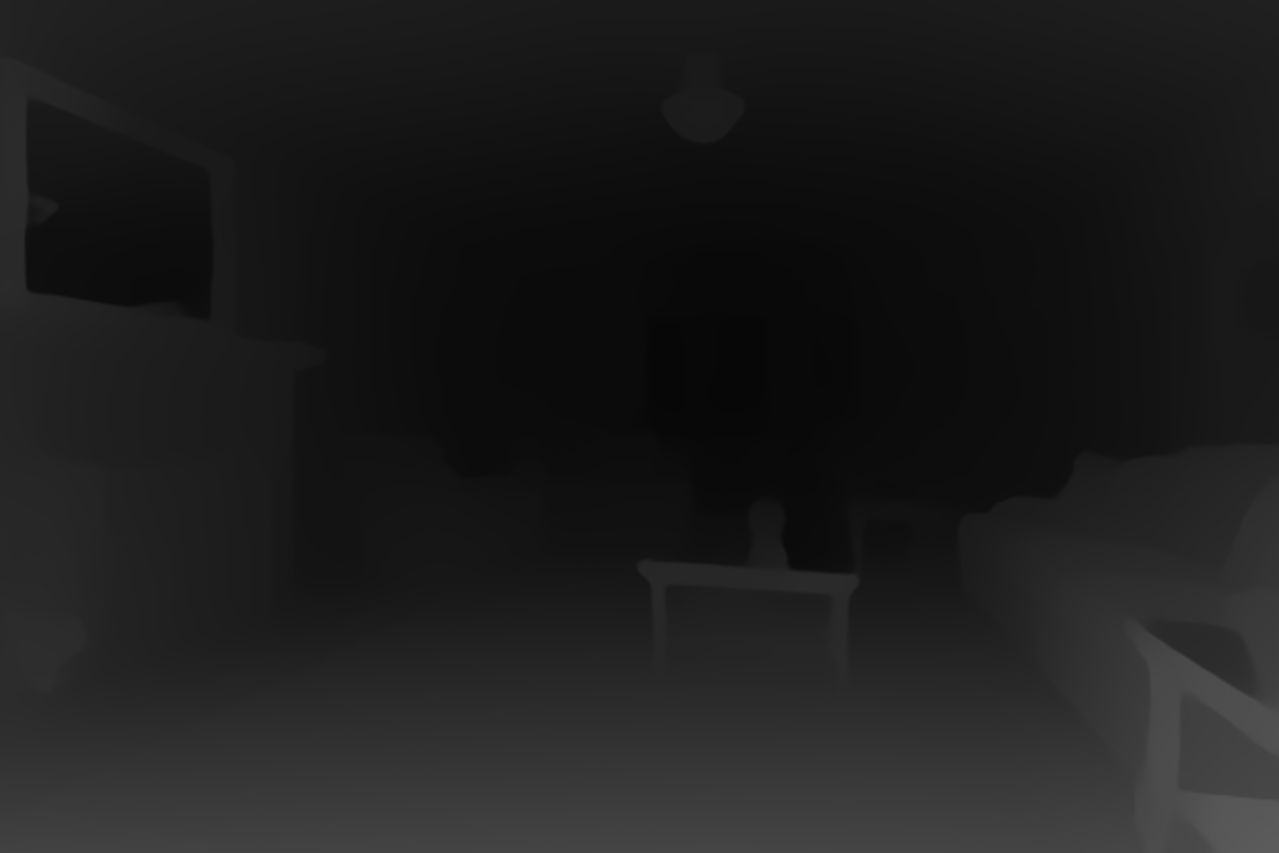}}
\end{subfigure}
\raisebox{.7\height}{ \rotatebox[origin=]{90}{Low Level}}
\begin{subfigure}{2.0in}
{\includegraphics[width=2.0in,height=2.0in,keepaspectratio]{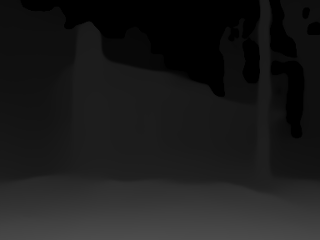}}
\end{subfigure}\\
\caption{Different viewing angles and their effect on depth displacement in corresponding disparity (inverse of depth) maps.}
\label{Angle}
\end{figure*}
This effect is heavily influenced by the camera angle i.e.  eye-level, low-level, and high-level, angles, each offering a unique perspective and contributing to the perceived elevation of distant objects, as shown in Fig. \ref{Angle}.\\
For an eye-level camera angle, the camera is positioned at the same height as the subject and looking straight ahead. The rotation matrix \(R\) is an identity matrix, and the translation vector \(t\) positions the camera at eye level. Hence, the horizon line is centered vertically at \( c_y \).\\
For a low camera angle, the camera is positioned below the subject and looking up. The rotation matrix \(\mathbf{R}\) tilts the camera upwards, and the translation vector \(\mathbf{t}\) positions the camera lower.
Let \(\theta\) be the upward tilt angle so the projection equations become:
\begin{equation} u' = f_x \frac{x_w}{\sin(-\theta) y_w + \cos(-\theta) z_w} + c_x \end{equation}
\begin{equation}v' = f_y \frac{\cos(-\theta) y_w - \sin(-\theta) z_w - l}{\sin(-\theta) y_w + \cos(-\theta) z_w} + c_y \end{equation}
The horizon line shifts higher in the image in comparison to eye-level camera image and therefore, objects further away converge to the horizon line and will appear in the higher portion of the image in comparison to object at the same depth captured at eye-level.\\
For a high camera angle, the camera is positioned above the subject and looking down. The rotation matrix \(\mathbf{R}\) tilts the camera downwards, and the translation vector \(\mathbf{t}\) positions the camera higher with the resulting projection equations as follows:
\begin{equation} u' = f_x \frac{x_w}{\sin(\theta) y_w + \cos(\theta) z_w} + c_x \end{equation}
\begin{equation} v' = f_y \frac{\cos(\theta) y_w - \sin(\theta) z_w + h}{\sin(\theta) y_w + \cos(\theta) z_w} + c_y \end{equation}
The horizon line appear lower in the image. Although objects at larger depth appear lower in the image in comparison to when the camera is held at eye-level, but they will still be higher up in the image than the object closer to the camera.\\
These equations show how the camera angles affect the perspective and the resulting image coordinates. In all three cases, close objects appear larger and more dominant, while distant objects, appearing higher up, seem smaller and further away.
\section{Image orientation estimation}
\label{sec:Image orientation estimation}
The proposed method leverages the distribution of depth in natural images. Recognizing that depth distribution is not arbitrary but follows a discernible pattern, our approach, highlighted in the Algorithm \ref{alg1} begins by examining the depth distribution across the image to infer its orientation.\\
If the input image does not have a depth map, we obtain it using the state-of-the-art monocular depth estimation method Midas \cite {Ranftl2020} and through defocus cues. 
\begin{algorithm}
\caption{Image Orientation Estimation}
\label{alg1}
\begin{algorithmic}[1]

\State REQUIRE image
\State ENSURE refinedOrientation
\State $depthMap \leftarrow \text{depthEstimation}(image)$
\State $quadrants \leftarrow \text{divideIntoQuadrants}(depthMap)$
\State $maxDepthQuadrant \leftarrow \text{findMaxDepthQuadrant}(quadrants)$
\State $coarseOrientation \leftarrow \text{estimateCoarseOrientation}(maxDepthQuadrant)$
\State $refinedOrientation \leftarrow \text{None}$
\State FOR {$angle$ in $rotation Angles$}
      \State \hspace{0.1in} $rotatedDepthMap \leftarrow \text{rotate}(depthMap, angle)$
      \State \hspace{0.1in} $dgcScore \leftarrow \text{calculateDGC}(rotatedDepthMap)$
      \State \hspace{0.1in} $hsaScore \leftarrow \text{calculateHSA}(rotatedDepthMap)$
     \State \hspace{0.1in} $combScore \leftarrow \text{weightedAvg.}(dgcScore, hsaScore)$
    \State \hspace{0.1in} IF {$combScore$ is optimal}
    \State \hspace{0.3in}$refinedOrientation \leftarrow angle$
    \State \hspace{0.1in} ENDIF
\State ENDFOR
\State RETURN $refinedOrientation$
\end{algorithmic}
\end{algorithm}
The depth map is divided into four quadrants: top, bottom, left, and right. This division is predicated on the observation that the distribution of depth in these quadrants varies depending on the orientation of the image. Each quadrant, therefore, contains depth information that potentially indicates the correct orientation of the image. For each quadrant, we calculate the total depth magnitude. This is achieved by summing the depth values within each quadrant as given below:
\begin{equation}
M(Q) = \sum_{i \in Q} D_i,
\end{equation}
where $M(Q)$ represents the total depth magnitude for a given quadrant Q. Based on the calculated depth magnitudes, the quadrant with the highest total depth value suggests the coarse orientation of the image, as given below.
\begin{equation}
\theta_C = \underset{Q \in {Q_{\text{top}}, Q_{\text{bottom}}, Q_{\text{left}}, Q_{\text{right}}}}{\arg\max} \text{Magnitude}(Q)
\end{equation}
The coarse estimate $\theta_C$ can be 0, 90, 180, or 270 degrees, corresponding to the image being upright, rotated left, upside down, or rotated right, respectively.
This coarse orientation provides a crucial starting point for further refining image orientation. It narrows down the possible orientations of the image, allowing for more focused and computationally efficient subsequent processing. The depth map is then further divided into finer quadrants with an increment of 45\textdegree. For instance, if the coarse estimate identifies the first quadrant (top quadrant) as having the maximum cumulative depth, then further refinement focuses specifically within  0\textdegree to 90\textdegree range.\\
The division of the image into smaller quadrants is crucial for defining the range of rotation angles to be explored in later steps, particularly for the DGC and HSA. Relying solely on the coarse prediction could inaccurately define this range. For instance, an image rotated by 80\textdegree might be incorrectly categorized in the second quadrant because of its proximity to 90\textdegree, which is the threshold for the next quadrant instead of the correct first quadrant. Nonetheless, this coarse estimation is vital for accurately establishing the lower boundary of the search range. In a similar vein, an image rotated by 130\textdegree might be misclassified as 135\textdegree, when finer quadrants are estimated, setting the starting range from 135\textdegree rather than 90\textdegree. Here, the initial prediction correctly places it in the quadrant starting at 90\textdegree, thereby accurately setting the lower boundary. 
\subsection{Depth gradient consistency}
Depth gradients (changes in depth across the scene) tend to follow certain patterns. Typically, these gradients are consistent within specific regions of the image. For instance, in an image of a natural landscape, the depth gradient is relatively uniform across the sky or ground areas. This uniformity is disrupted if the image is incorrectly oriented. By assessing the consistency of these gradients at various rotational angles, we can precisely identify the angle that best aligns with the natural depth distribution of the scene, thereby achieving a finer scale, up to a 10\degree estimate.\\
{The depth gradient consistency calculation involves systematically evaluating local depth gradients within pre-defined box regions in the rotated depth map \(D_{rot}\). These box regions are of size \(10 \times 10\), and they systematically cover the depth map to ensure robust assessment across the entire image. The depth gradient for each box is calculated by averaging the vertical depth gradients (differences in depth values between adjacent pixels) within the box. The Depth Gradient Consistency for a given rotation angle \(\theta\) is then calculated as the sum of the absolute vertical gradients across all boxes. This is mathematically expressed as:}
\begin{equation}
\text{DGC}(\theta) = \sum_{\text{boxes}} \left| \nabla_y D_{\text{rot}}(\theta) \right|,
\end{equation}
where \( \text{DGC}(\theta) \) denotes the Depth Gradient Consistency for the rotation angle \(\theta\), \(\nabla_y\) is the vertical gradient operator, and \(D_{rot}(\theta)\) is the depth map rotated by \(\theta\) degrees.
\subsection{Horizontal Symmetry Analysis}
Horizontal Symmetry Analysis also provides additional cues in assessing image orientation, particularly when utilizing depth information. This is because many natural and man-made scenes tend to exhibit a certain level of symmetry along the horizontal axis, especially when the image is correctly oriented. This symmetry can be observed in various contexts, ranging from natural landscapes to architectural structures, where the left and right halves of an image often mirror each other to a certain degree. An imbalance in this symmetry might suggest an incorrect orientation of the image.
To quantitatively evaluate the horizontal symmetry, the depth map \(D_{rot}\), after being rotated by an angle \(\theta\), is divided into left and right halves. The HSA score is then calculated by summing the absolute differences between corresponding pixels in these two halves. This is mathematically expressed as:
\begin{equation}
\text{HSA}(\theta) = \sum_{i} \left| D_{\text{rot,left}}(i) - D_{\text{rot,right}}(i) \right|,
\end{equation}
where \( \text{HSA}(\theta) \) denotes the Horizontal Symmetry Analysis score for the rotation angle \(\theta\), and \(D_{rot,left}(i)\) and \(D_{rot,right}(i)\) are the depth values at pixel \(i\) in the left and right halves of the rotated depth map, respectively.
\\
{Once the correct coarse orientation is identified, the algorithm refines the angle within a specific range based on the quadrant. This range is typically defined as \(\pm 45^\circ\) around the coarse angle estimate. For instance, if the coarse angle is \(90^\circ\), the fine-scale refinement is performed within the range \([45^\circ, 135^\circ]\). Within this range, the depth map is rotated in fine increments of \(10^\circ\). For each rotated depth map (\(D_{rot}(\theta)\)), the algorithm evaluates the orientation quality by calculating the DGC and HSA scores.} \textcolor{black}{These scores are then combined into a single objective function:
\begin{equation}
C(\theta) = \alpha \cdot \text{DGC}(\theta) + \beta \cdot \text{HSA}(\theta),
\end{equation}
where \(\alpha\) and \(\beta\) are weights that balance the influence of DGC and HSA, respectively. The algorithm iterates through all angles in the range, computing \(C(\theta)\) for each candidate angle.
The angle that minimizes the cost function \(C(\theta)\) is selected as the refined orientation:
\begin{equation}
\theta_{opt} = \arg\min_\theta C(\theta)
\end{equation}
This process ensures that the orientation aligns with the natural depth distribution and symmetry characteristics of the scene, refining the coarse angle to a fine-scale estimate. The final refined angle \(\theta_{opt}\) is then returned as the output of the algorithm.}
{The assignment of weights \(\alpha\) and \(\beta\) is a critical aspect of the objective function. These weights balance the influence of depth gradient consistency and horizontal symmetry in determining the image orientation. The current implementation uses \(\alpha = 0.8\) and \(\beta = 0.2\), emphasizing depth gradient consistency as the primary driver of orientation estimation. However, sensitivity analysis reveals that small adjustments to \(\alpha\) and \(\beta\) can significantly affect performance. This is because DGC plays a dominant role in structured and planar scenes, while HSA contributes more in symmetrical settings, such as architectural imagery.
The weights were chosen empirically to achieve optimal performance across diverse datasets, but they can be further tuned for application-specific scenarios or specific image types. Future work may explore adaptive weight selection based on scene characteristics or iterative optimization methods to balance the trade-off between DGC and HSA dynamically.}
{
\subsection{Depth from Defocus:}
\label{DOD}
The proposed orientation estimation method do not require, precise depth measurement at pixel level, rather the relative depth variations across different regions of the image, therefore, we also use defocus cues to infer depth. Each quadrant of the input image (same as the depth map quadrant discussed previously) is analyzed independently to estimate blur characteristics. For each quadrant 
\( Q_i \), we compute a blur kernel \( K_i \) using a defocus blur estimation technique \cite{chen2021blind}. The intensity of blur, \( B_i \), is derived from \( K_i \) by analyzing the spread of the kernel.\\
To mitigate noise in the blur kernel, a thresholding operation is applied:
\[
K_i'(x, y) =
\begin{cases}
K_i(x, y), & \text{if } K_i(x, y) \geq \tau, \\
0, & \text{otherwise}.
\end{cases}
\]
Here, \( \tau \) is the threshold value. Based on the principle of depth from defocus, regions with higher blur intensity \( B_i \) are inferred to be further away from the camera. The quadrant with the maximum blur intensity \( B_{\text{max}} = \max(B_1, B_2, B_3, B_4) \) is considered to be the furthest. \\
Depth-from-defocus techniques are typically applied to images that exhibit significant defocus blur \cite{11475880},\cite{ALAM201941}, as the variation in blur across the scene provides a strong cue for estimating relative depth. However, the proposed method addresses the challenge of utilizing defocus analysis in images that are relatively sharper, with limited blur variations. These sharper images, often characterized by better depth of field, pose unique challenges as the difference in estimated blur kernel characteristics across image regions (quadrants) becomes less pronounced. This necessitates a more nuanced approach to kernel analysis. To effectively analyze defocus cues in such images, we dynamically adjust the noise reduction threshold applied to the blur kernels. To handle cases with negligible differences in blur intensity across quadrants, we introduce a confidence measure. If the difference between the maximum and minimum blur intensities across quadrants \( (B_{\text{max}} - B_{\text{min}}) \) is below a predefined threshold \( \epsilon \), the estimation is labeled as low confidence. In such cases, the threshold \( \tau \) is dynamically adjusted as follows:
\begin{equation}
\tau' = \tau \cdot \gamma, \quad \gamma < 1,
\end{equation}
where \( \gamma \) is a scaling factor that ensures \( \tau \) decreases iteratively. This iterative adjustment continues until:
\begin{equation}
\Delta B \geq \epsilon,
\end{equation}
ensuring sufficient differentiation between quadrants. This dynamic adjustment is critical for leveraging defocus cues in sharper images. It allows the algorithm to retain meaningful kernel contributions while filtering out noise, enhancing the reliability of quadrant-level blur intensity comparisons. \\  
The relative positions of the quadrants and the identified farthest region guide the estimation of image orientation, as mentioned above in case of traditional depth map scenario. The process inherently accounts for perspective effects and spatial blur variations. Other methods that enable depth estimation from a single image, discussed in the \ref{literature} section, provide several high-level cues for depth inference. However, the challenge in the literature arises at the level of detail, where pixel-level accuracy and fine details are most critical, However, in the proposed algorithm such detail information is not required.}
\color{black}

\section{Experimental results}
\label{results}
\begin{figure*}[!h]
\centering
\hspace{-0.50in}
\raisebox{.7\height}{ \rotatebox[origin=]{90}{Input}}
\begin{subfigure}{1.2in}
{\includegraphics[width=1.2in,height=1.0in,keepaspectratio]{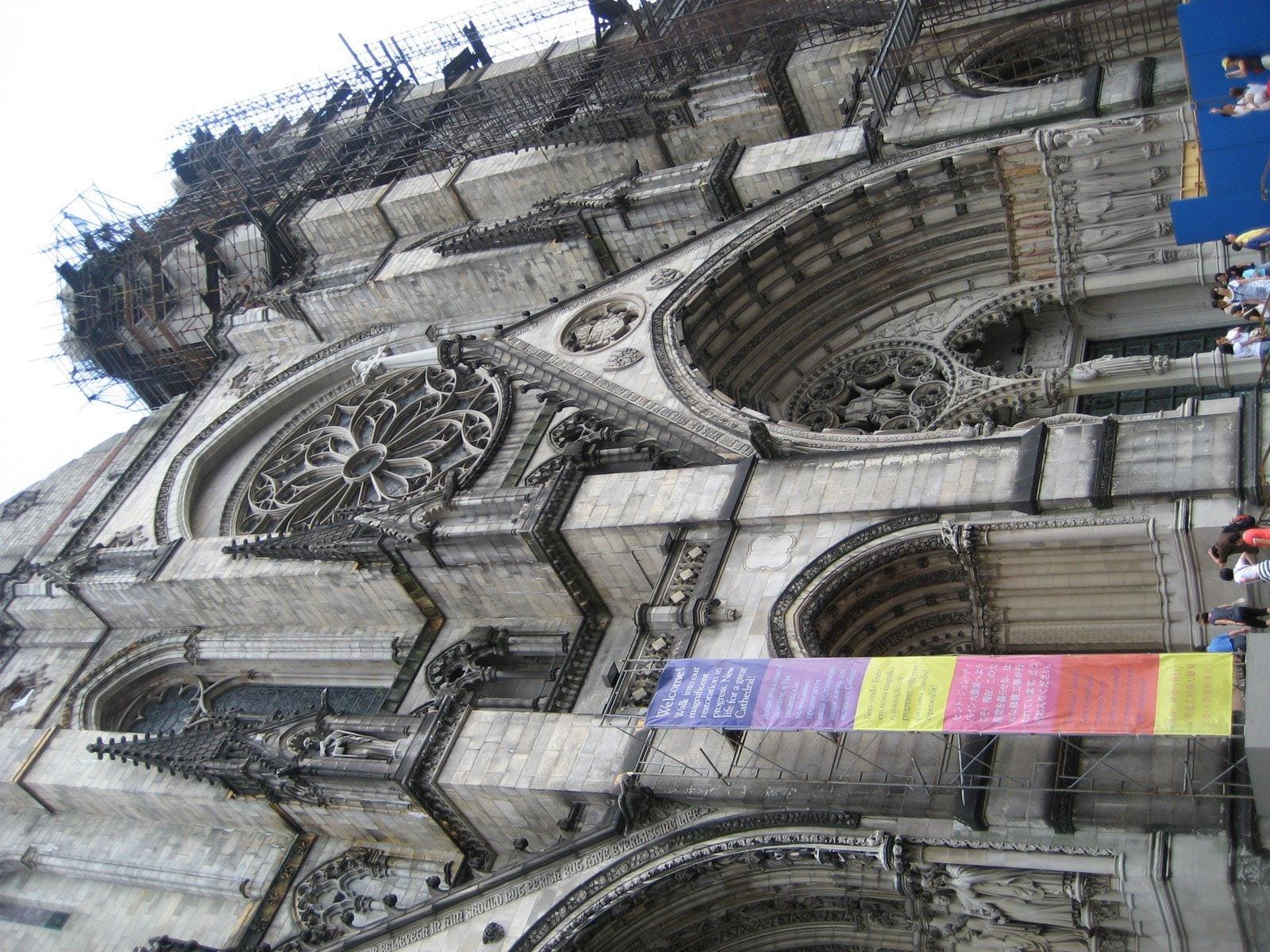}}
\end{subfigure}
\begin{subfigure}{1.2in}
{\includegraphics[width=1.2in,height=1.0in,keepaspectratio]{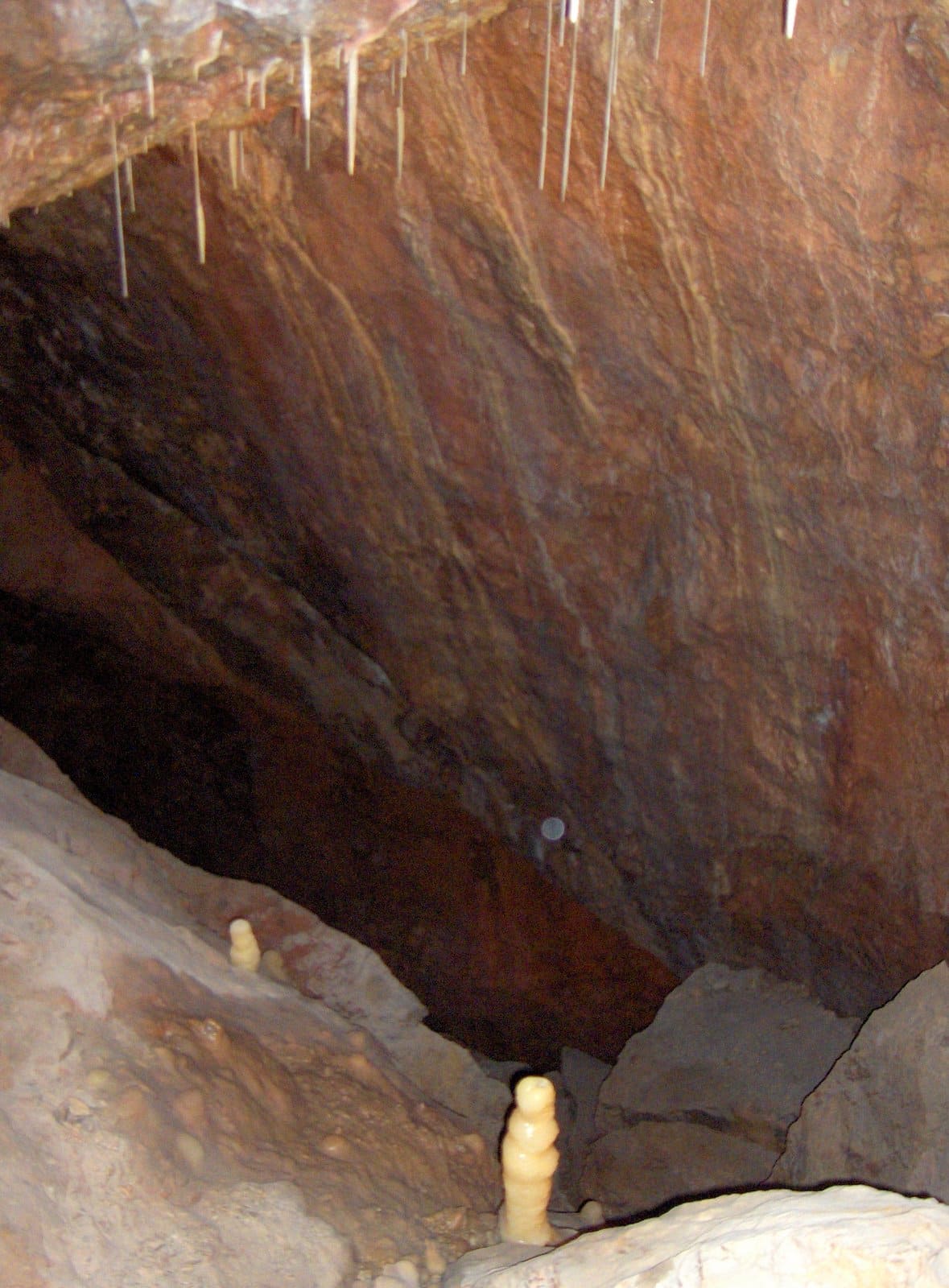}}
\end{subfigure}
\begin{subfigure}{1.2in}
\hspace{-0.5in}
{\includegraphics[width=1.2in,height=1.0in,keepaspectratio]{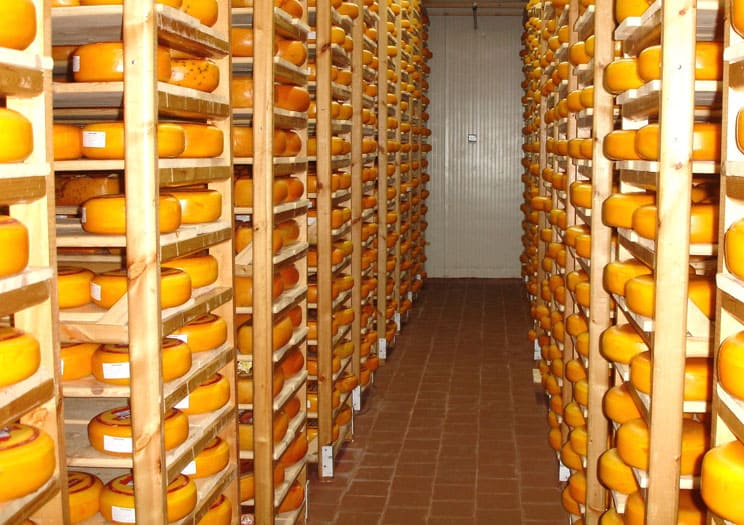}}
\end{subfigure}
\hspace{-0.5in}
\begin{subfigure}{1.2in}
{\includegraphics[width=1.2in,height=1.0in,keepaspectratio]{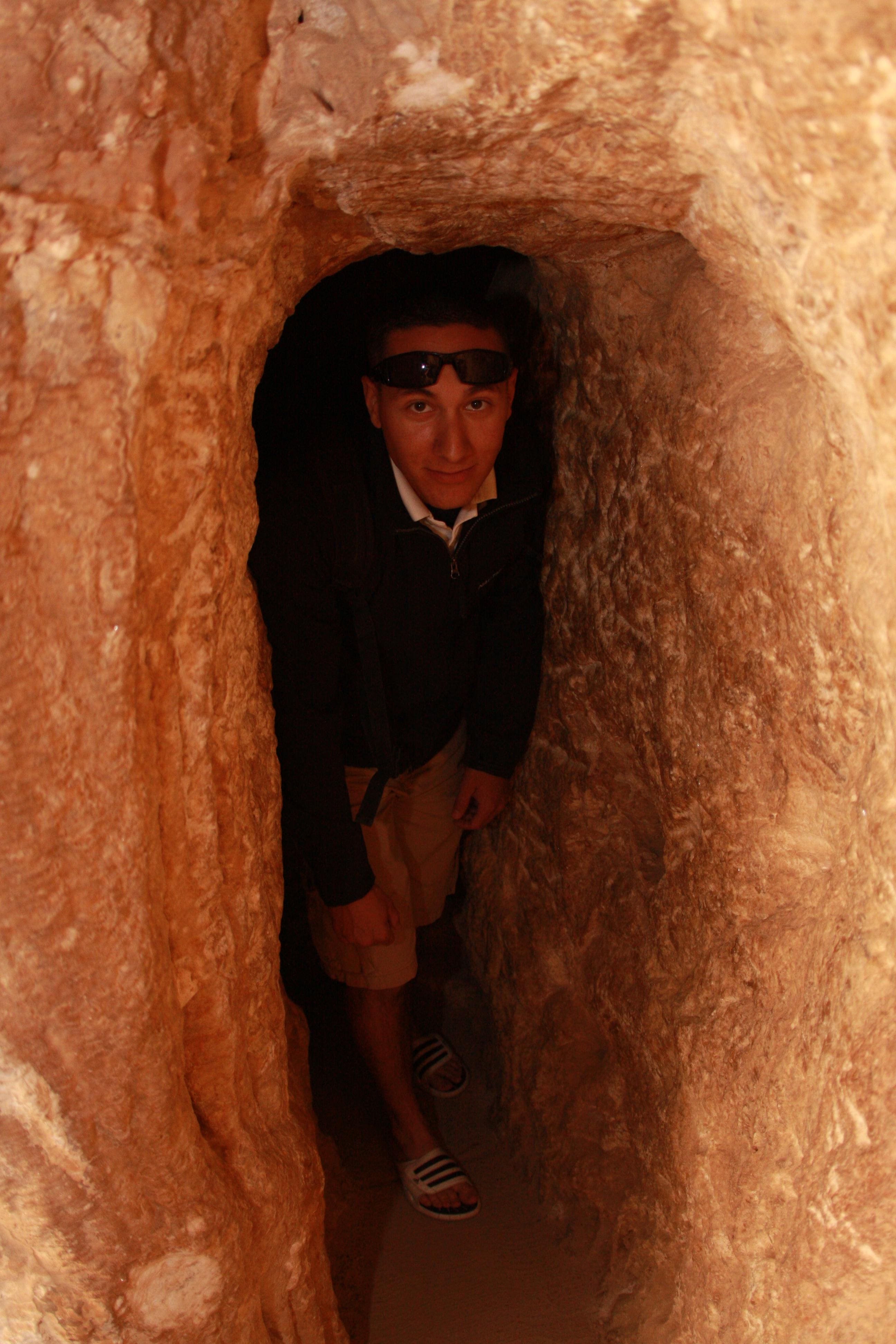}}
\end{subfigure}
\hspace{-0.6in}
\begin{subfigure}{1.2in}
{\includegraphics[width=1.2in,height=1.0in,keepaspectratio]{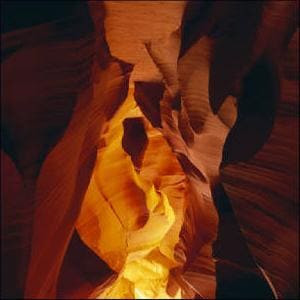}}
\end{subfigure}
\hspace{-0.2in}
\begin{subfigure}{1.2in}
{\includegraphics[width=1.2in,height=1.2in,keepaspectratio]{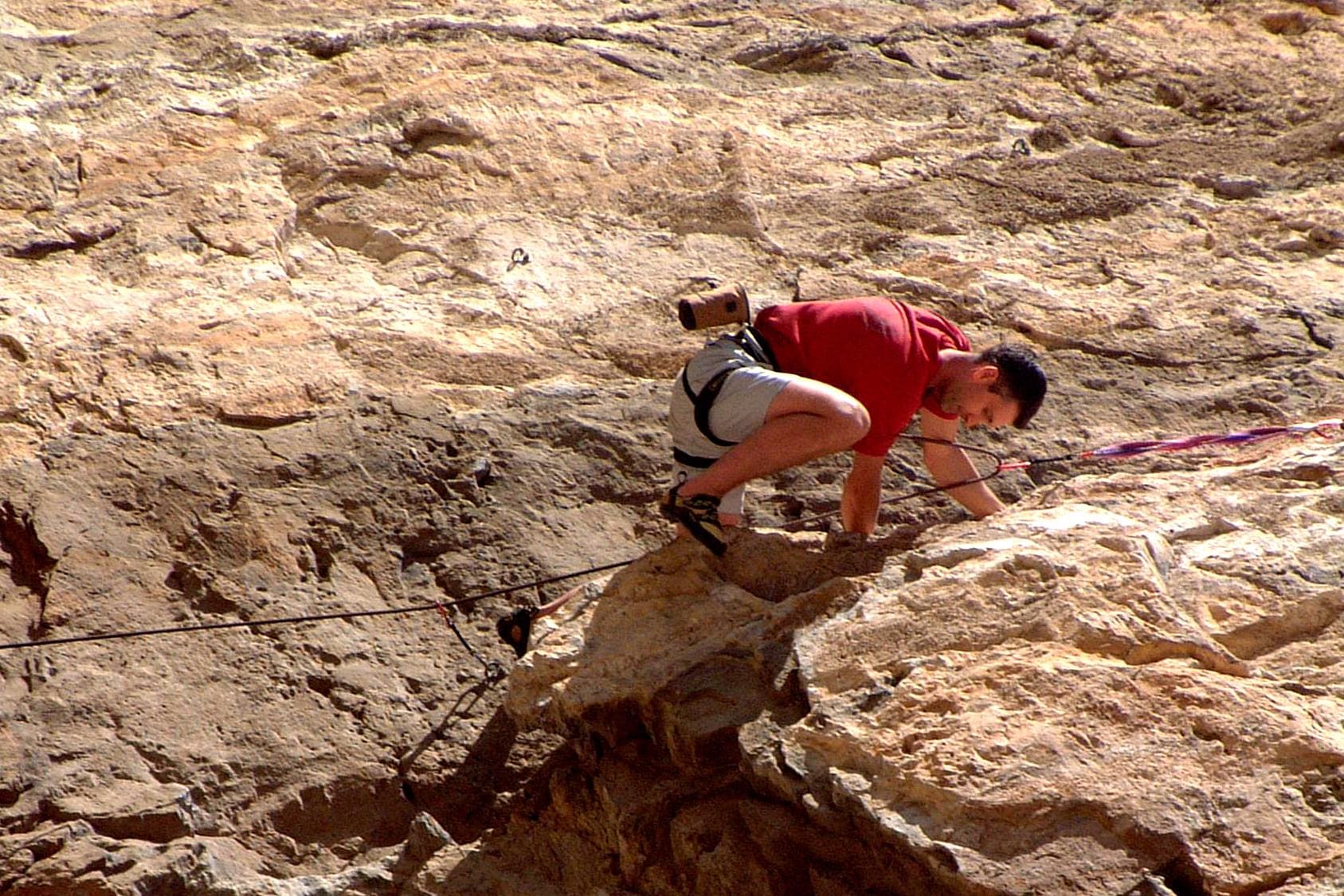}}
\end{subfigure}\\
\raisebox{.7\height}{ \rotatebox[origin=]{90}{Input}}
\begin{subfigure}{1.2in}
{\includegraphics[width=1.2in,height=1.2in,keepaspectratio]{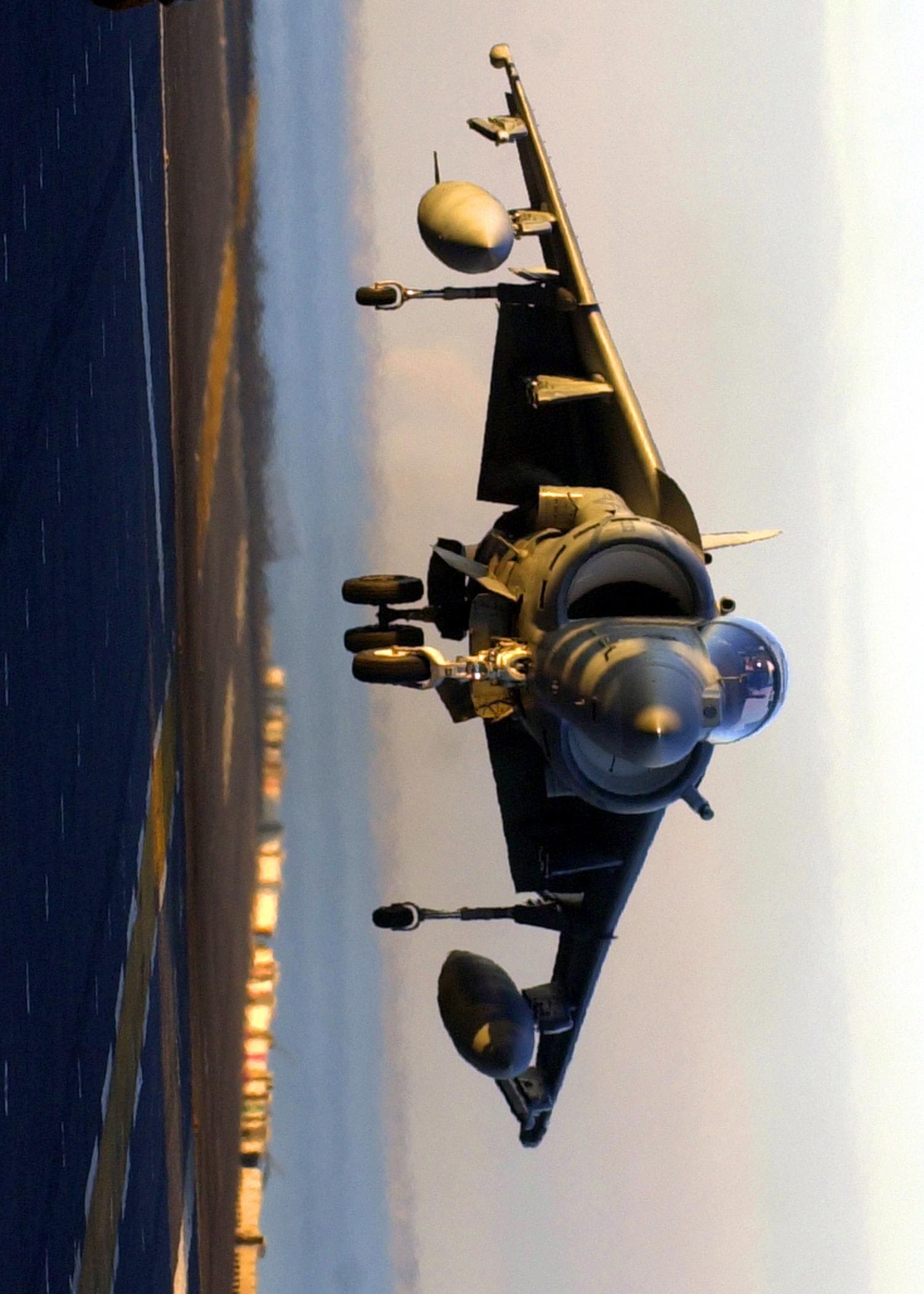}}
\end{subfigure}
\hspace{-0.4in}
\begin{subfigure}{1.2in}
{\includegraphics[width=1.2in,height=1.2in,keepaspectratio]{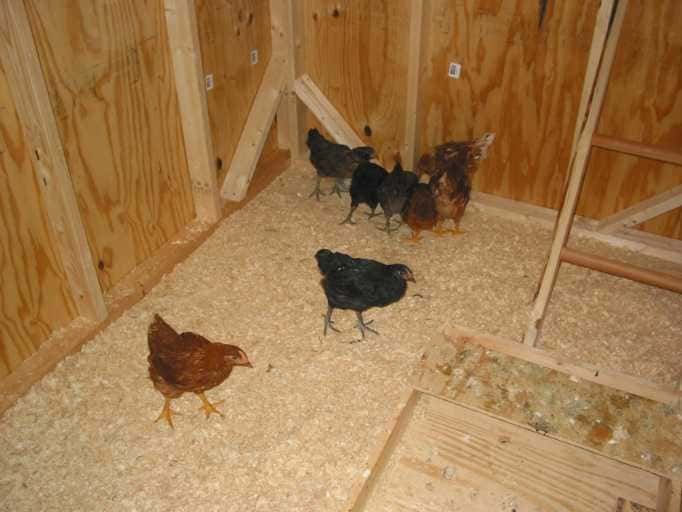}}
\end{subfigure}
\begin{subfigure}{1.2in}
{\includegraphics[width=1.2in,height=1.2in,keepaspectratio]{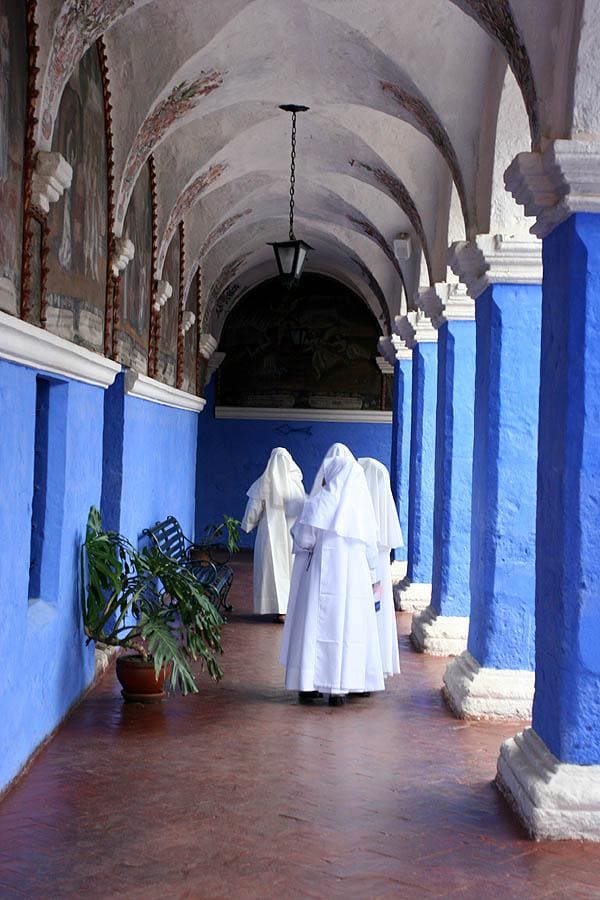}}
\end{subfigure}
\hspace{-0.4in}
\begin{subfigure}{1.2in}
{\includegraphics[width=1.2in,height=1.2in,keepaspectratio]{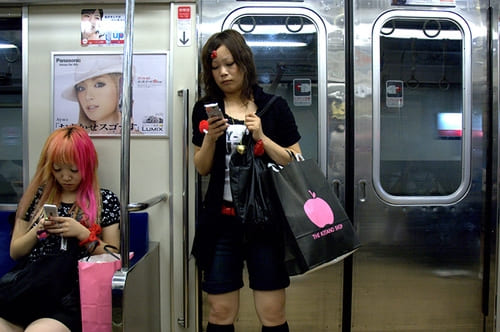}}
\end{subfigure}
\begin{subfigure}{1.2in}
{\includegraphics[width=1.2in,height=1.2in,keepaspectratio]{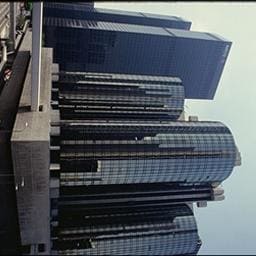}}
\end{subfigure}
\begin{subfigure}{1.2in}
{\includegraphics[width=1.2in,height=1.2in,keepaspectratio]{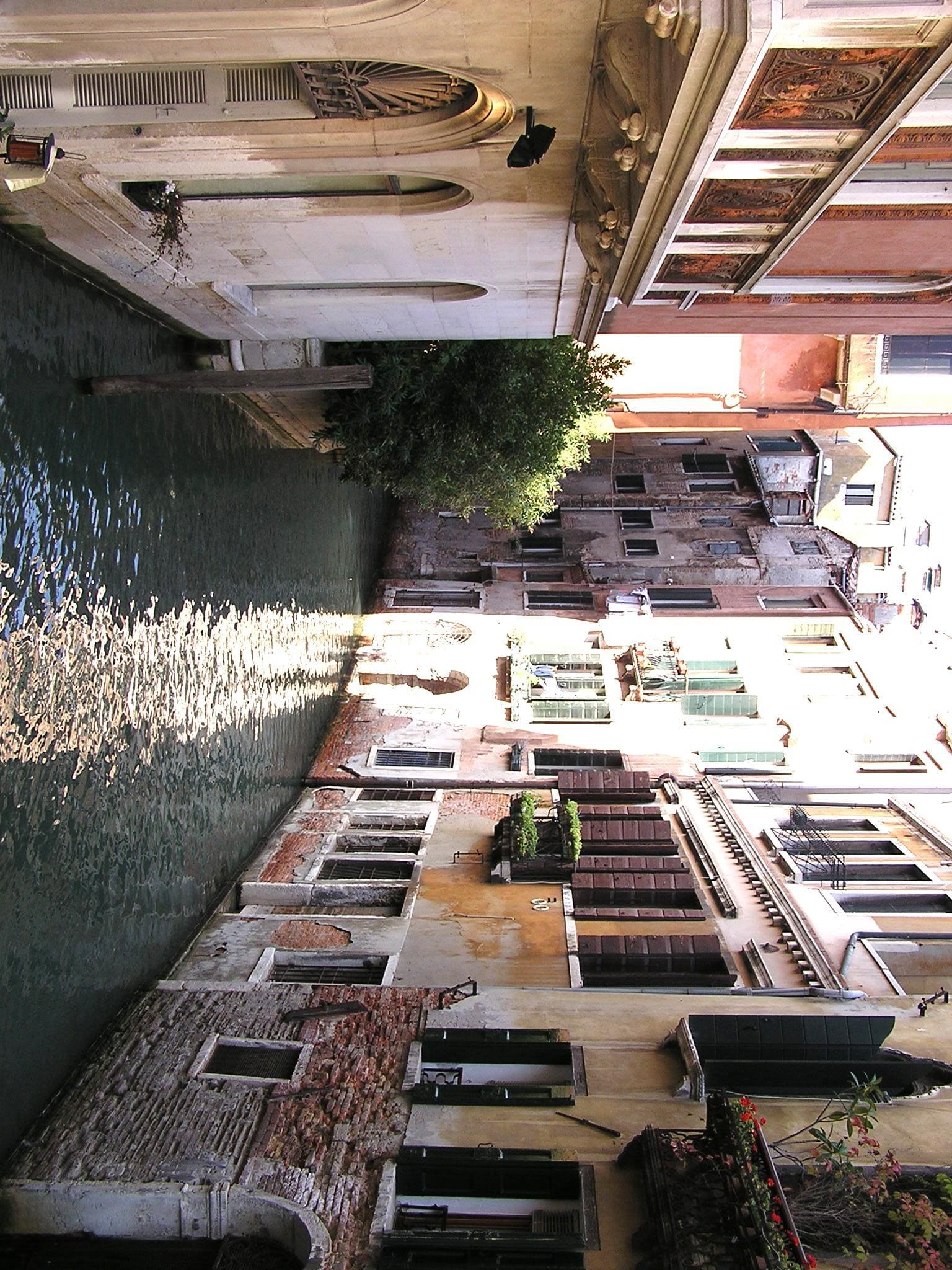}}
\end{subfigure}\\
\hspace{-0.15in}
\raisebox{.7\height}{ \rotatebox[origin=]{90}{Depth-map}}
\begin{subfigure}{1.2in}
{\includegraphics[width=1.2in,height=1.2in,keepaspectratio]{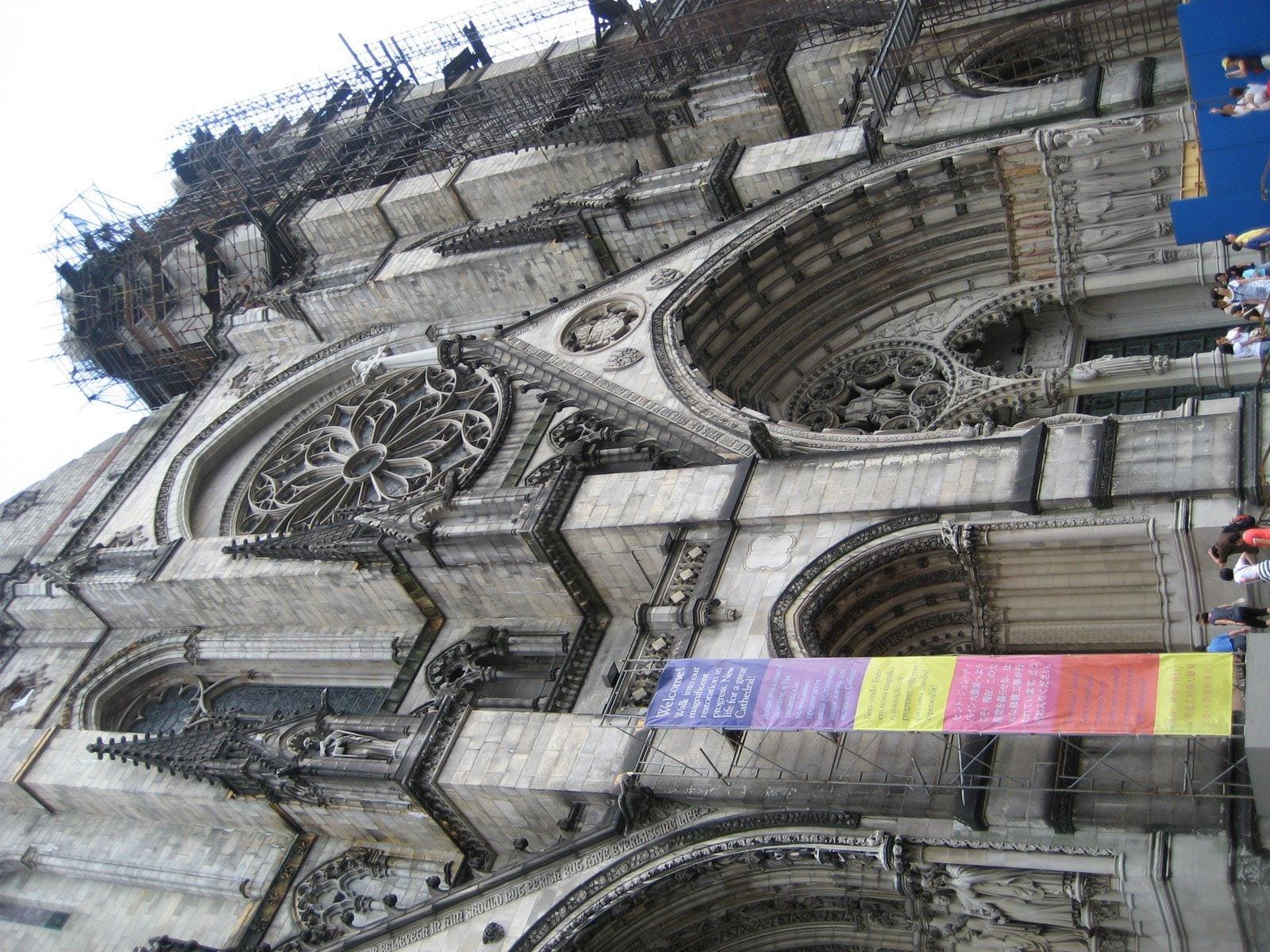}}
\end{subfigure}
\begin{subfigure}{1.2in}
{\includegraphics[width=1.2in,height=1.2in,keepaspectratio]{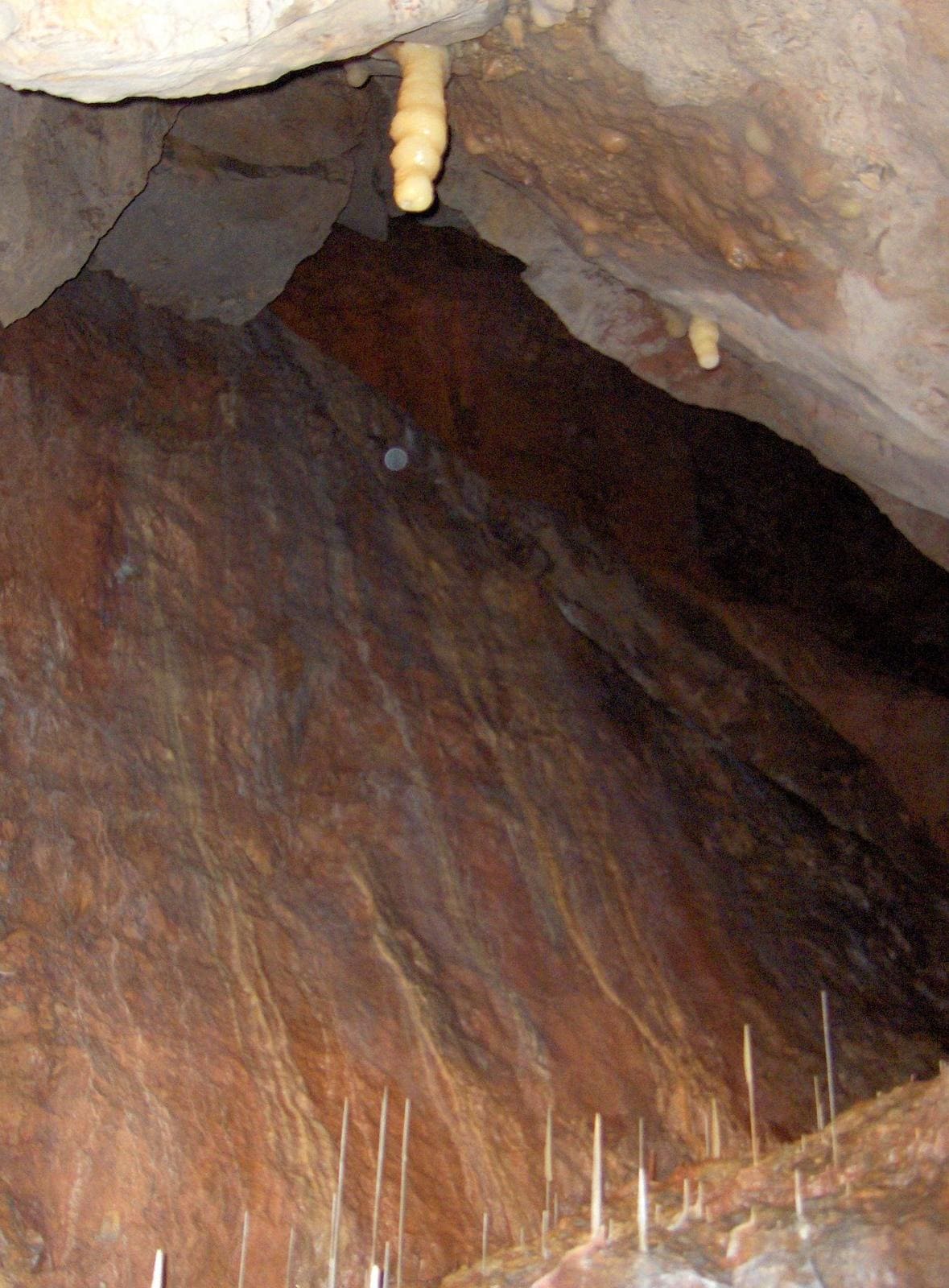}}
\end{subfigure}
\begin{subfigure}{1.2in}
\hspace{-0.35in}
{\includegraphics[width=1.2in,height=1.2in,keepaspectratio]{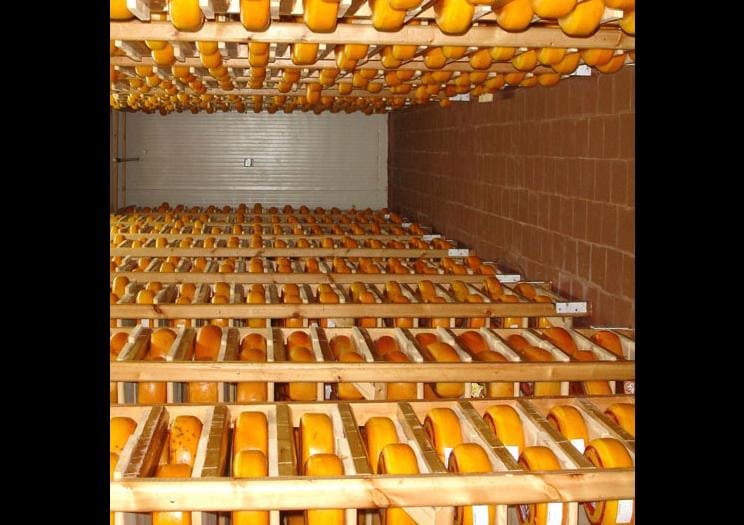}}
\end{subfigure}
\hspace{-0.4in}
\begin{subfigure}{1.2in}
{\includegraphics[width=1.2in,height=1.2in,keepaspectratio]{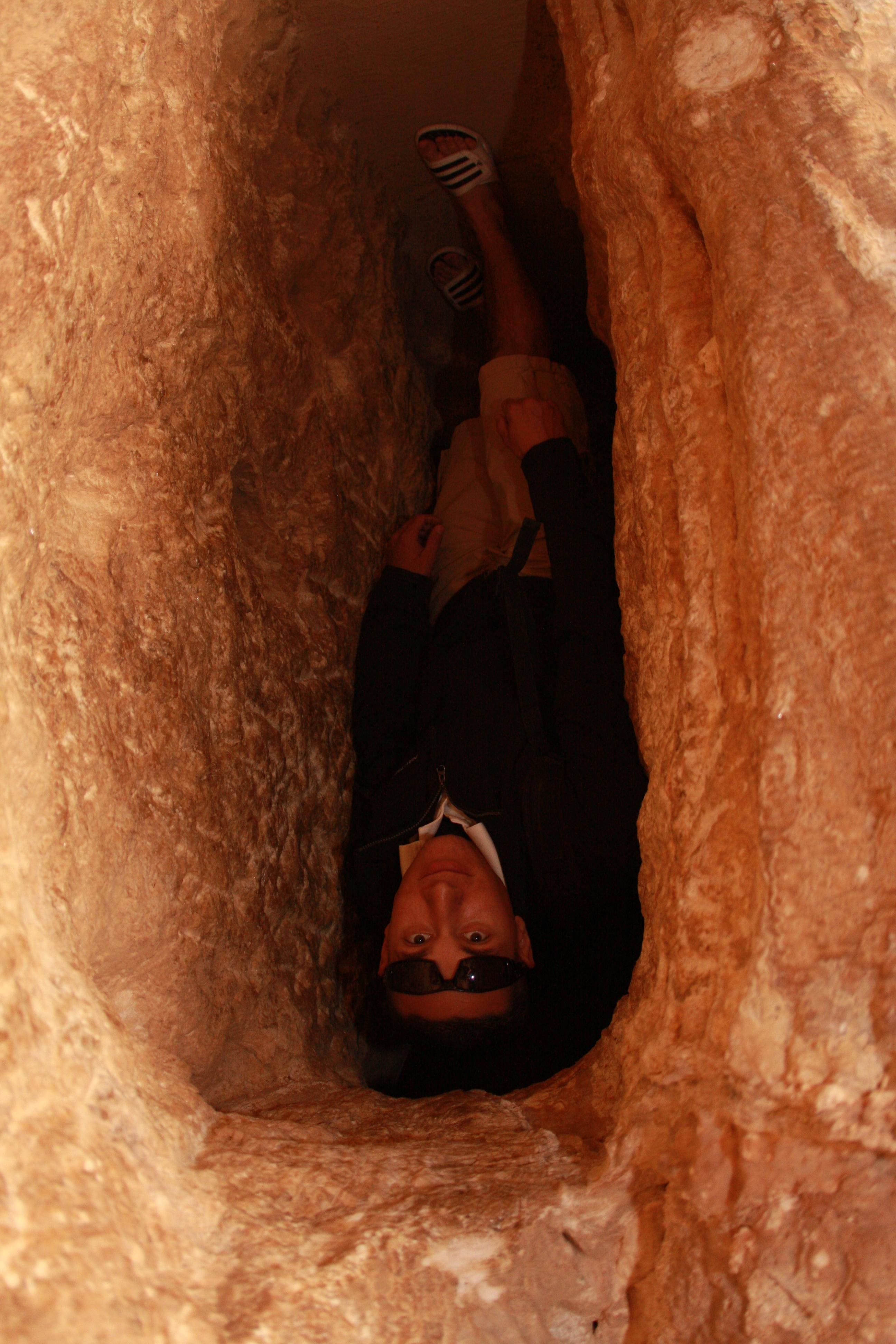}}
\end{subfigure}
\hspace{-0.5in}
\begin{subfigure}{1.2in}
{\includegraphics[width=1.2in,height=1.2in,keepaspectratio]{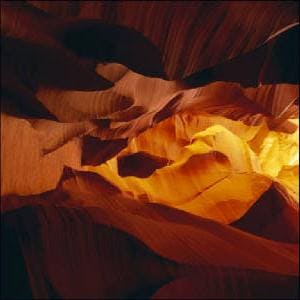}}
\end{subfigure}
\hspace{-0.1in}
\begin{subfigure}{1.2in}
{\includegraphics[width=1.2in,height=1.2in,keepaspectratio]{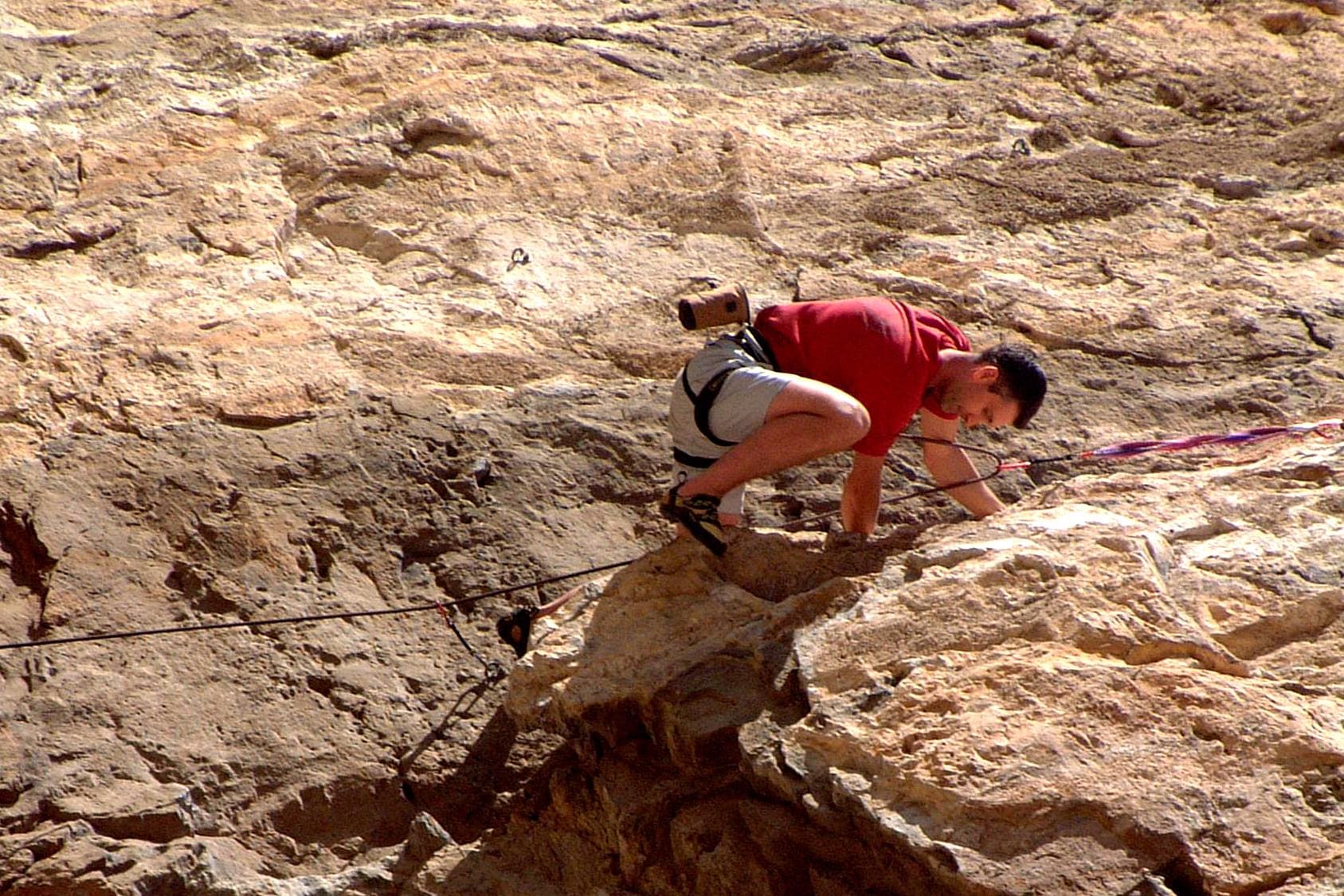}}
\end{subfigure}\\
\raisebox{.7\height}{ \rotatebox[origin=]{90}{Depth-map}}
\begin{subfigure}{1.2in}
{\includegraphics[width=1.2in,height=1.2in,keepaspectratio]{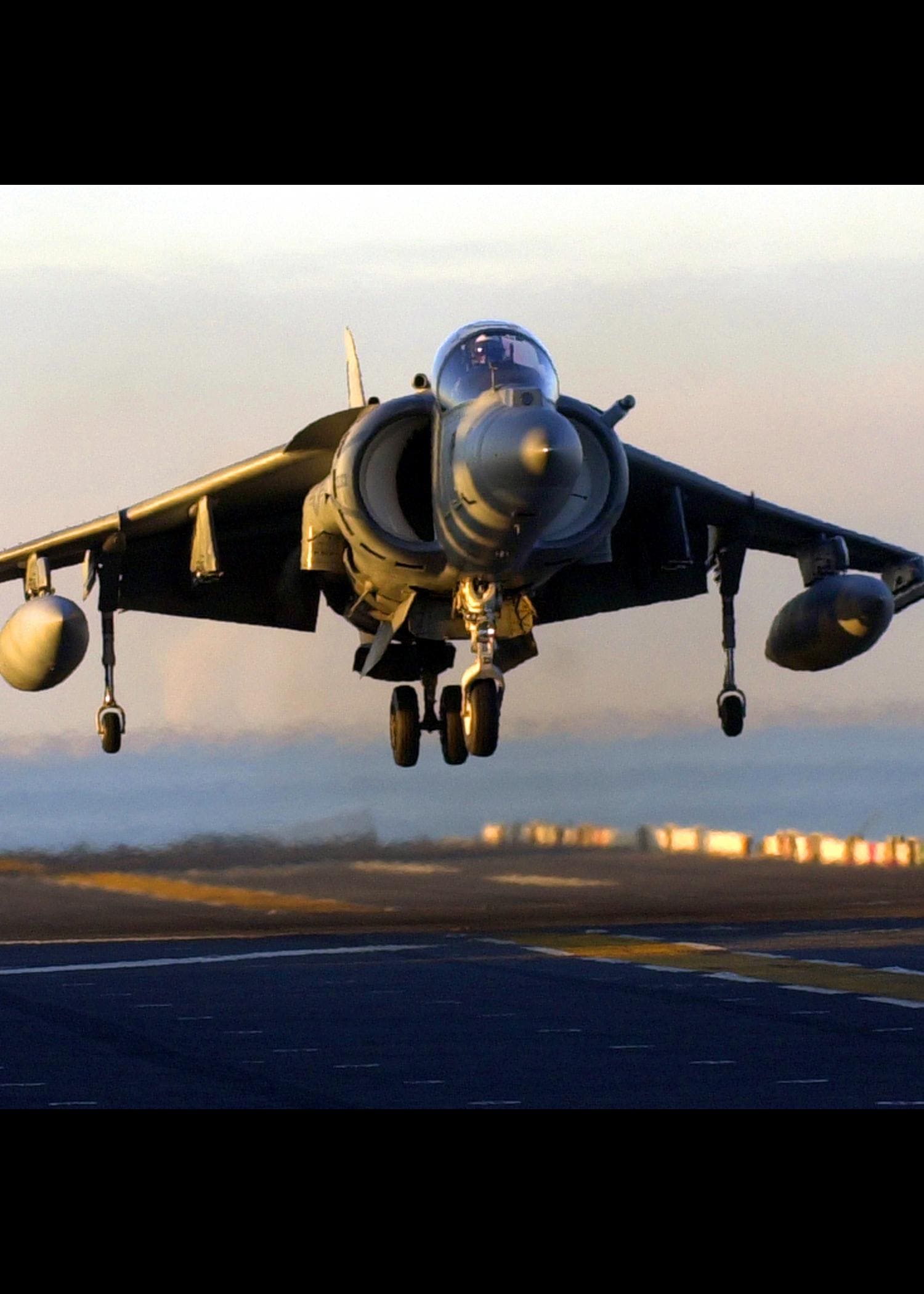}}
\end{subfigure}
\hspace{-0.4in}
\begin{subfigure}{1.2in}
{\includegraphics[width=1.2in,height=1.2in,keepaspectratio]{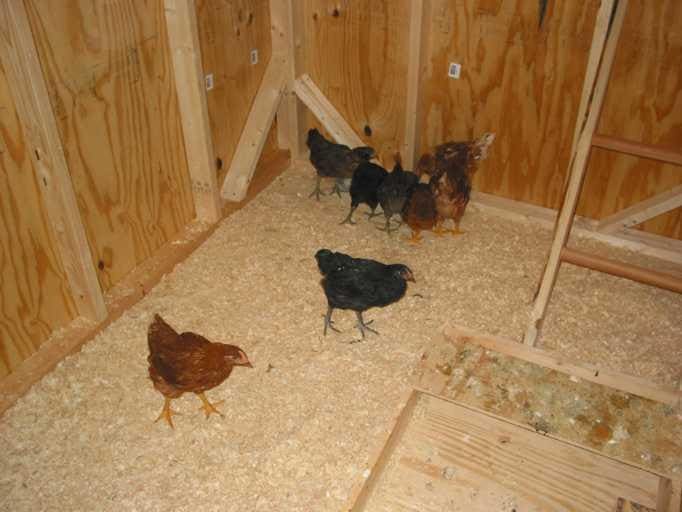}}
\end{subfigure}
\begin{subfigure}{1.2in}
{\includegraphics[width=1.2in,height=1.2in,keepaspectratio]{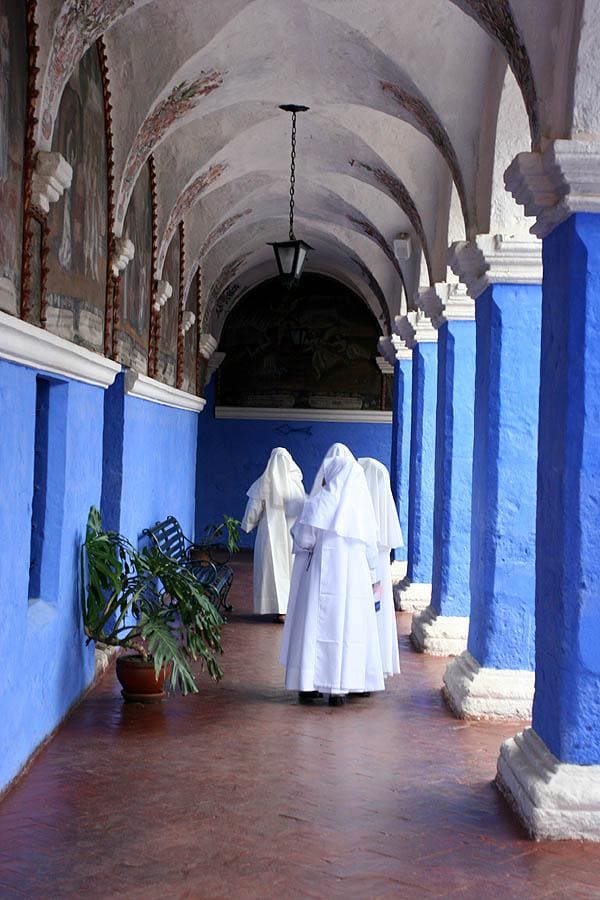}}
\end{subfigure}
\hspace{-0.4in}
\begin{subfigure}{1.2in}
{\includegraphics[width=1.2in,height=1.2in,keepaspectratio]{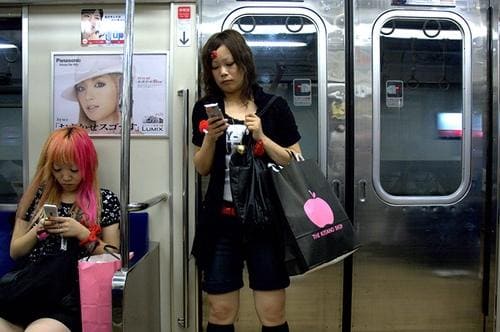}}
\end{subfigure}
\begin{subfigure}{1.2in}
{\includegraphics[width=1.2in,height=1.2in,keepaspectratio]{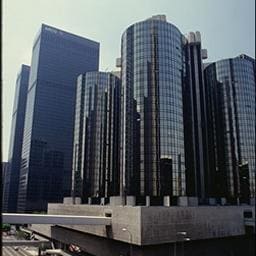}}
\end{subfigure}
\begin{subfigure}{1.2in}
{\includegraphics[width=1.2in,height=1.2in,keepaspectratio]{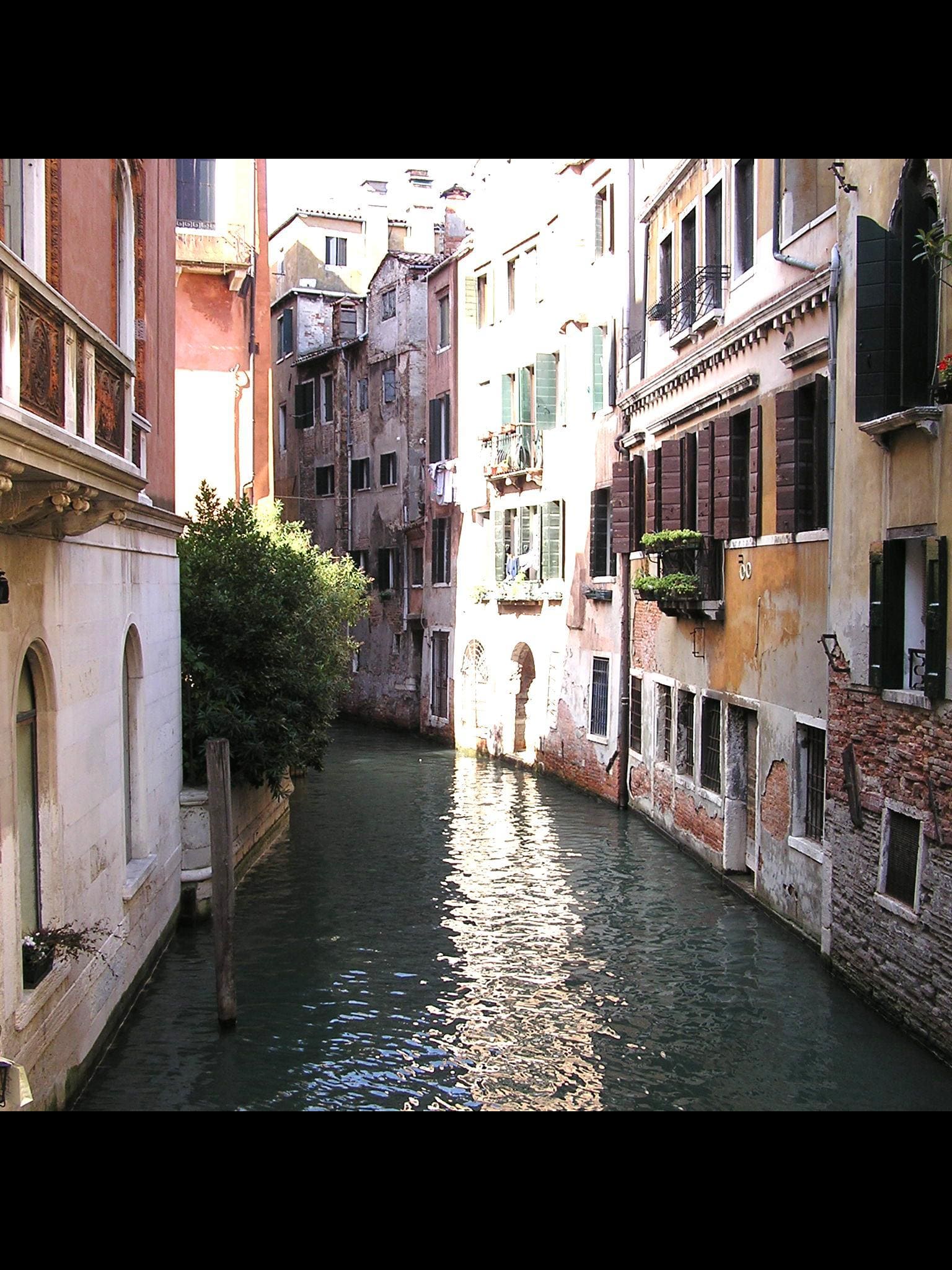}}
\end{subfigure}\\
\hspace{-0.25in}
\raisebox{.7\height}{ \rotatebox[origin=]{90}{Dfd}}
\begin{subfigure}{1.2in}
{\includegraphics[width=1.2in,height=1.2in,keepaspectratio]{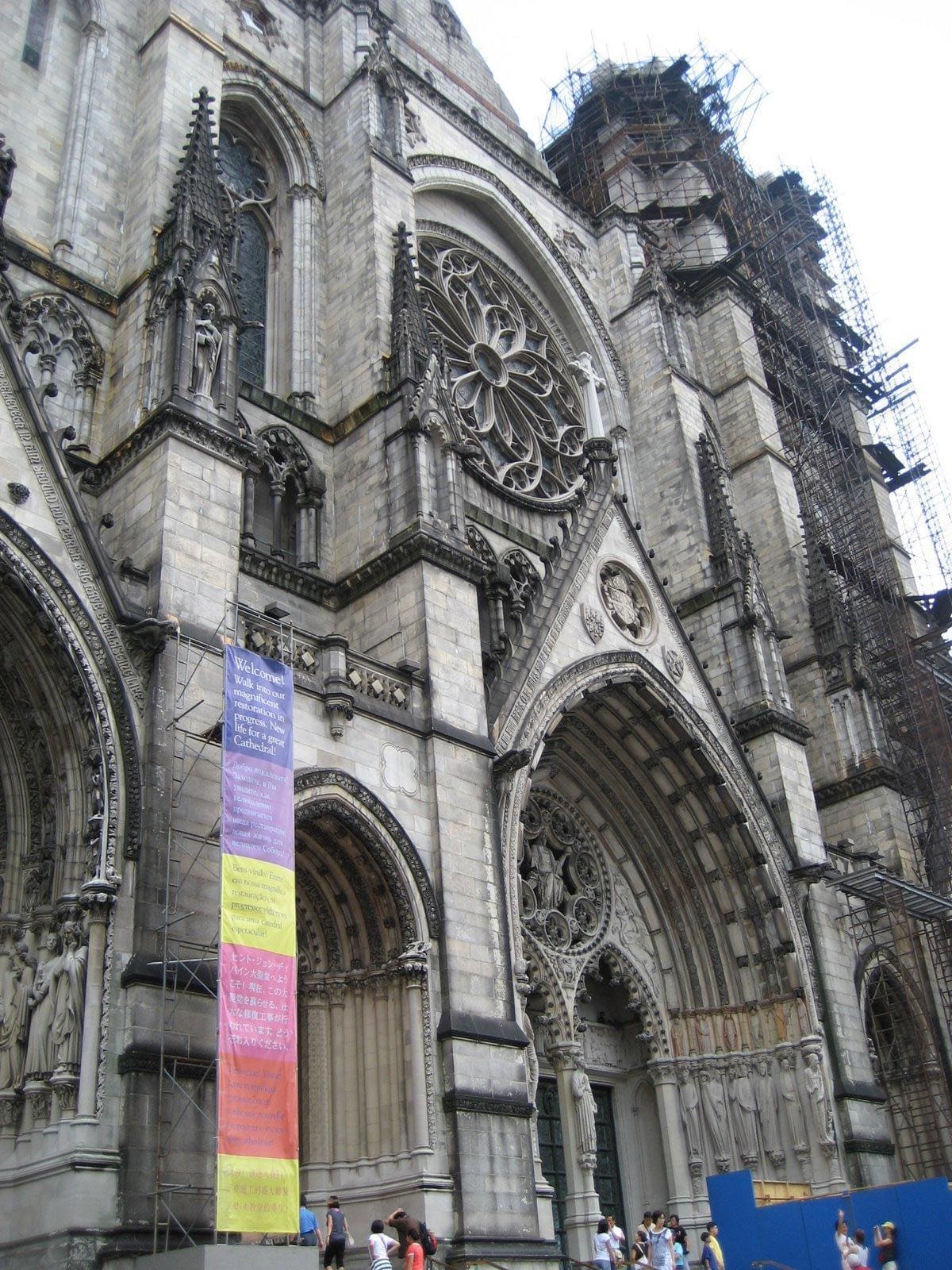}}
\end{subfigure}
\hspace{-0.35in}
\begin{subfigure}{1.2in}
{\includegraphics[width=1.2in,height=1.2in,keepaspectratio]{images/2fixed.jpg}}
\end{subfigure}
\hspace{-0.35in}
\begin{subfigure}{1.2in}
{\includegraphics[width=1.2in,height=1.2in,keepaspectratio]{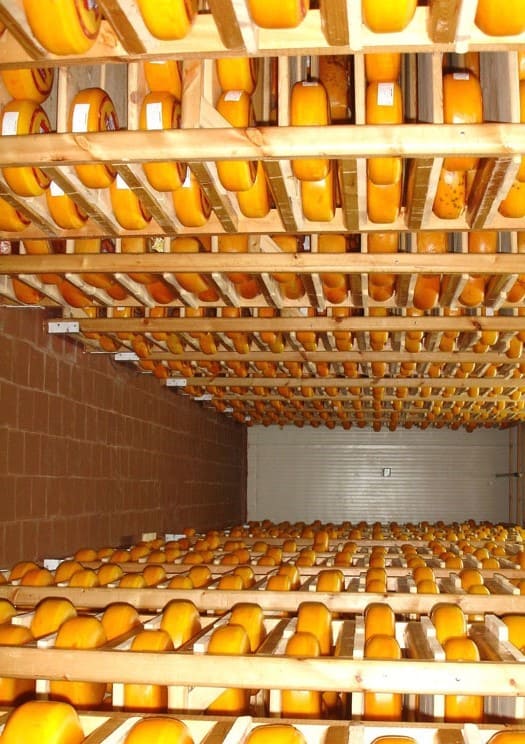}}
\end{subfigure}
\hspace{-0.35in}
\begin{subfigure}{1.2in}
{\includegraphics[width=1.2in,height=1.2in,keepaspectratio]{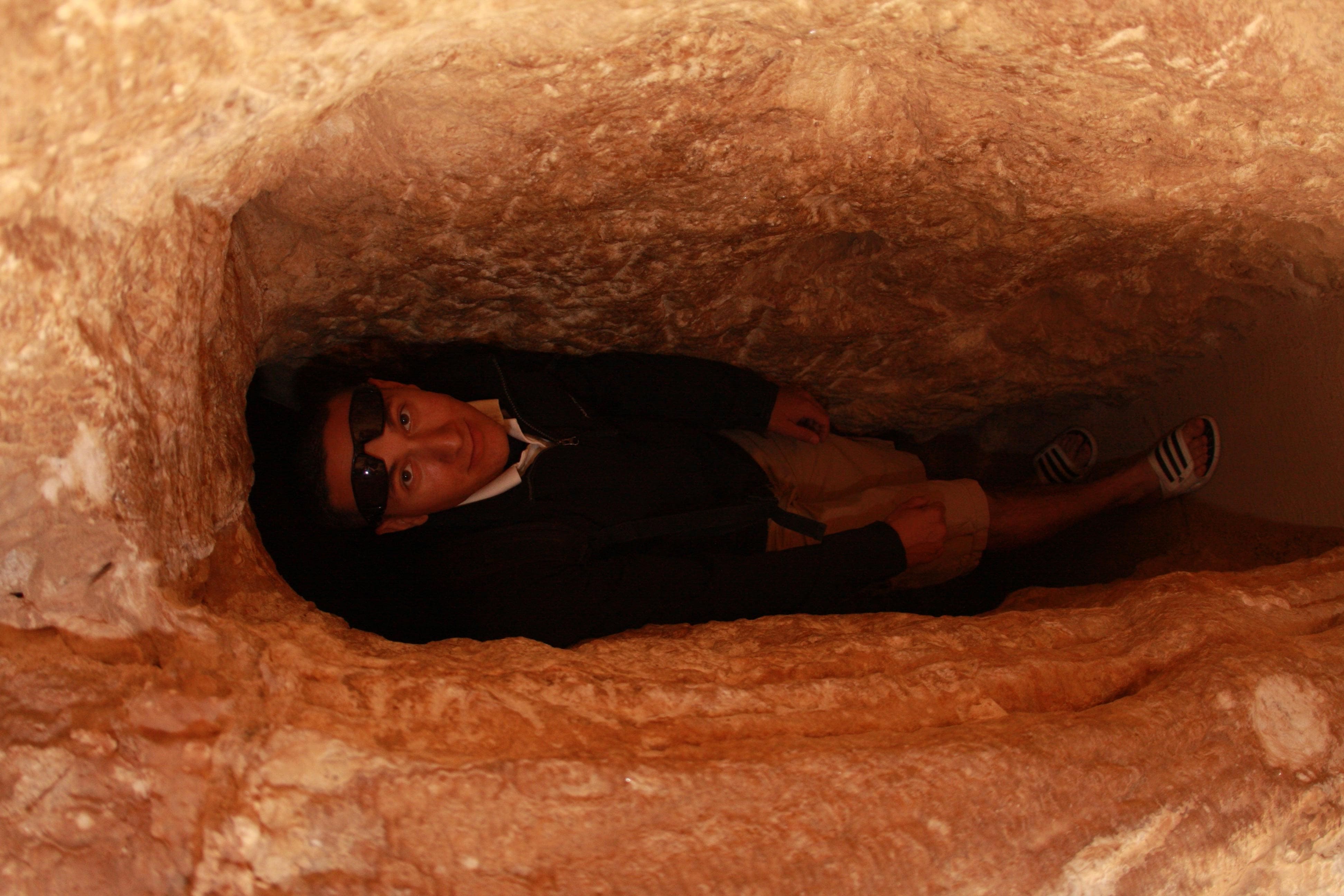}}
\end{subfigure}
\begin{subfigure}{1.2in}
{\includegraphics[width=1.2in,height=1.2in,keepaspectratio]{images/15.jpg}}
\end{subfigure}
\begin{subfigure}{1.2in}
{\includegraphics[width=1.2in,height=1.2in,keepaspectratio]{images/22fixed.jpg}}
\end{subfigure}\\
\raisebox{.7\height}{ \rotatebox[origin=]{90}{Dfd}}
\begin{subfigure}{1.2in}
{\includegraphics[width=1.2in,height=1.2in,keepaspectratio]{images/6fixed.jpg}}
\end{subfigure}
\hspace{-0.4in}
\begin{subfigure}{1.2in}
{\includegraphics[width=1.2in,height=1.2in,keepaspectratio]{images/4fixed.jpg}}
\end{subfigure}
\begin{subfigure}{1.2in}
{\includegraphics[width=1.2in,height=1.2in,keepaspectratio]{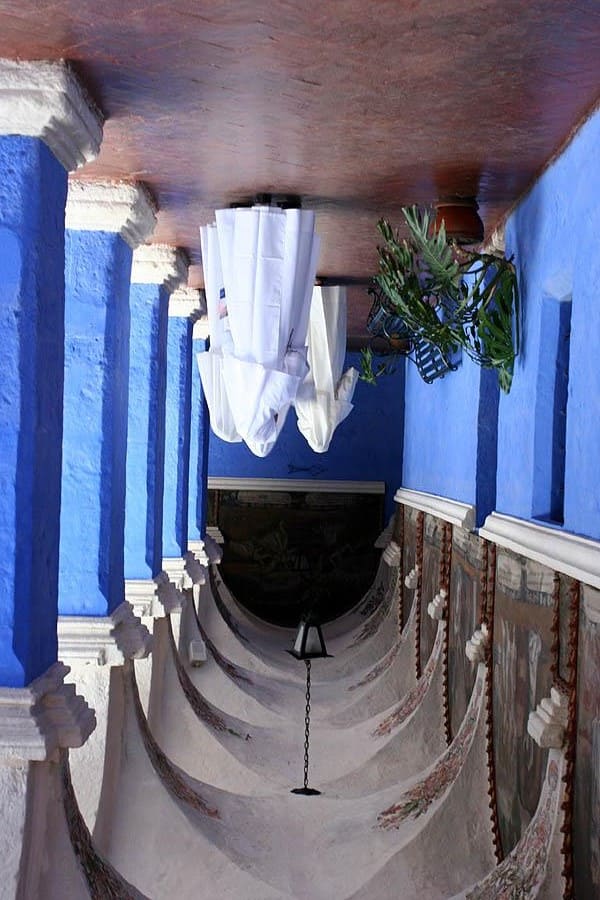}}
\end{subfigure}
\hspace{-0.4in}
\begin{subfigure}{1.2in}
{\includegraphics[width=1.2in,height=1.2in,keepaspectratio]{images/9fixed.jpg}}
\end{subfigure}
\begin{subfigure}{1.2in}
{\includegraphics[width=1.2in,height=1.2in,keepaspectratio]{images/10fixed.jpg}}
\end{subfigure}
\begin{subfigure}{1.2in}
{\includegraphics[width=1.2in,height=1.2in,keepaspectratio]{images/21fixed.jpg}}
\end{subfigure}
\caption{Predictions of the proposed method on the subset of images used to produce the results in Table \ref{Summary}. Orientation predictions by the proposed method using two different depth estimation techniques. (Depth-map) Image orientations predicted by the proposed method using complete depth-maps. (Dfd) Image orientations predicted by the proposed method using depth from defous estimates.}
\label{fig:badPredict}
\end{figure*}

\begin{figure*}[!h]
\centering
{\includegraphics[trim={40 0 20 18 },clip,width=7.75in,keepaspectratio]{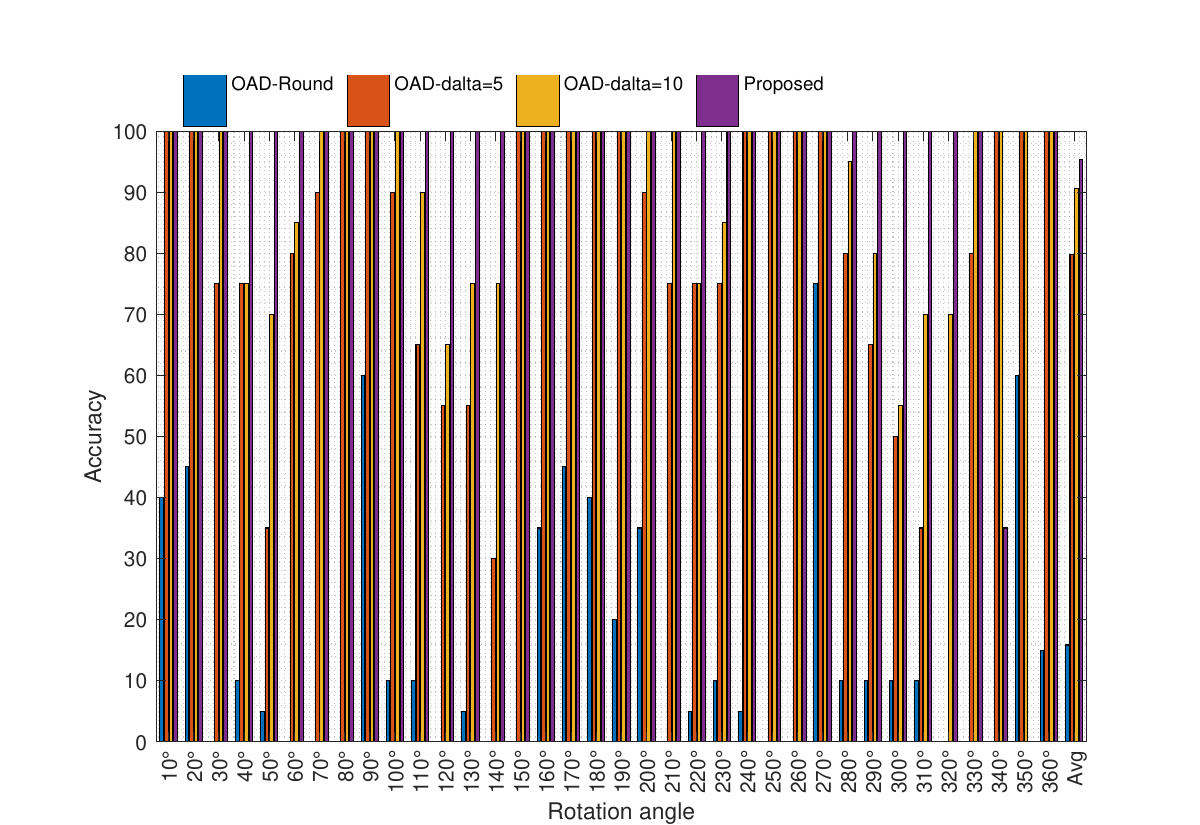}}
\caption{Comparison of the Proposed method with state-of-the-art OAD \cite{maji2020deep} on fine-scale (10\textdegree increment) image orientation estimation. Delta represents the acceptable error margin in calculating the accuracy.}
\label{fig:fineQuant}
\end{figure*}

\begin{figure*}[!h]
\centering
\begin{subfigure}{1.7in}
{\includegraphics[width=1.7in,height=1.7in,keepaspectratio]{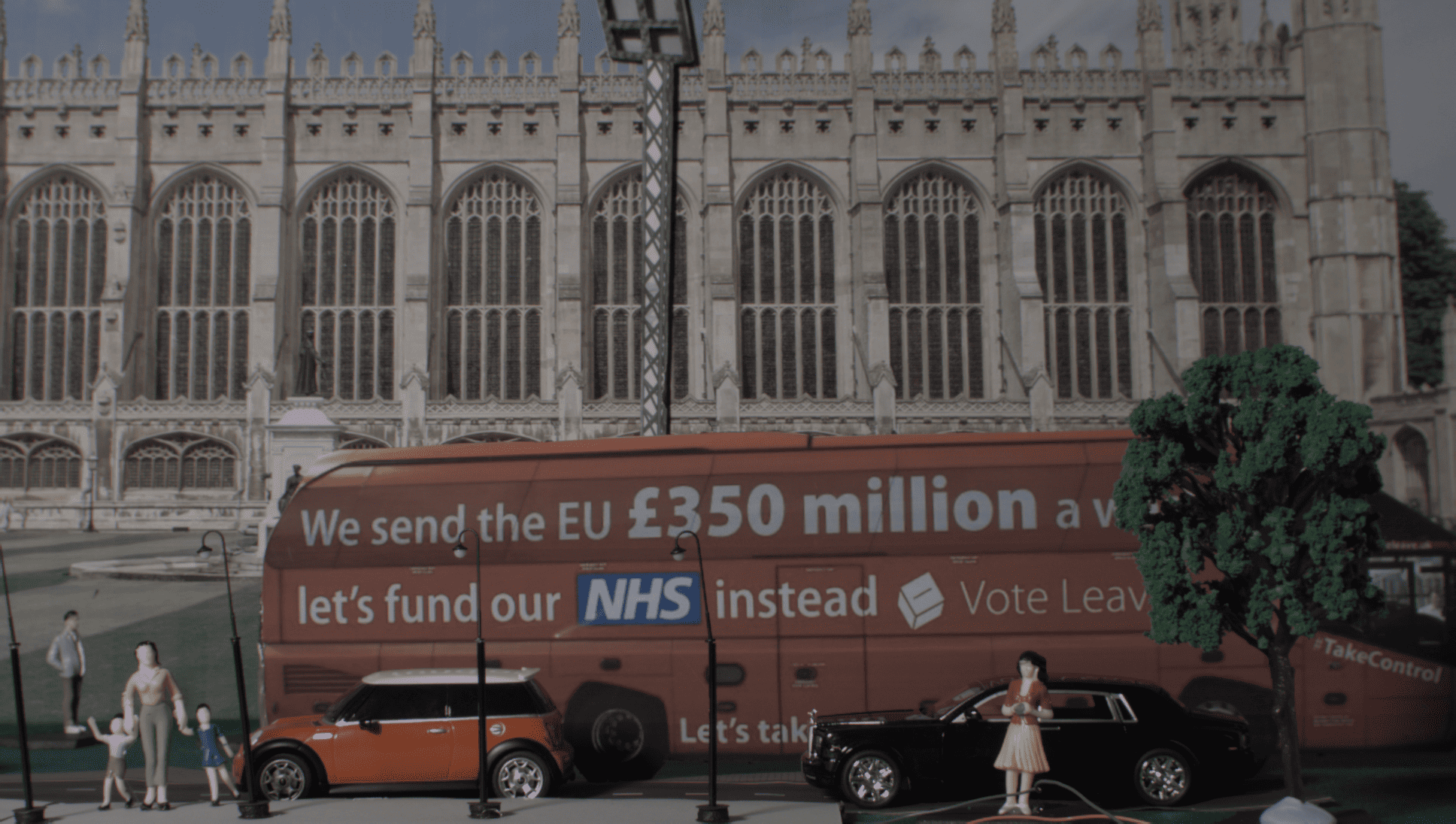}}
\end{subfigure}
\begin{subfigure}{1.7in}
{\includegraphics[width=1.7in,height=1.7in,keepaspectratio]{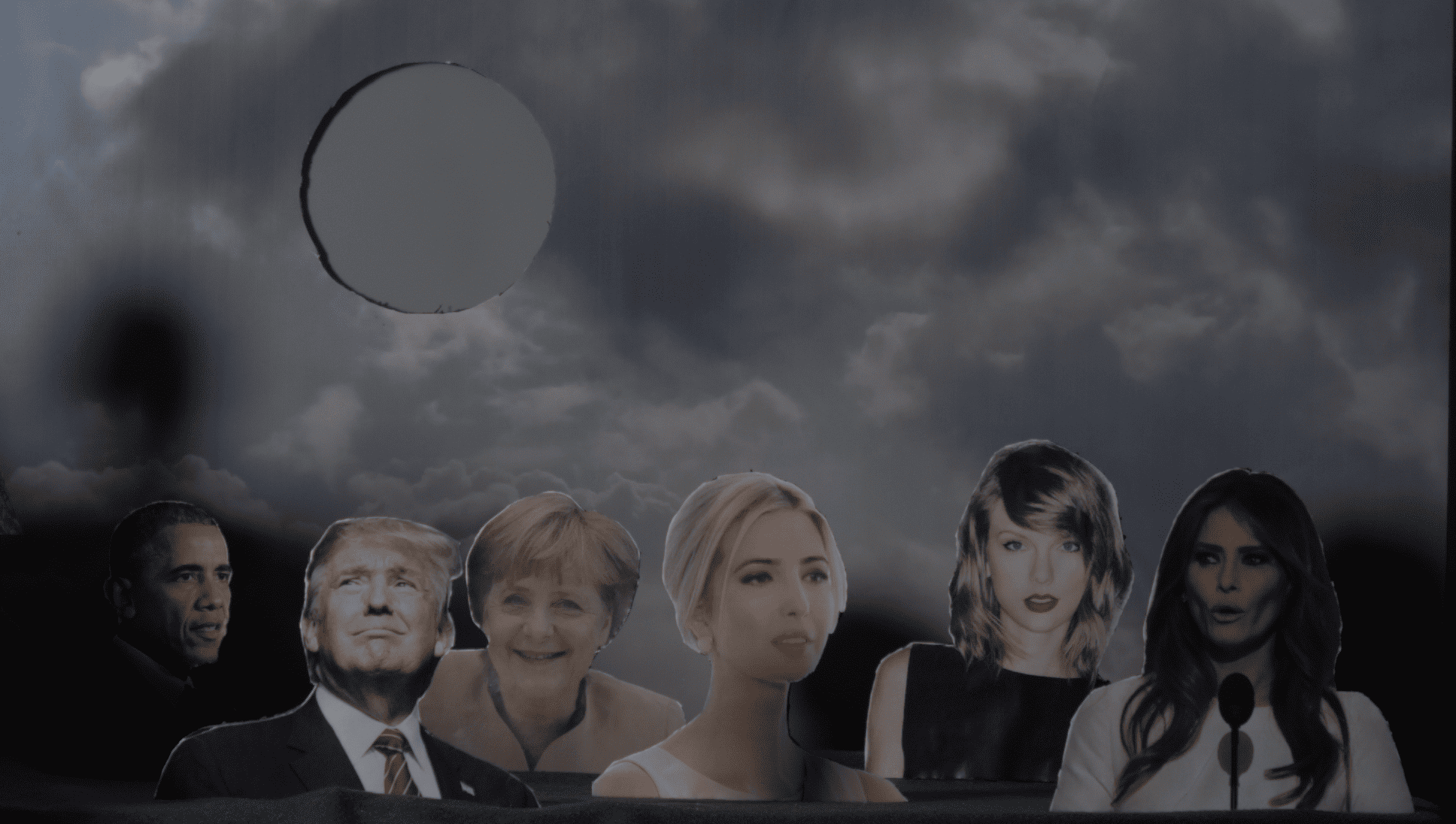}}
\end{subfigure}
\begin{subfigure}{1.7in}
{\includegraphics[width=1.7in,height=1.7in,keepaspectratio]{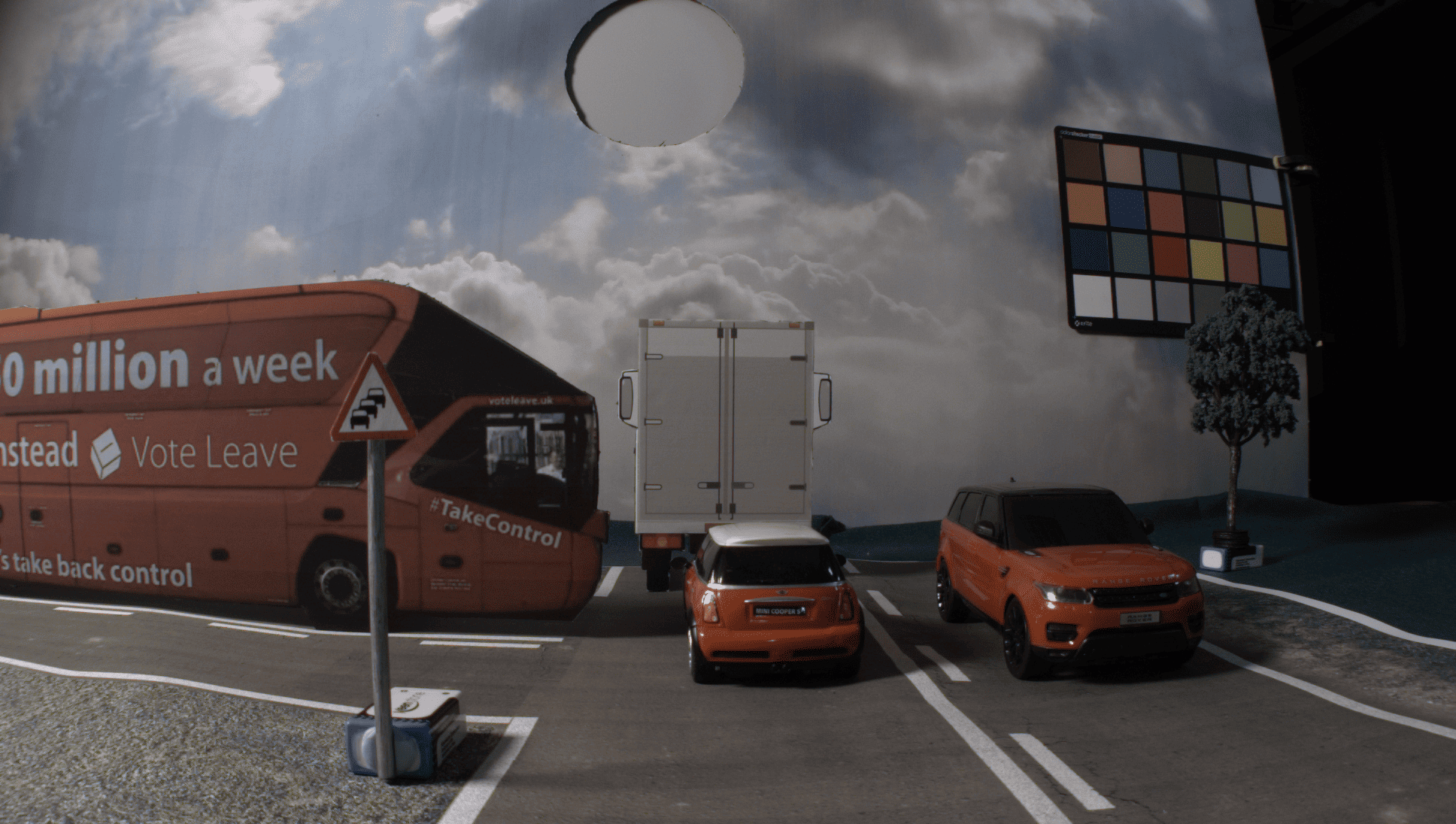}}
\end{subfigure}
\begin{subfigure}{1.7in}
{\includegraphics[width=1.7in,height=1.7in,keepaspectratio]{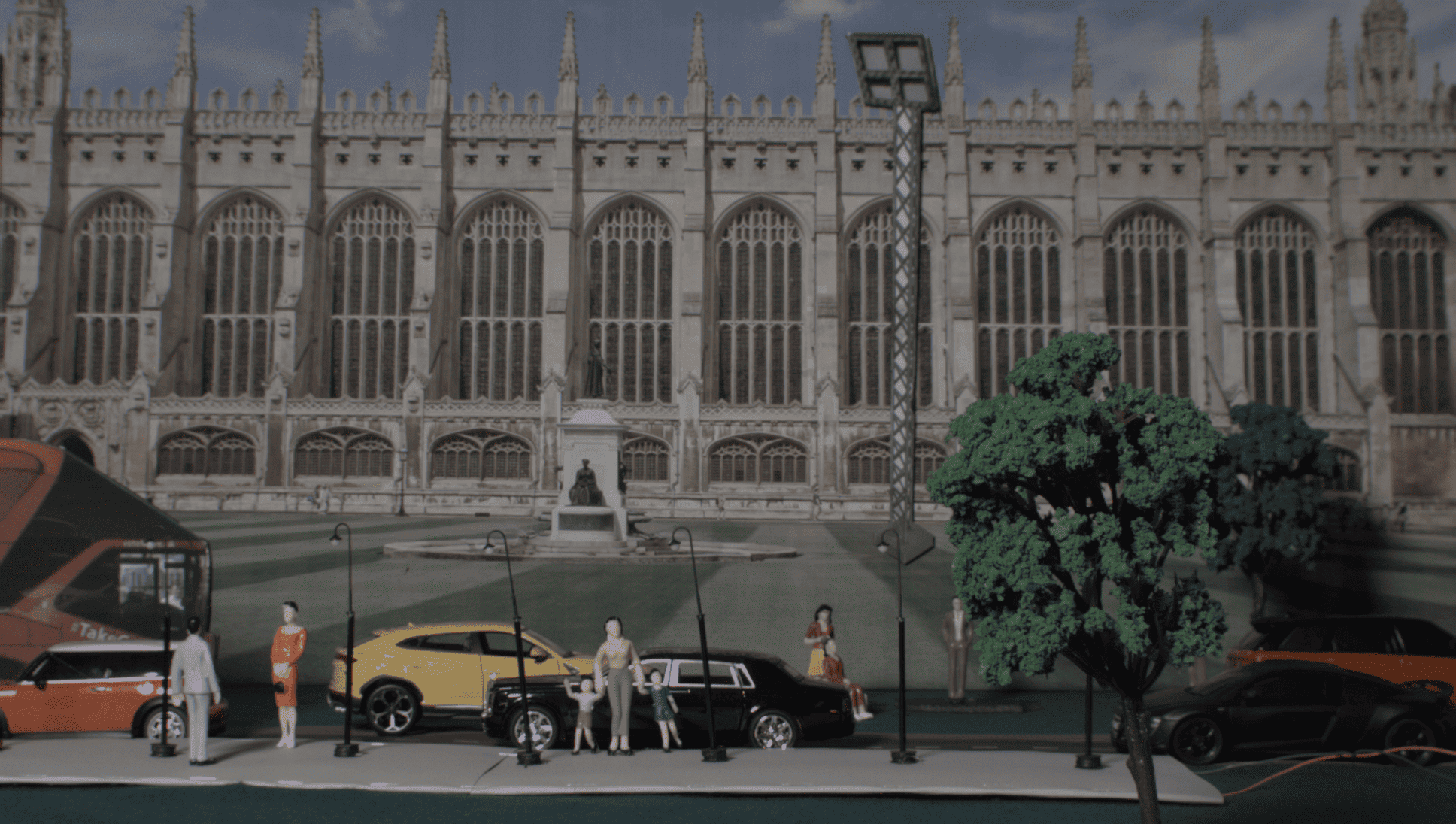}}
\end{subfigure}
\caption{A subset of images selected from four different categories to produce results in Fig. \ref{fig:fineQuant}}
\label{fig:Subset}
\end{figure*}

\begin{figure*}[!h]
\centering
\begin{subfigure}{1.5in}
{\includegraphics[width=1.5in,height=1.5in,keepaspectratio]{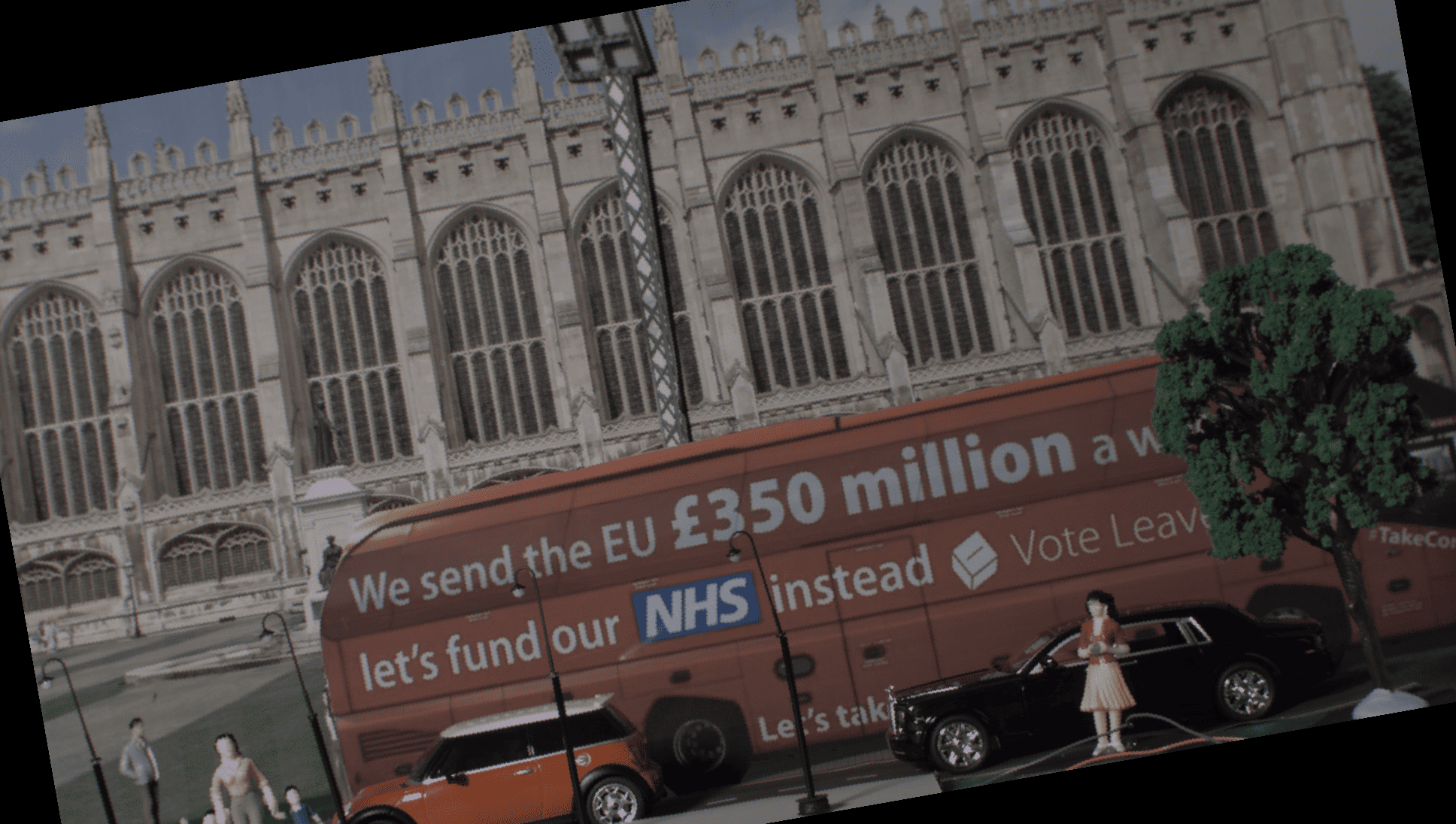}}
\end{subfigure}
\begin{subfigure}{1.5in}
{\includegraphics[width=1.5in,height=1.5in,keepaspectratio]{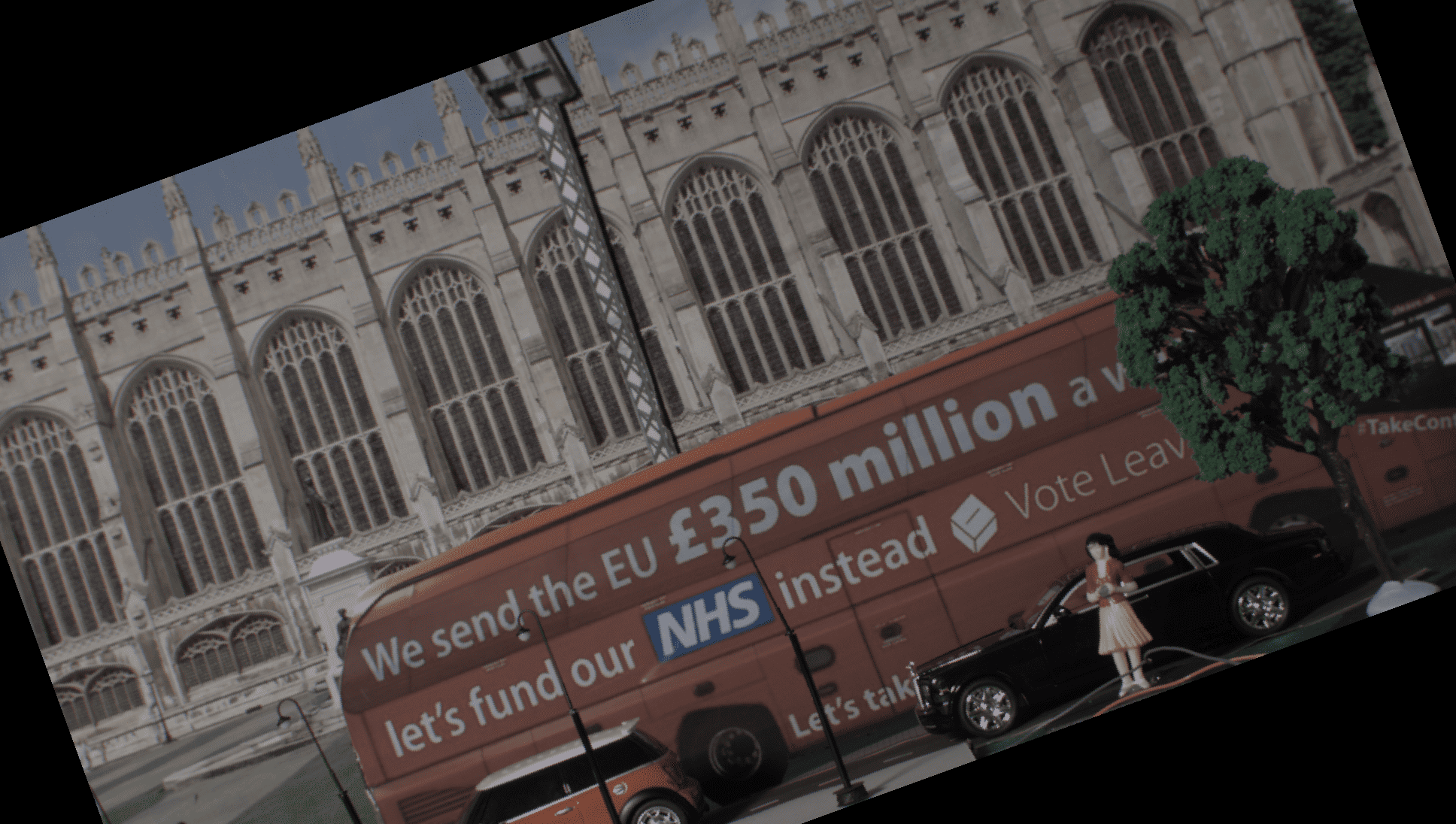}}
\end{subfigure}
\begin{subfigure}{1.5in}
{\includegraphics[width=1.5in,height=1.5in,keepaspectratio]{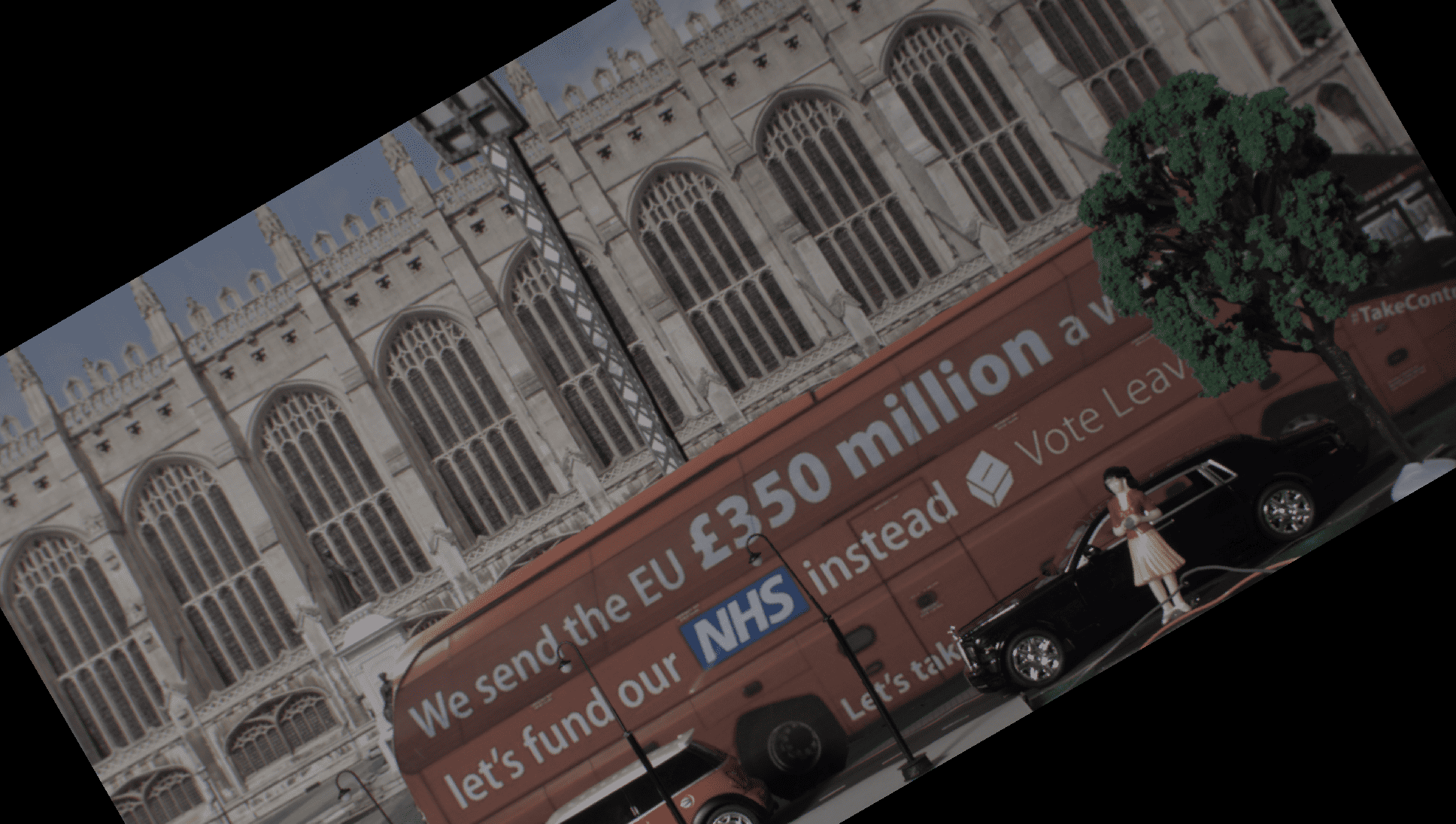}}
\end{subfigure}
\begin{subfigure}{1.5in}
{\includegraphics[width=1.5in,height=1.5in,keepaspectratio]{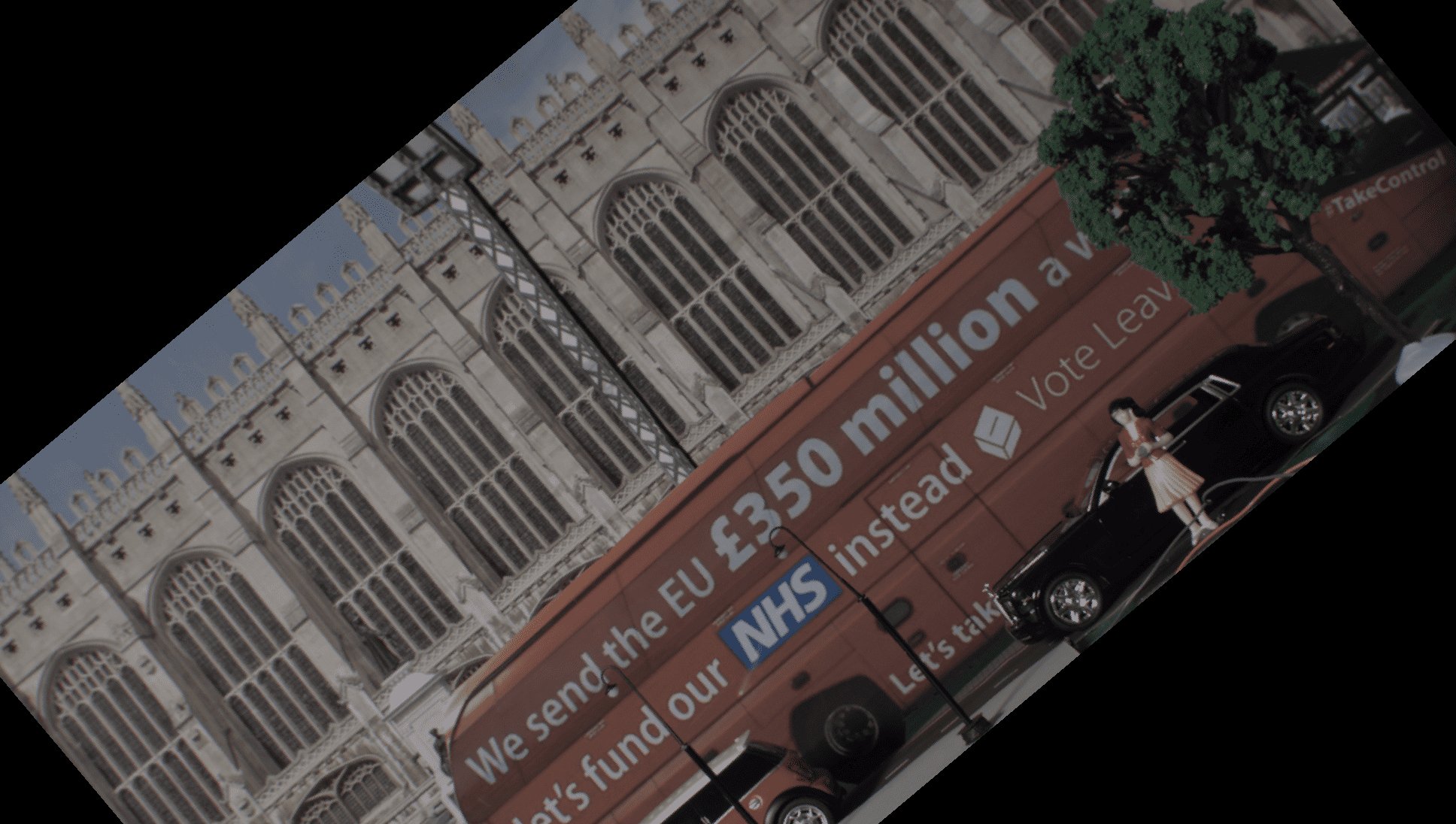}}
\end{subfigure}
\begin{subfigure}{1.5in}
{\includegraphics[width=1.5in,height=1.5in,keepaspectratio]{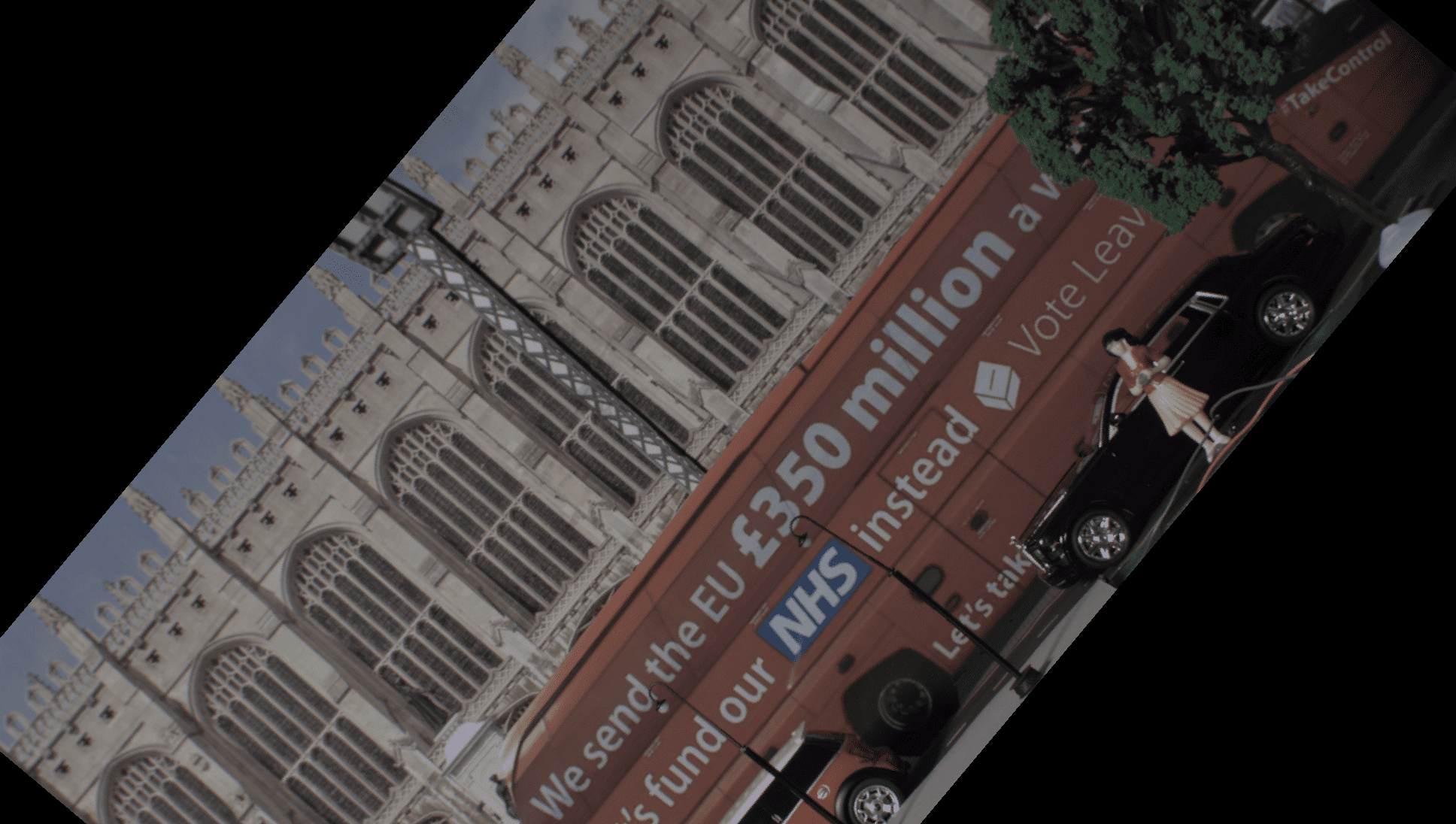}}
\end{subfigure}
\begin{subfigure}{1.5in}
{\includegraphics[width=1.5in,height=1.5in,keepaspectratio]{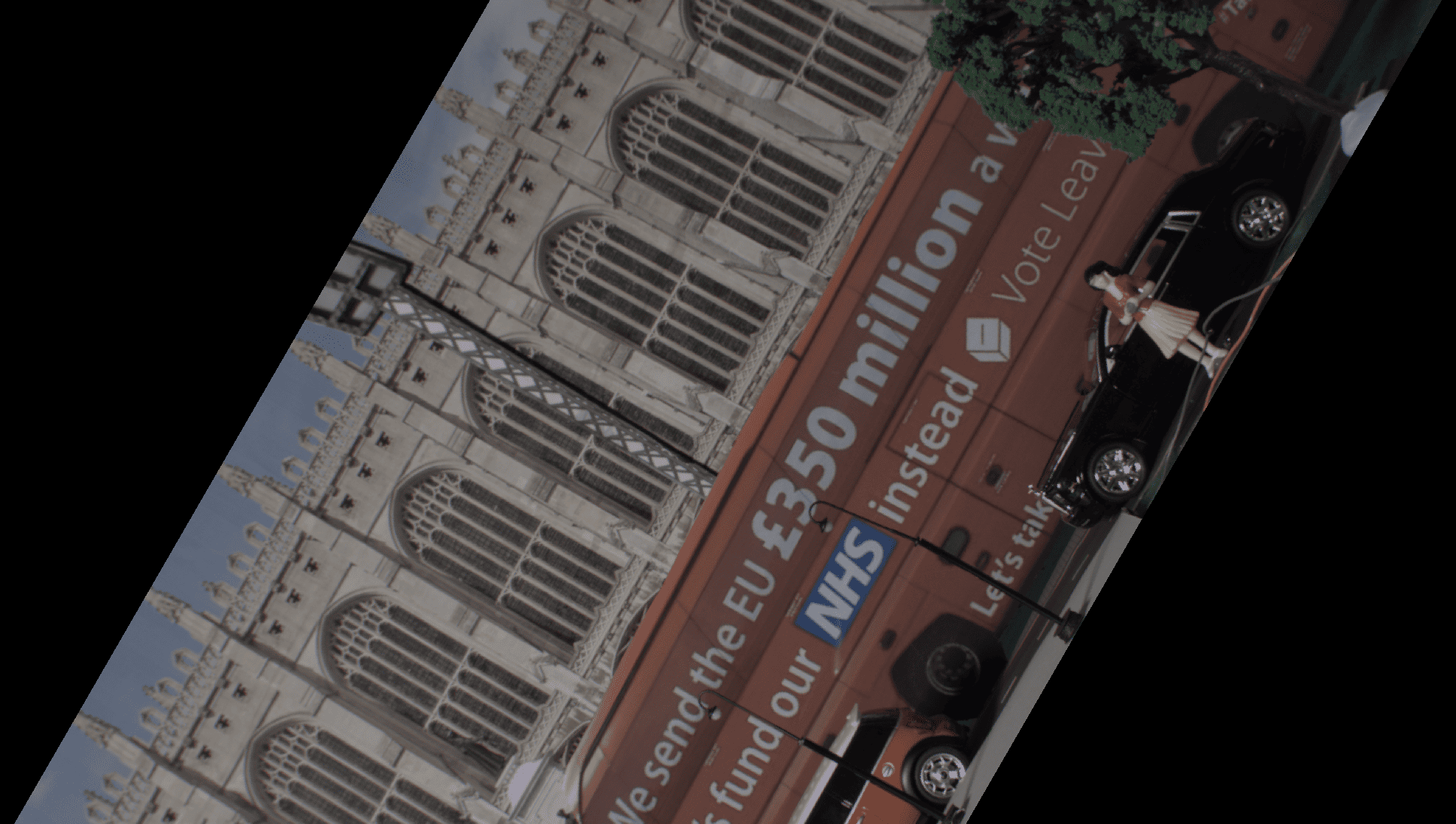}}
\end{subfigure}
\begin{subfigure}{1.5in}
{\includegraphics[width=1.5in,height=1.5in,keepaspectratio]{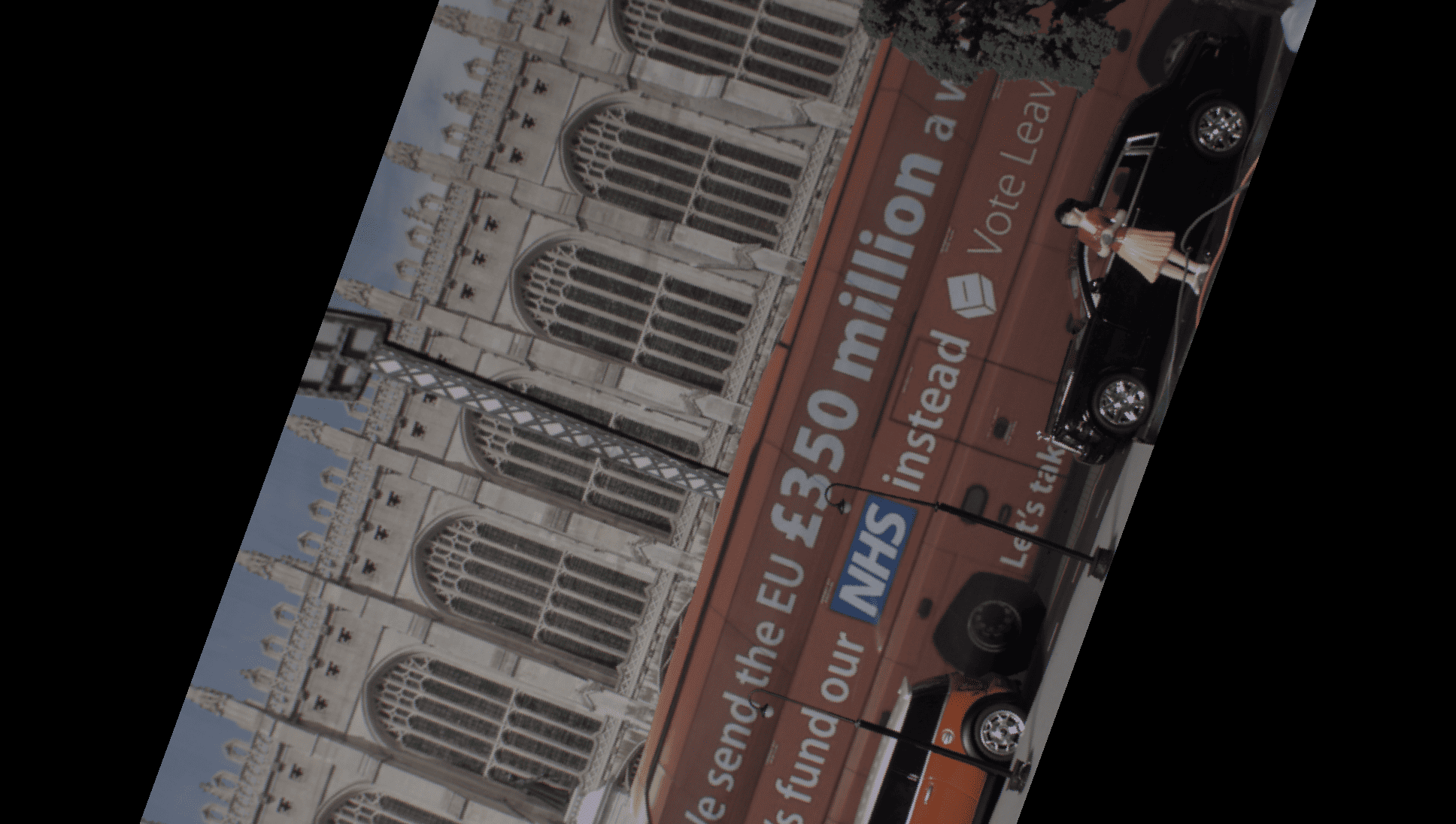}}
\end{subfigure}
\begin{subfigure}{1.5in}
{\includegraphics[width=1.5in,height=1.5in,keepaspectratio]{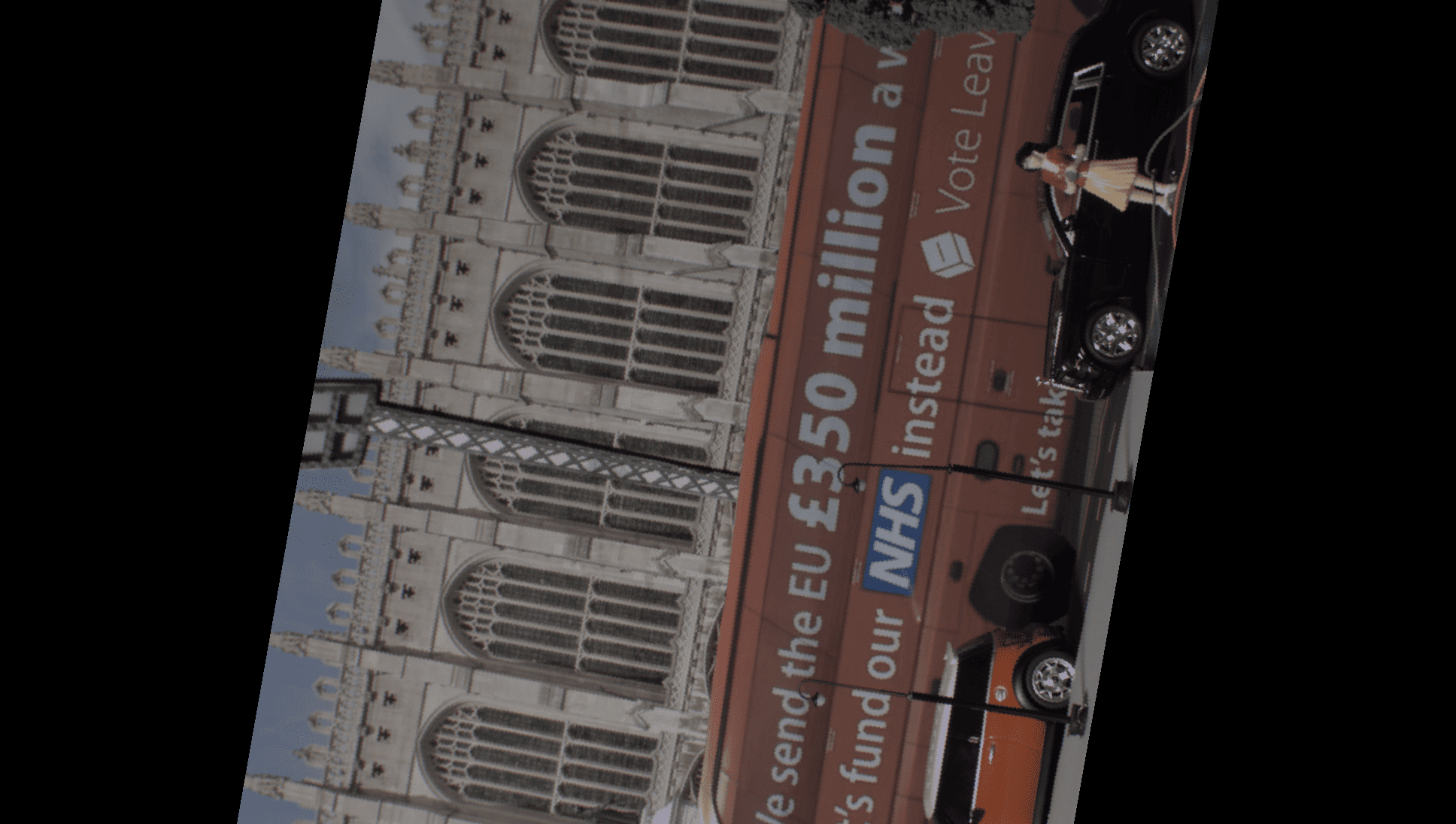}}
\end{subfigure}
\begin{subfigure}{1.5in}
{\includegraphics[width=1.5in,height=1.5in,keepaspectratio]{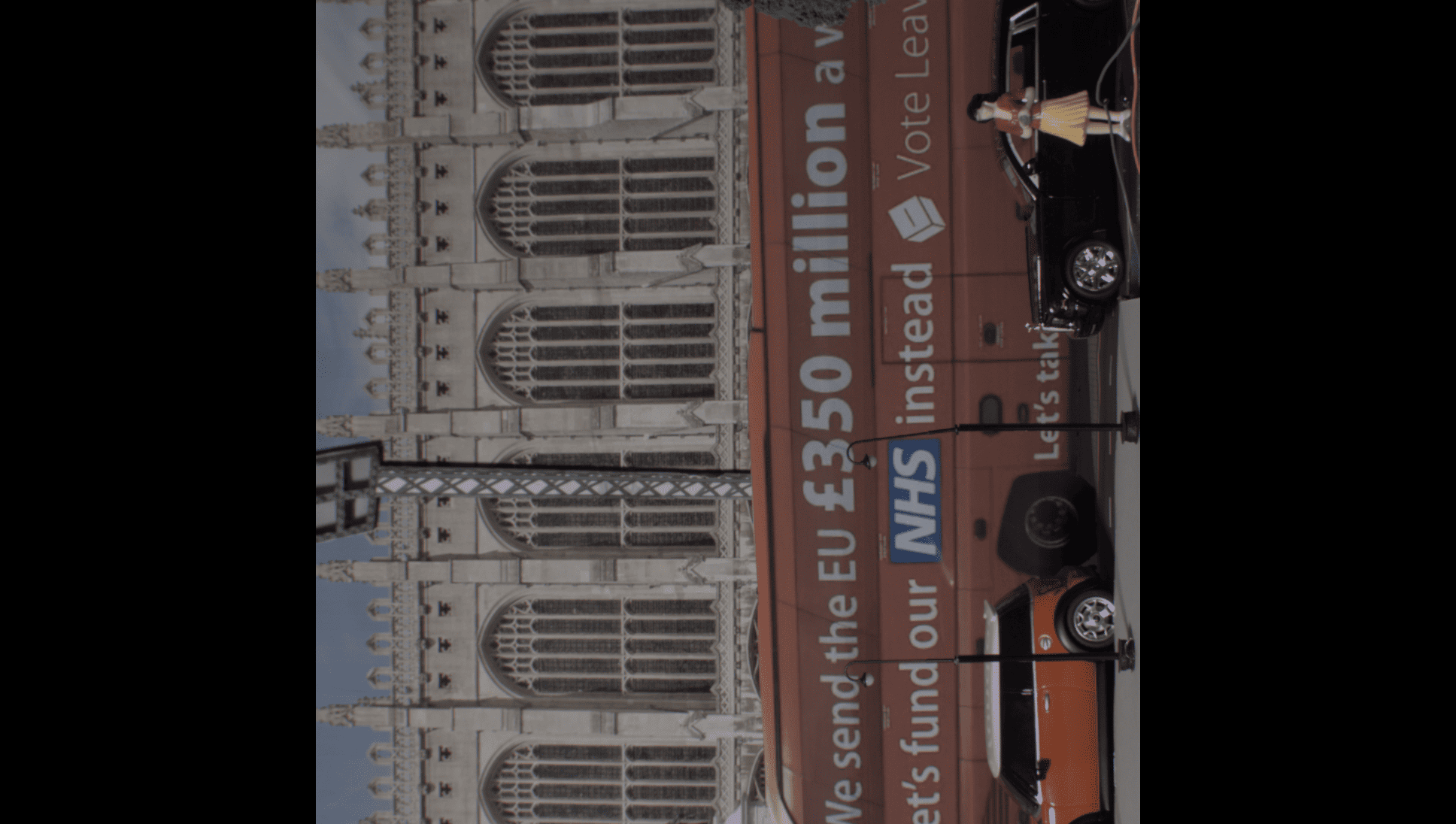}}
\end{subfigure}
\begin{subfigure}{1.5in}
{\includegraphics[width=1.5in,height=1.5in,keepaspectratio]{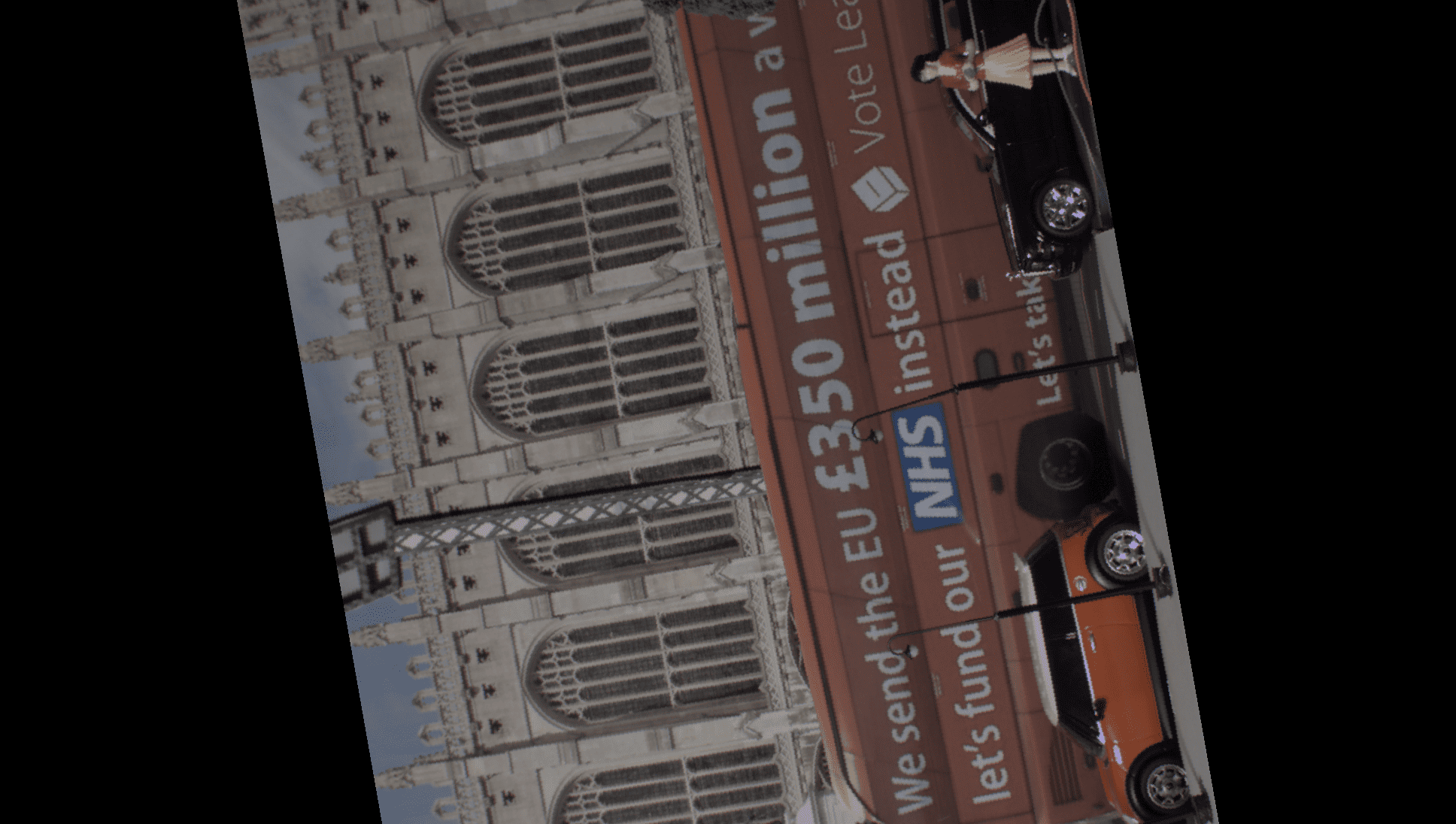}}
\end{subfigure}
\begin{subfigure}{1.5in}
{\includegraphics[width=1.5in,height=1.5in,keepaspectratio]{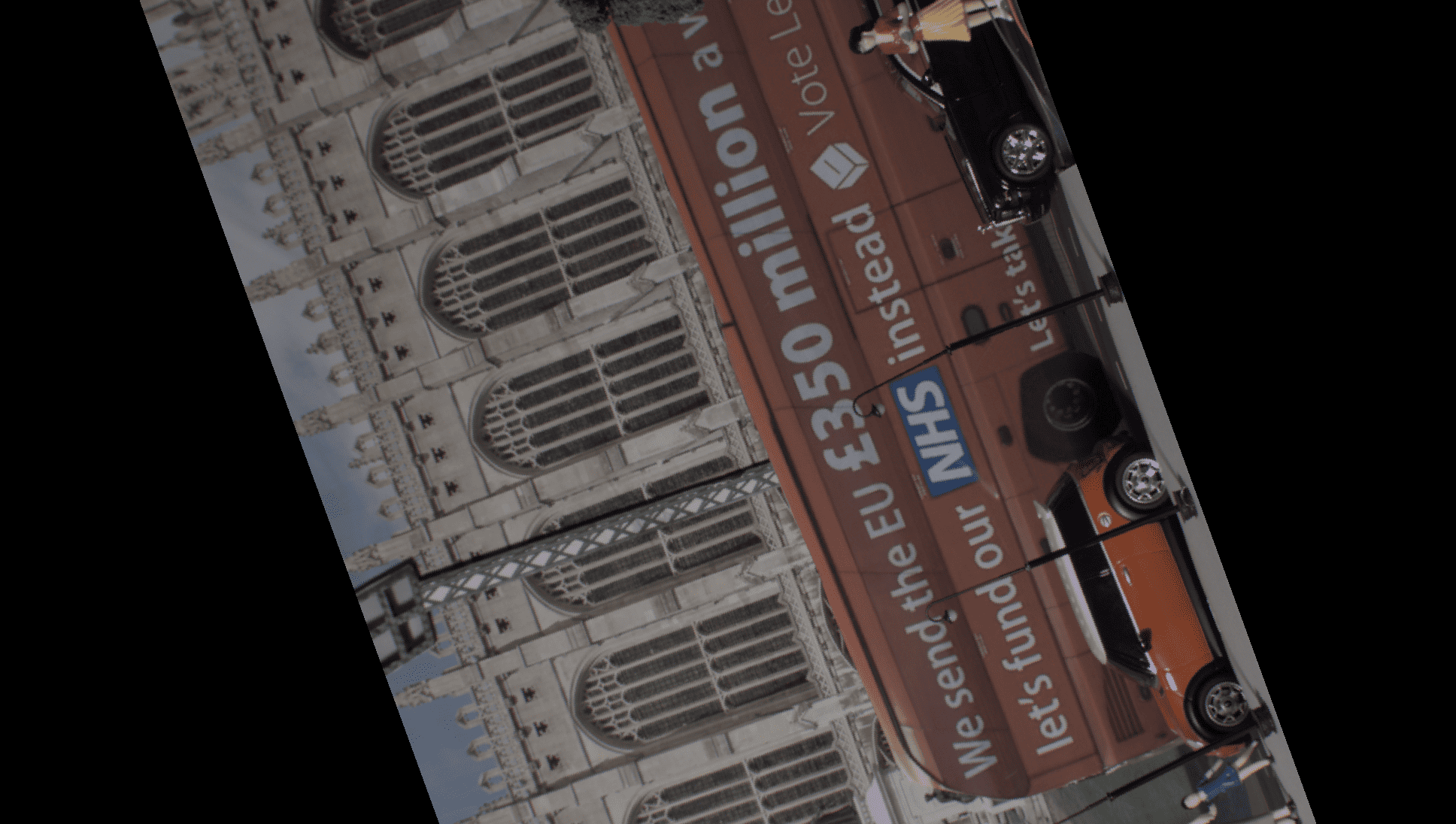}}
\end{subfigure}
\begin{subfigure}{1.5in}
{\includegraphics[width=1.5in,height=1.5in,keepaspectratio]{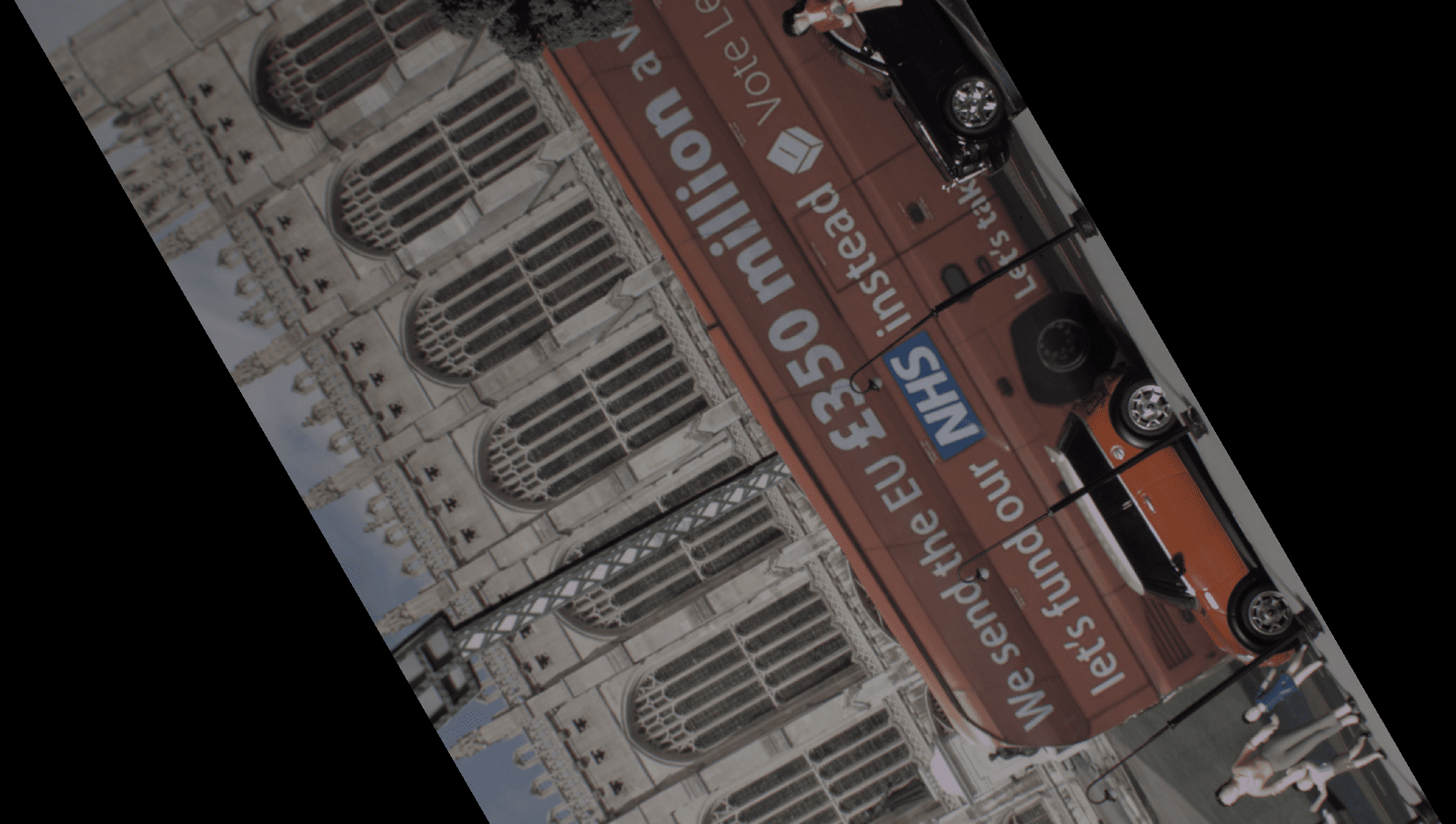}}
\end{subfigure}\\
\begin{subfigure}{1.5in}
{\includegraphics[width=1.5in,height=1.5in,keepaspectratio]{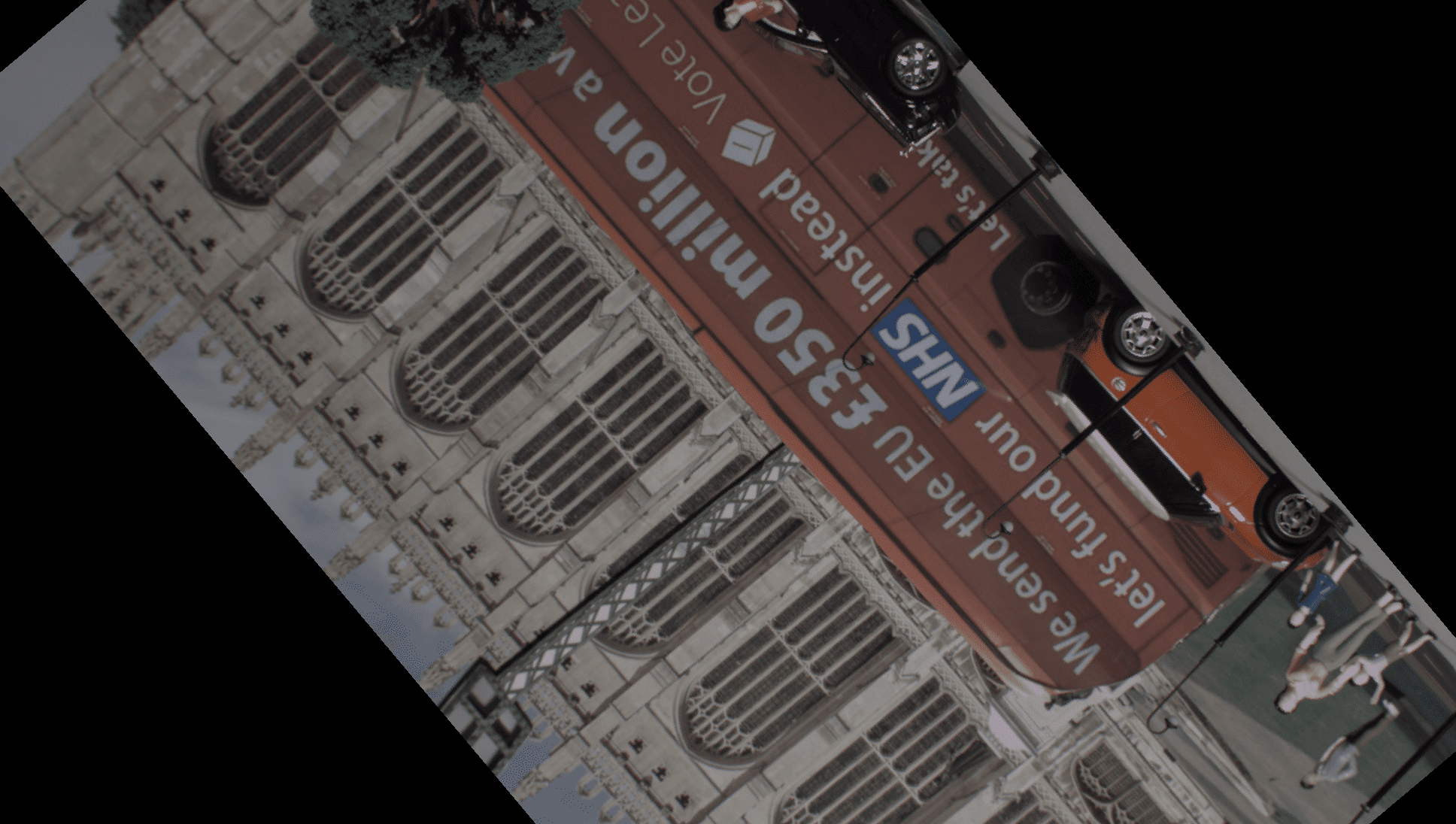}}
\end{subfigure}
\begin{subfigure}{1.5in}
{\includegraphics[width=1.5in,height=1.5in,keepaspectratio]{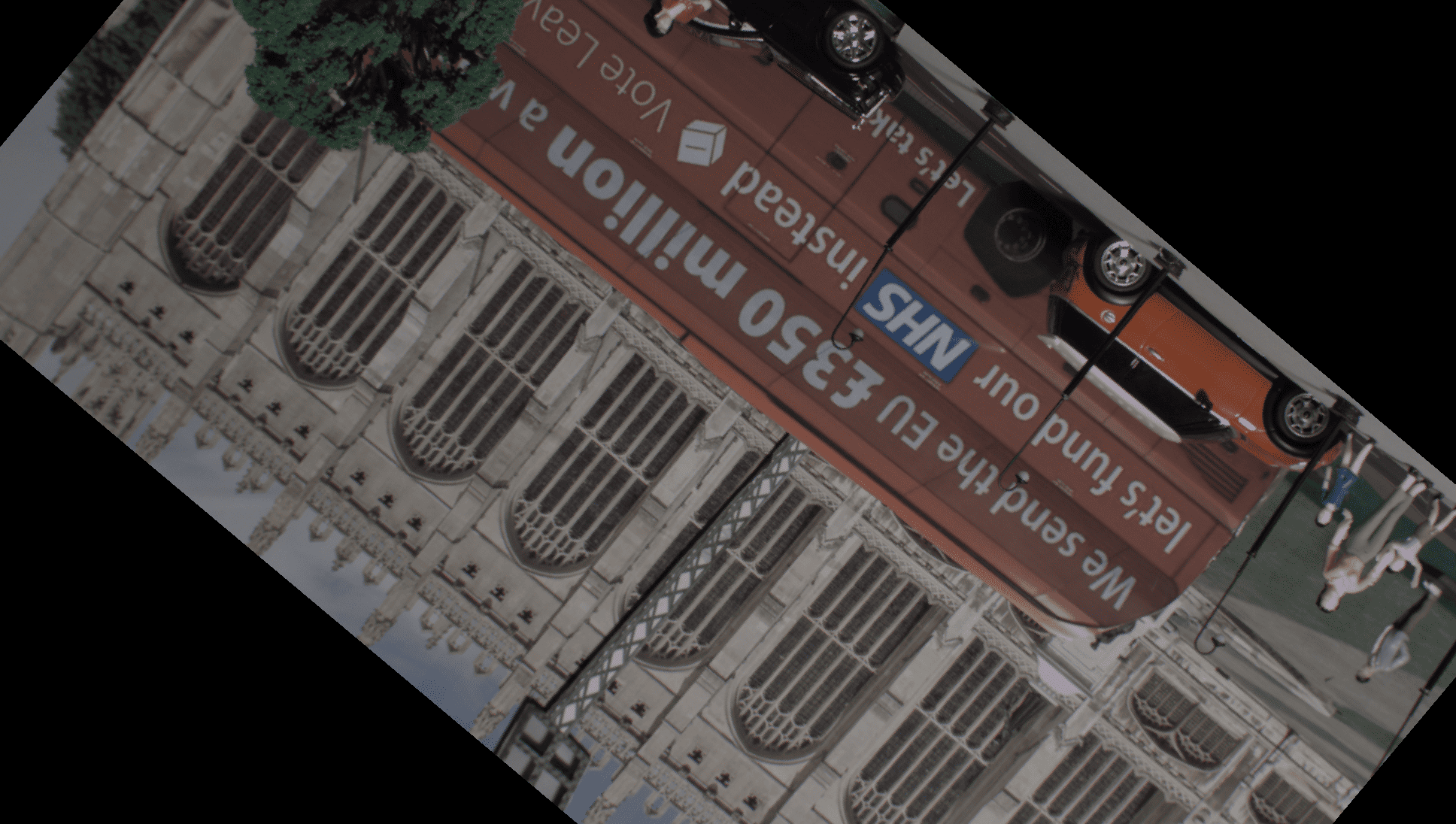}}
\end{subfigure}
\begin{subfigure}{1.5in}
{\includegraphics[width=1.5in,height=1.5in,keepaspectratio]{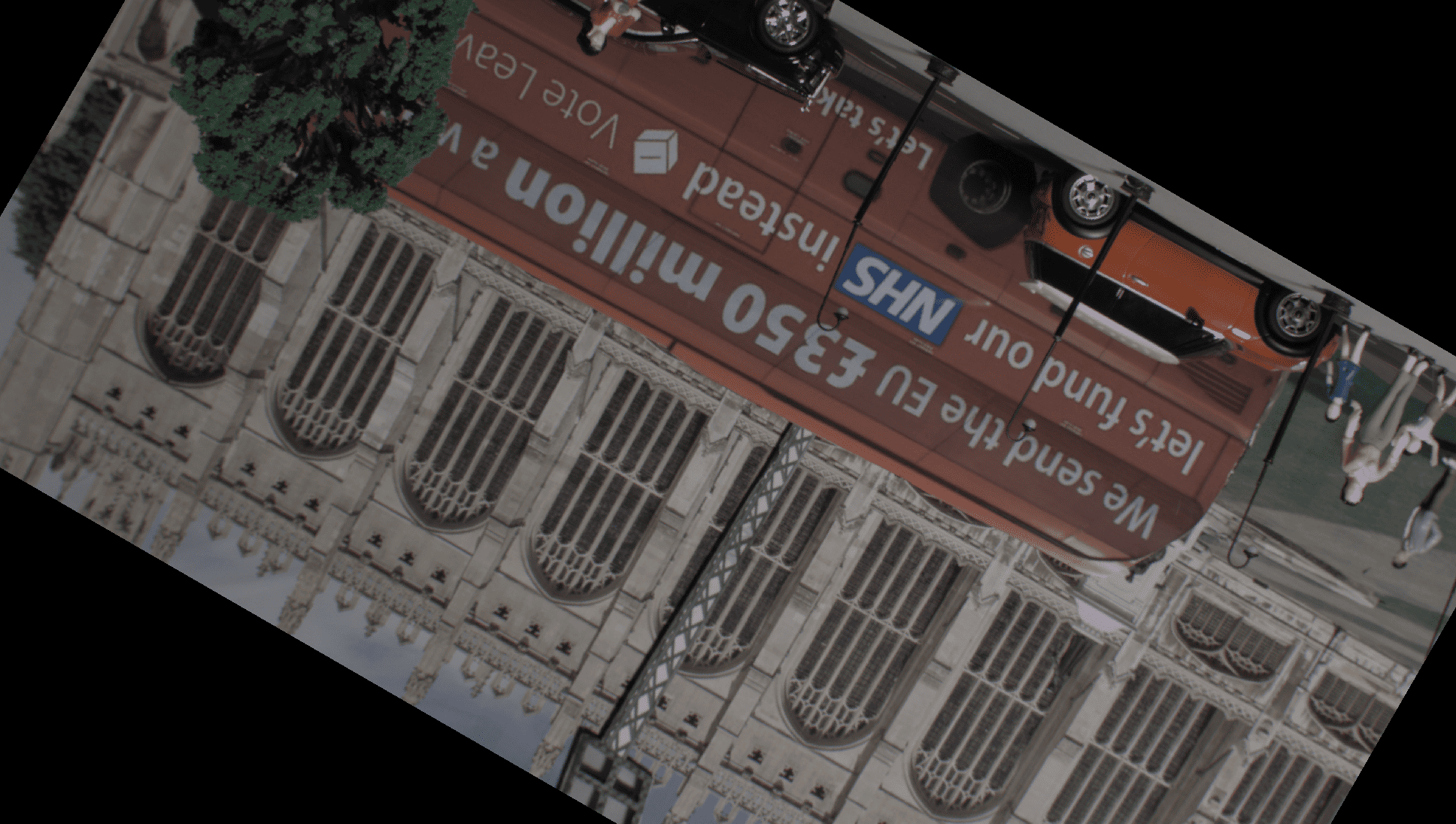}}
\end{subfigure}
\begin{subfigure}{1.5in}
{\includegraphics[width=1.5in,height=1.5in,keepaspectratio]{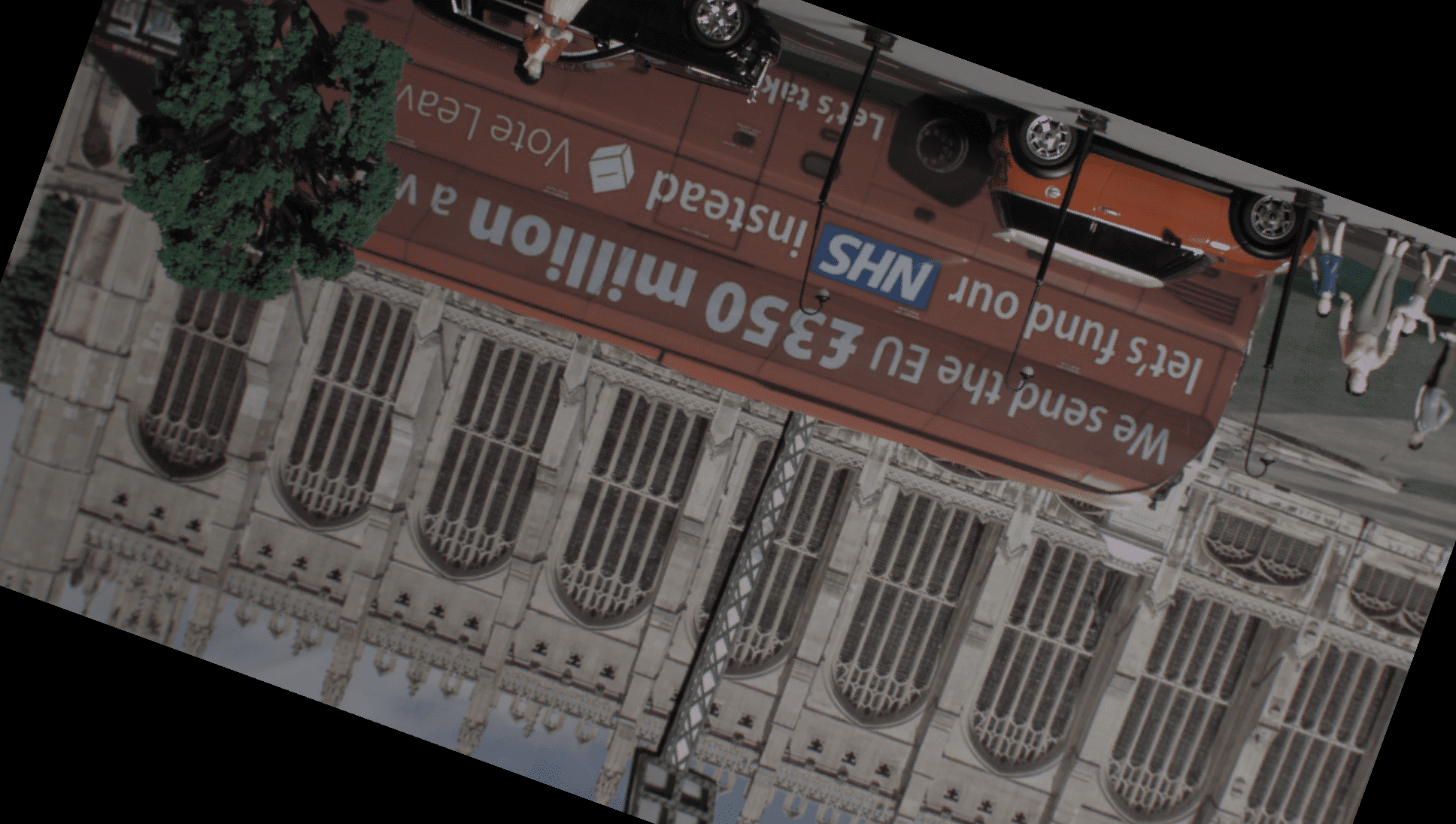}}
\end{subfigure}
\begin{subfigure}{1.5in}
{\includegraphics[width=1.5in,height=1.5in,keepaspectratio]{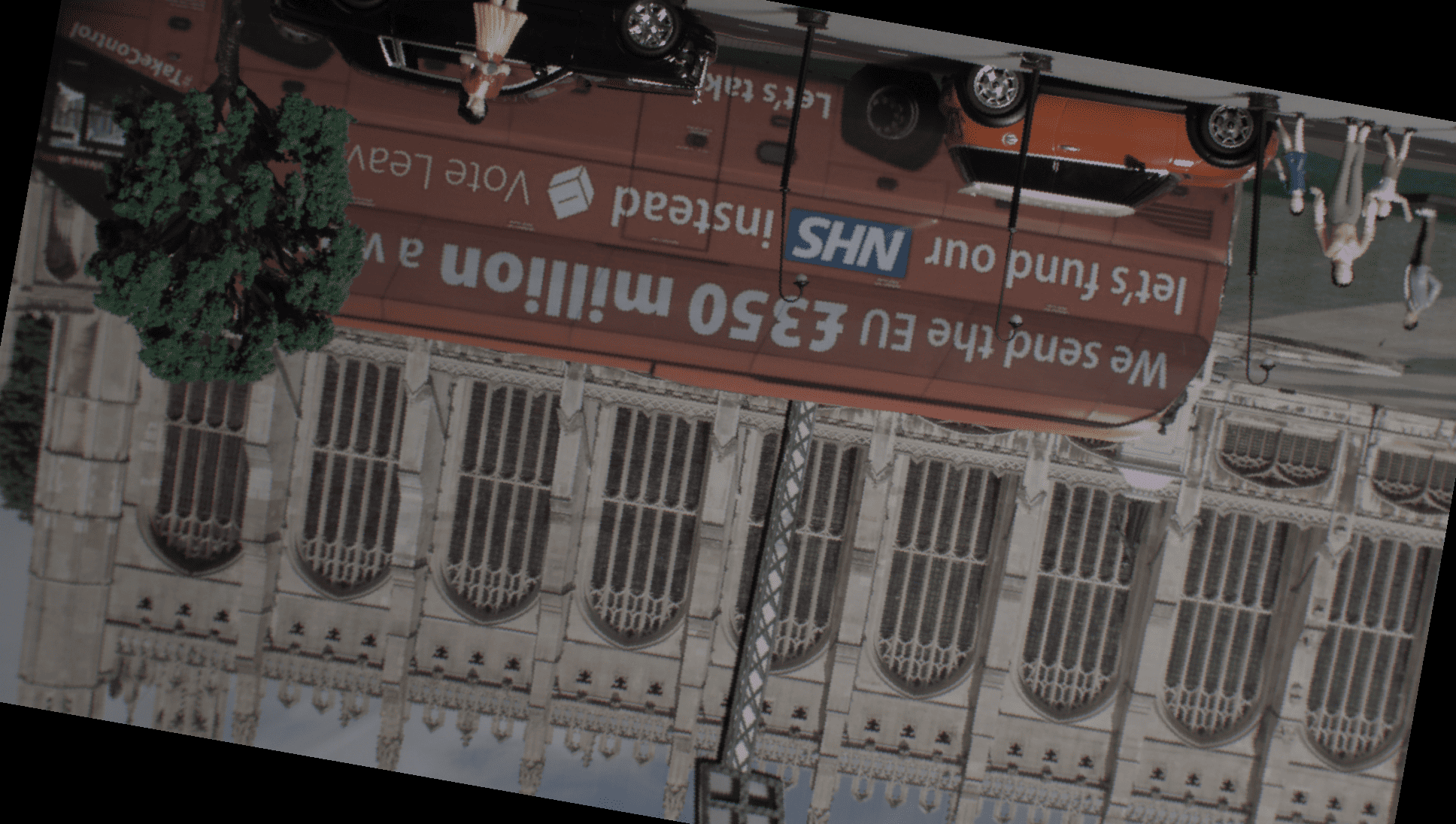}}
\end{subfigure}
\begin{subfigure}{1.5in}
{\includegraphics[width=1.5in,height=1.5in,keepaspectratio]{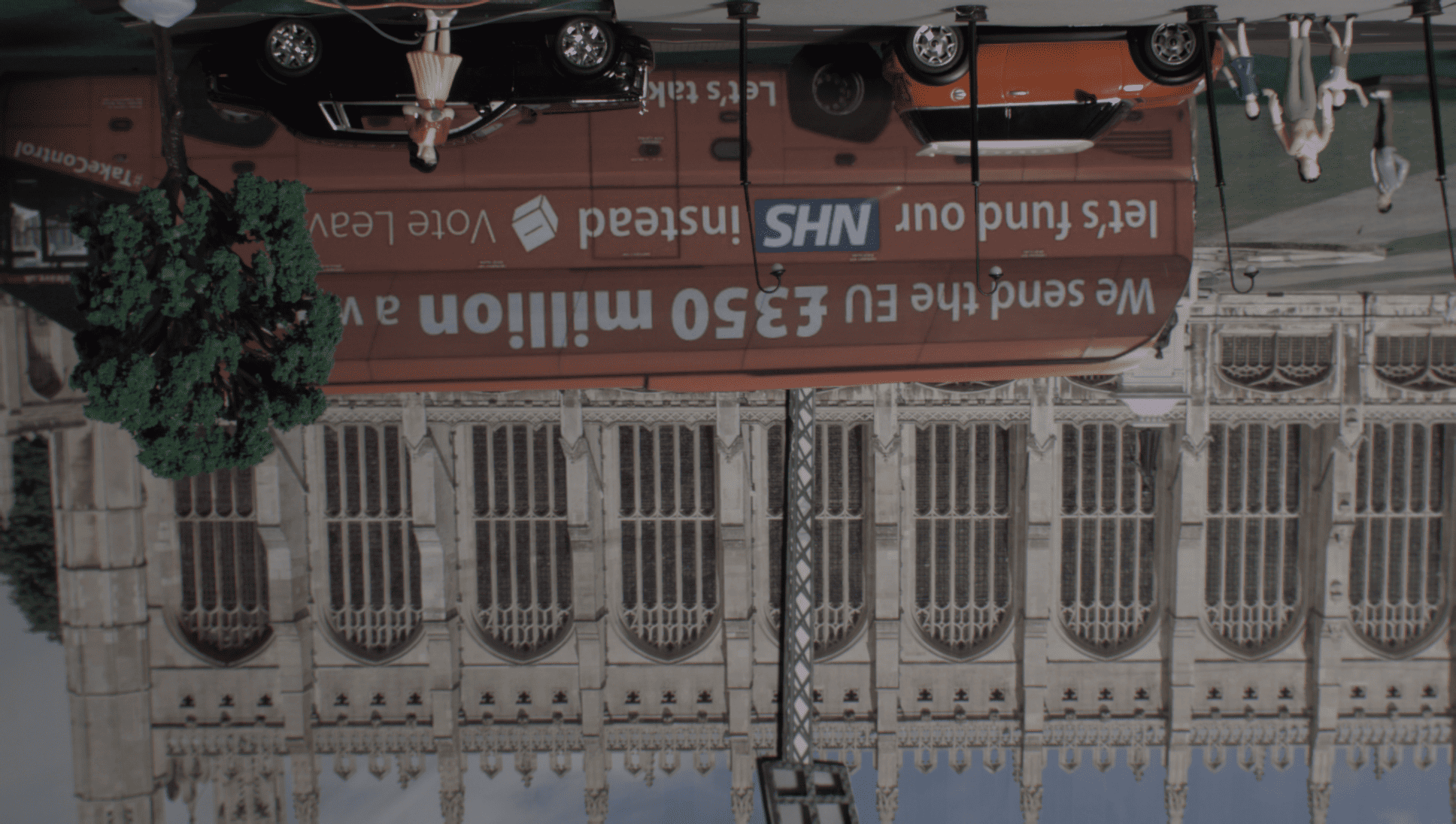}}
\end{subfigure}
\begin{subfigure}{1.5in}
{\includegraphics[width=1.5in,height=1.5in,keepaspectratio]{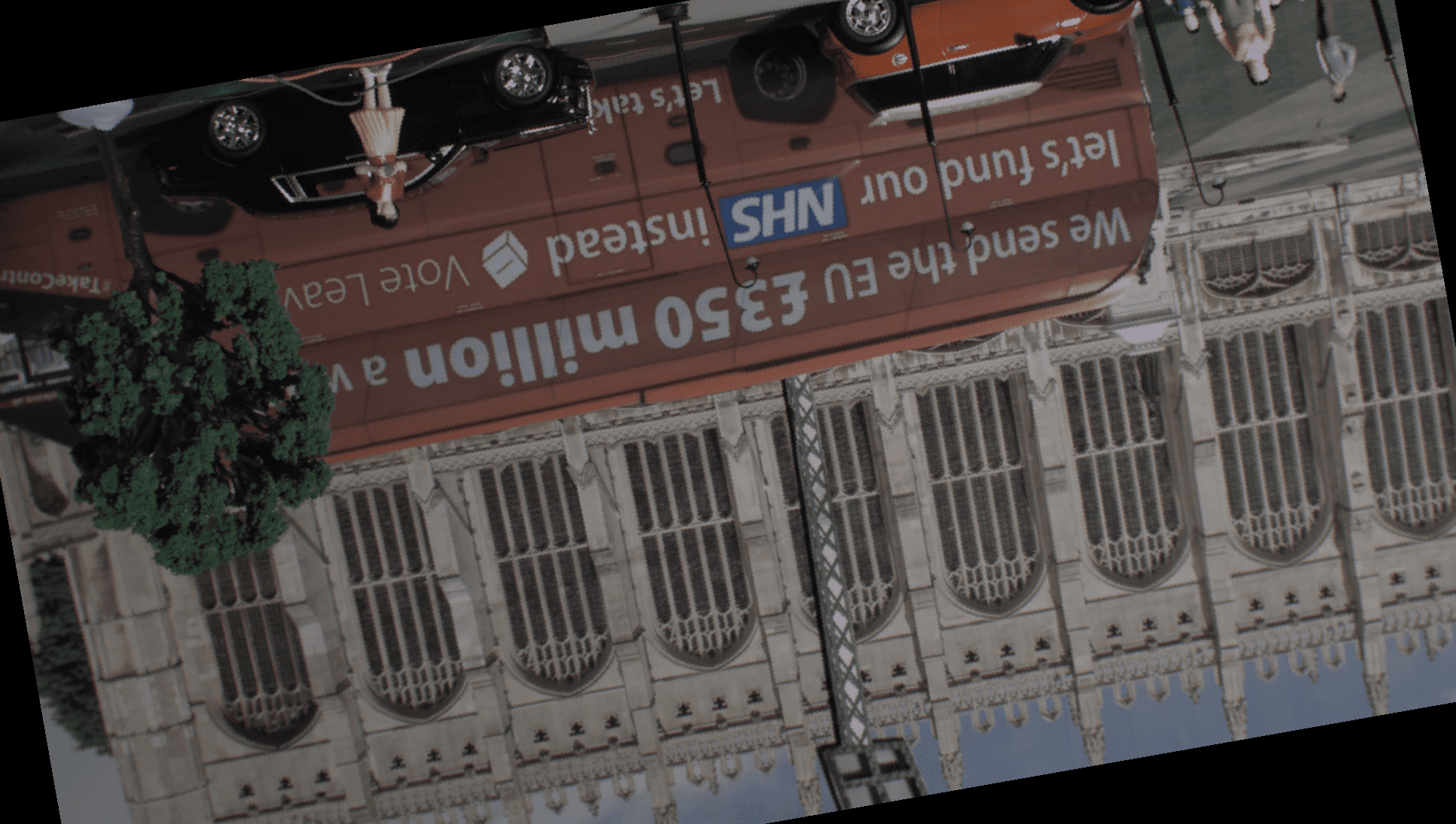}}
\end{subfigure}
\begin{subfigure}{1.5in}
{\includegraphics[width=1.5in,height=1.5in,keepaspectratio]{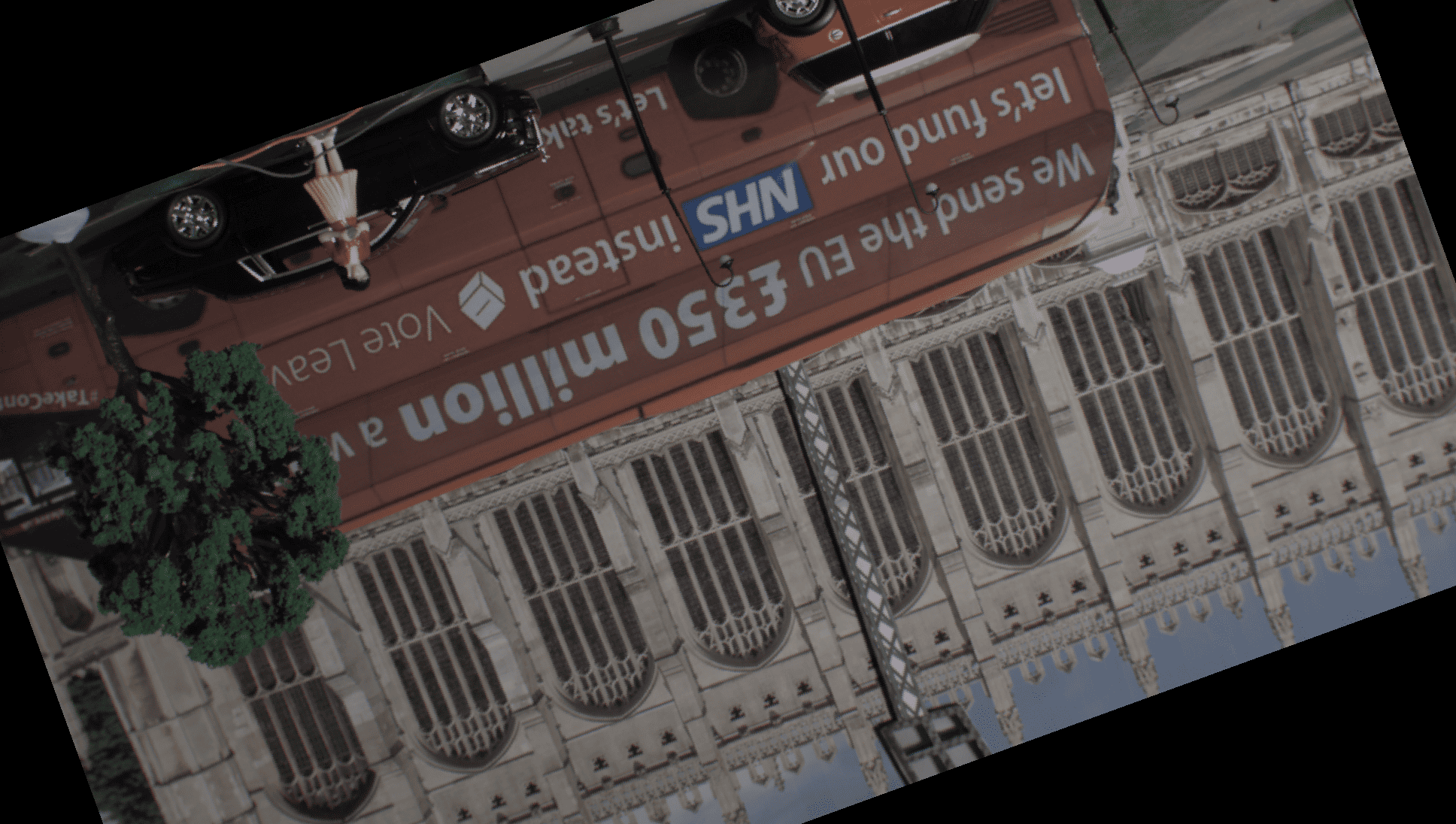}}
\end{subfigure}
\begin{subfigure}{1.5in}
{\includegraphics[width=1.5in,height=1.5in,keepaspectratio]{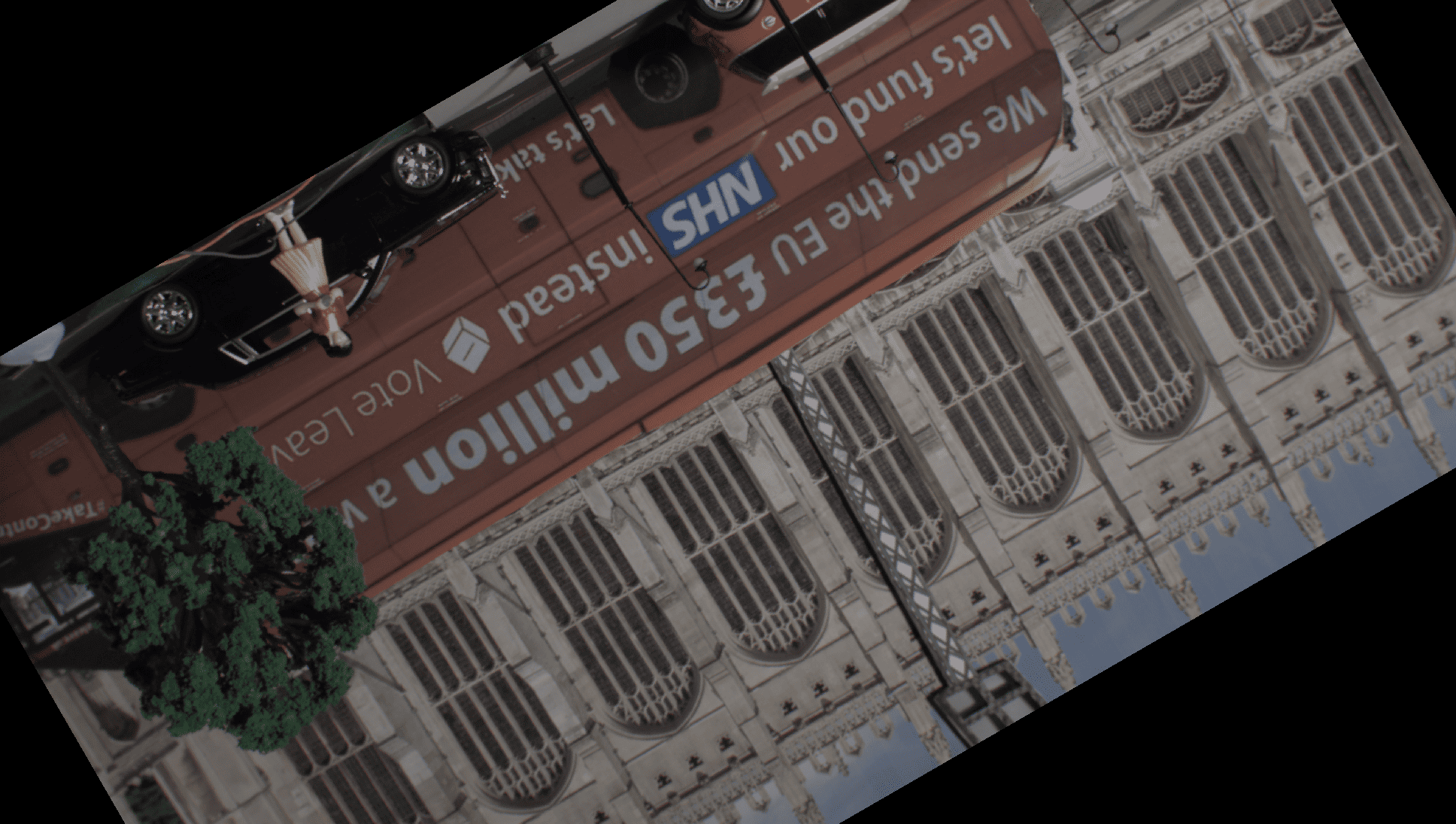}}
\end{subfigure}
\begin{subfigure}{1.5in}
{\includegraphics[width=1.5in,height=1.5in,keepaspectratio]{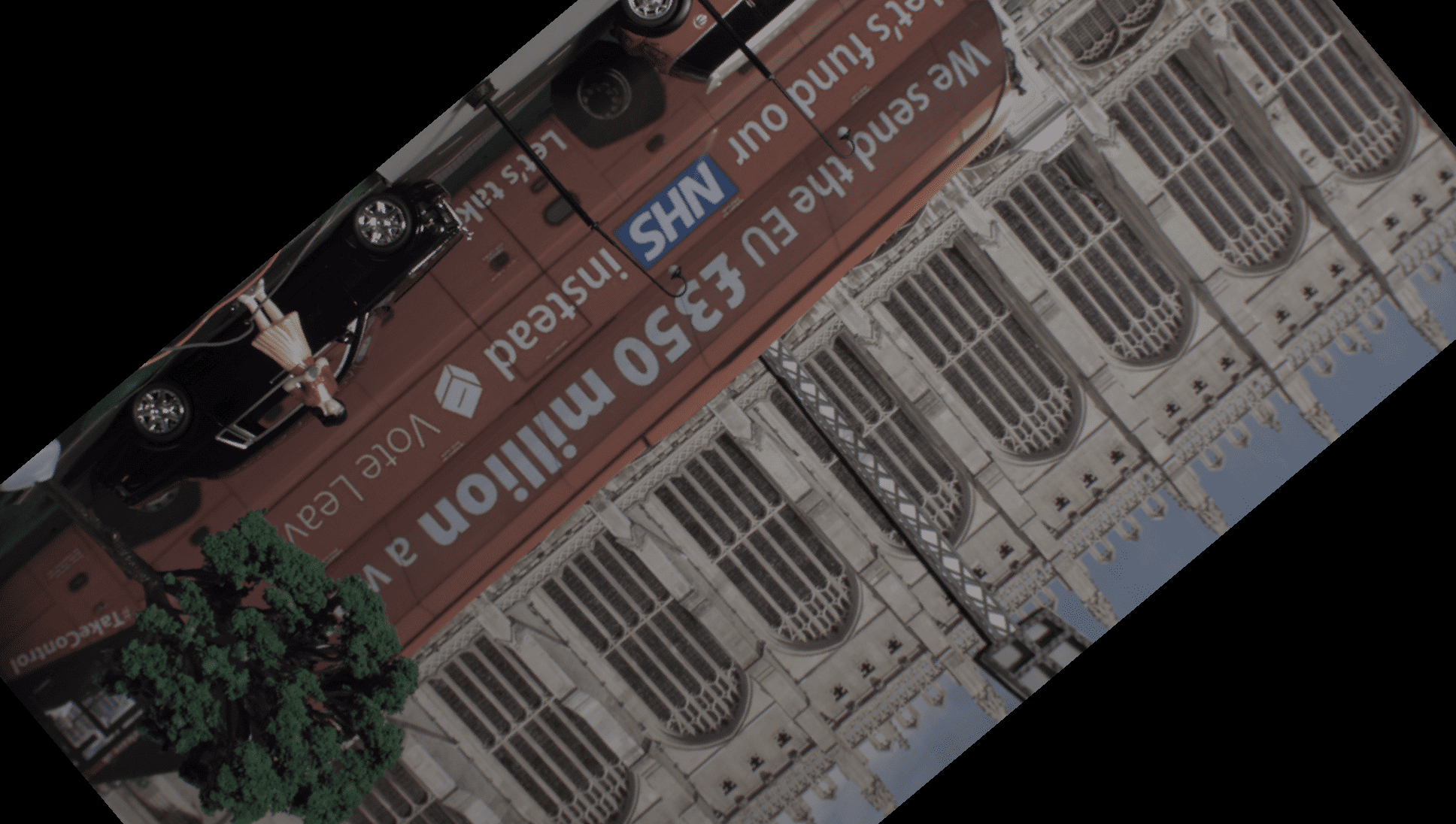}}
\end{subfigure}
\begin{subfigure}{1.5in}
{\includegraphics[width=1.5in,height=1.5in,keepaspectratio]{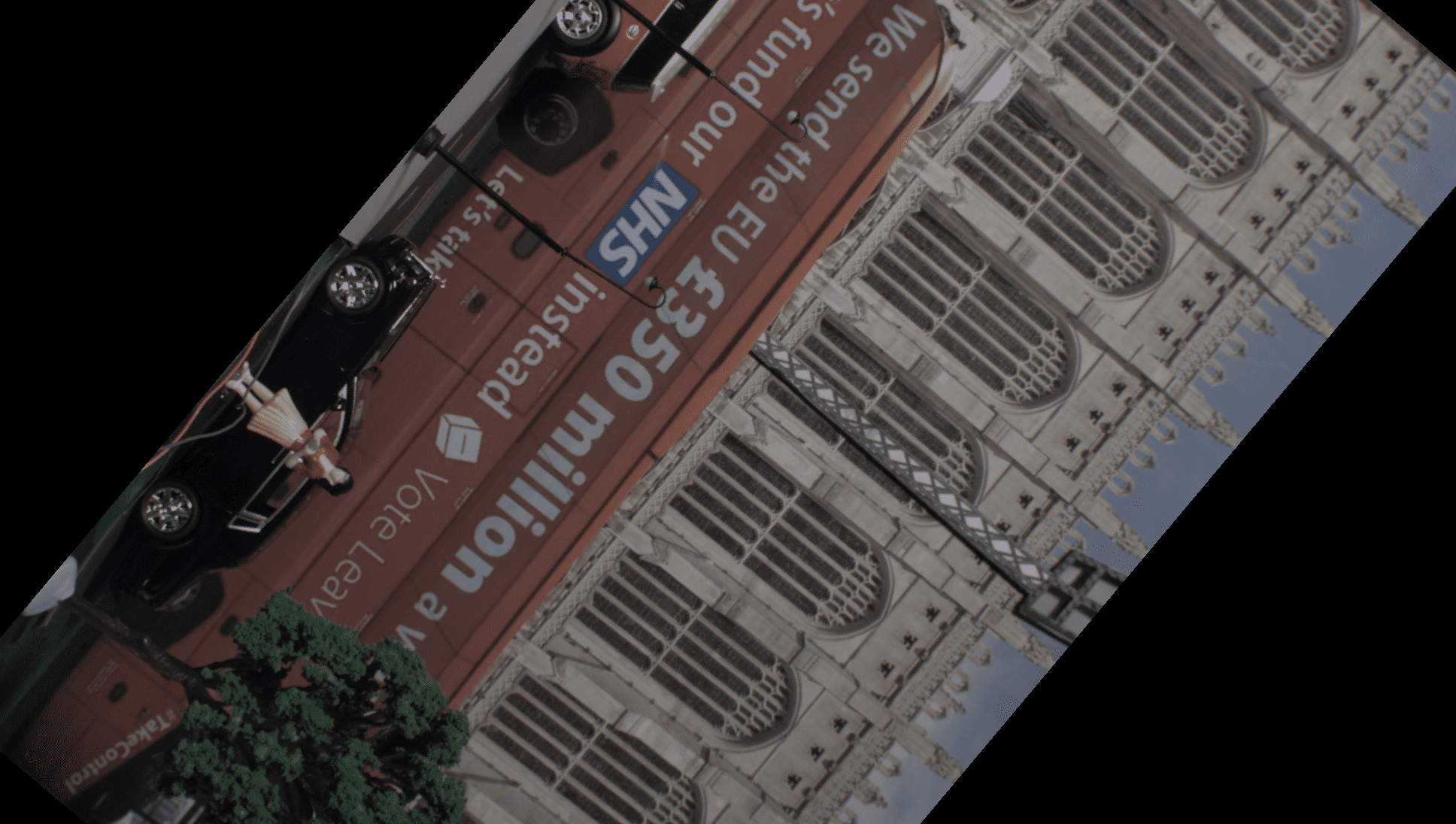}}
\end{subfigure}
\begin{subfigure}{1.5in}
{\includegraphics[width=1.5in,height=1.5in,keepaspectratio]{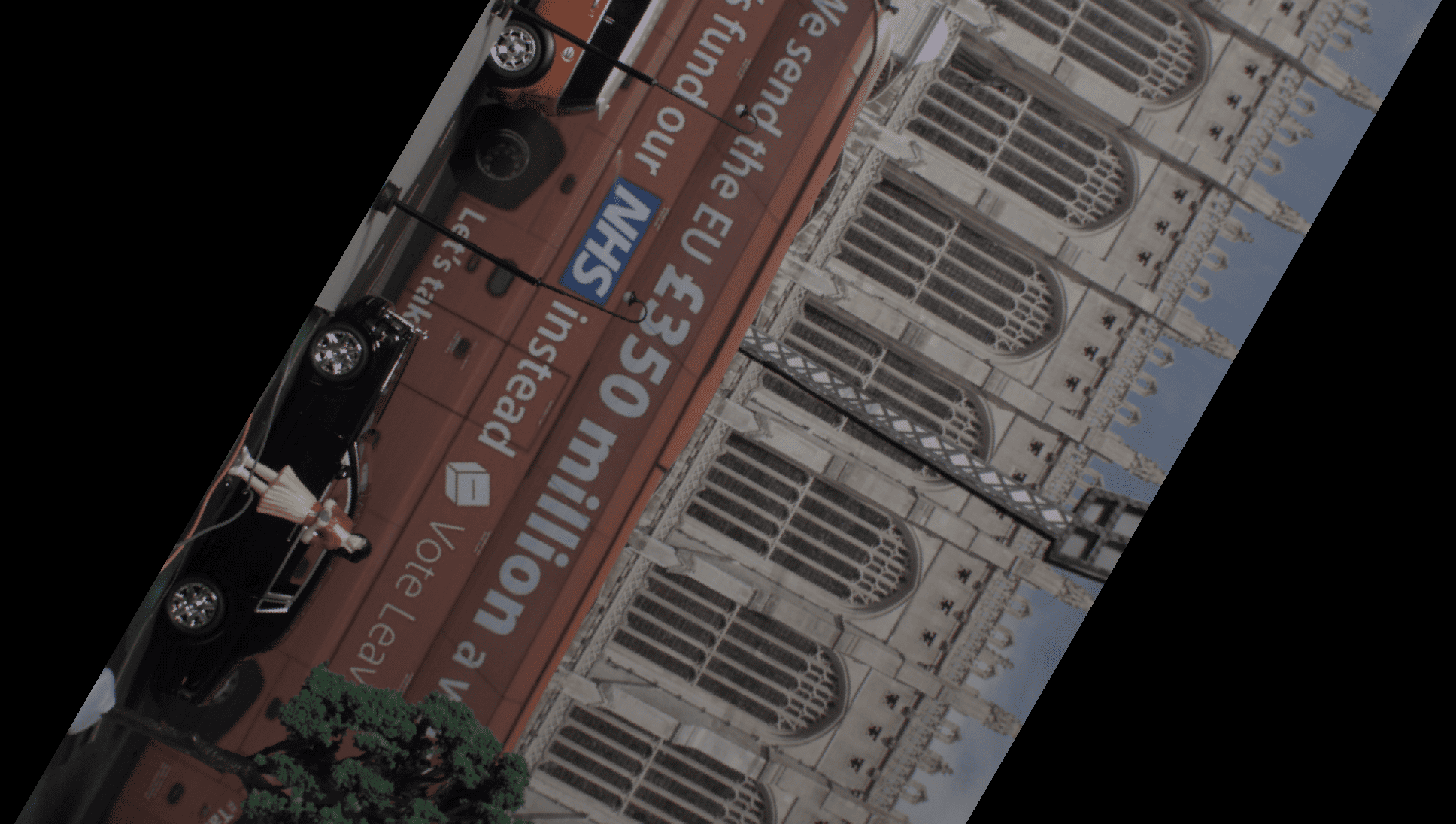}}
\end{subfigure}\\
\begin{subfigure}{1.5in}
{\includegraphics[width=1.5in,height=1.5in,keepaspectratio]{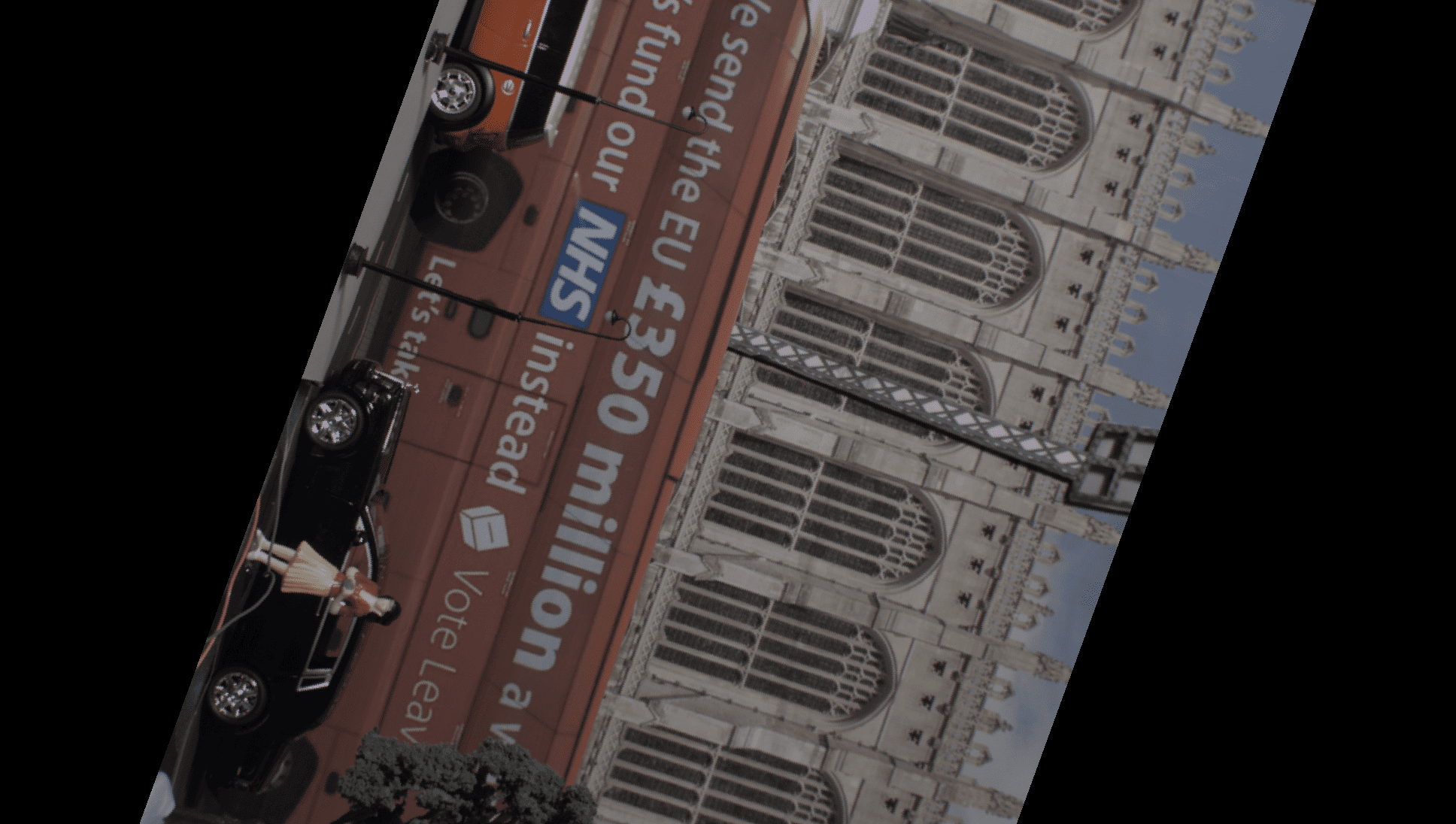}}
\end{subfigure}
\begin{subfigure}{1.5in}
{\includegraphics[width=1.5in,height=1.5in,keepaspectratio]{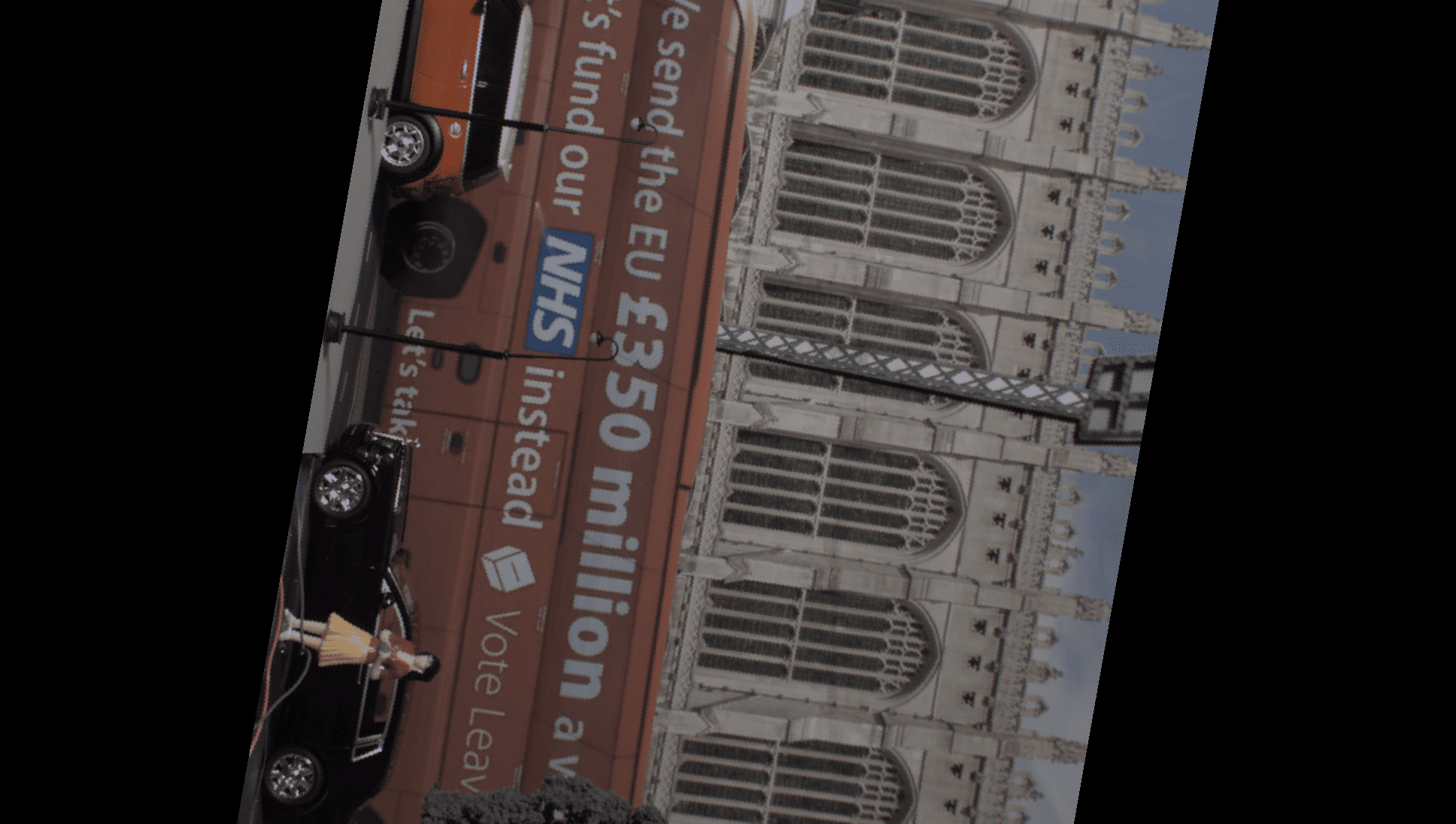}}
\end{subfigure}
\begin{subfigure}{1.5in}
{\includegraphics[width=1.5in,height=1.5in,keepaspectratio]{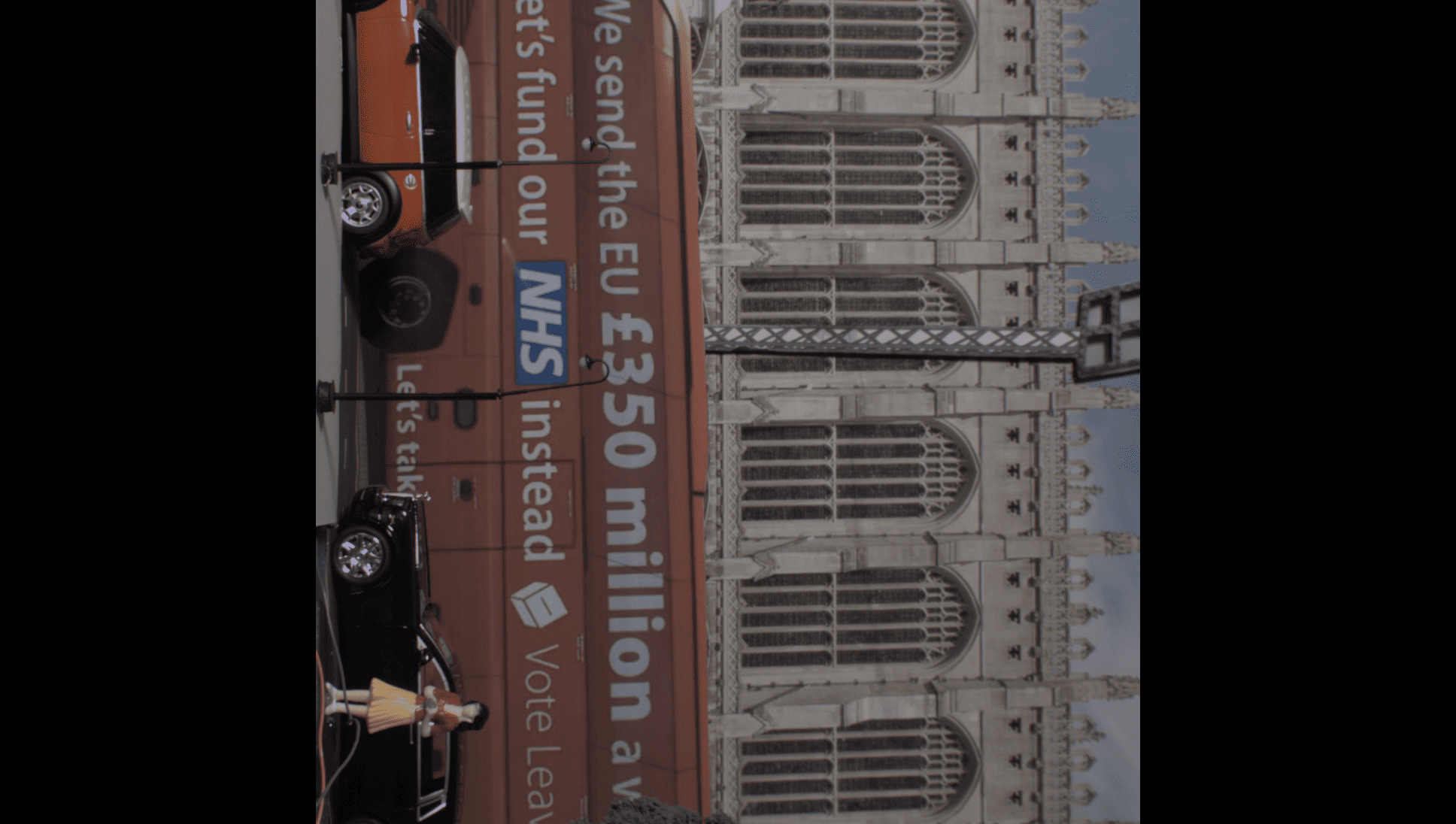}}
\end{subfigure}
\begin{subfigure}{1.5in}
{\includegraphics[width=1.5in,height=1.5in,keepaspectratio]{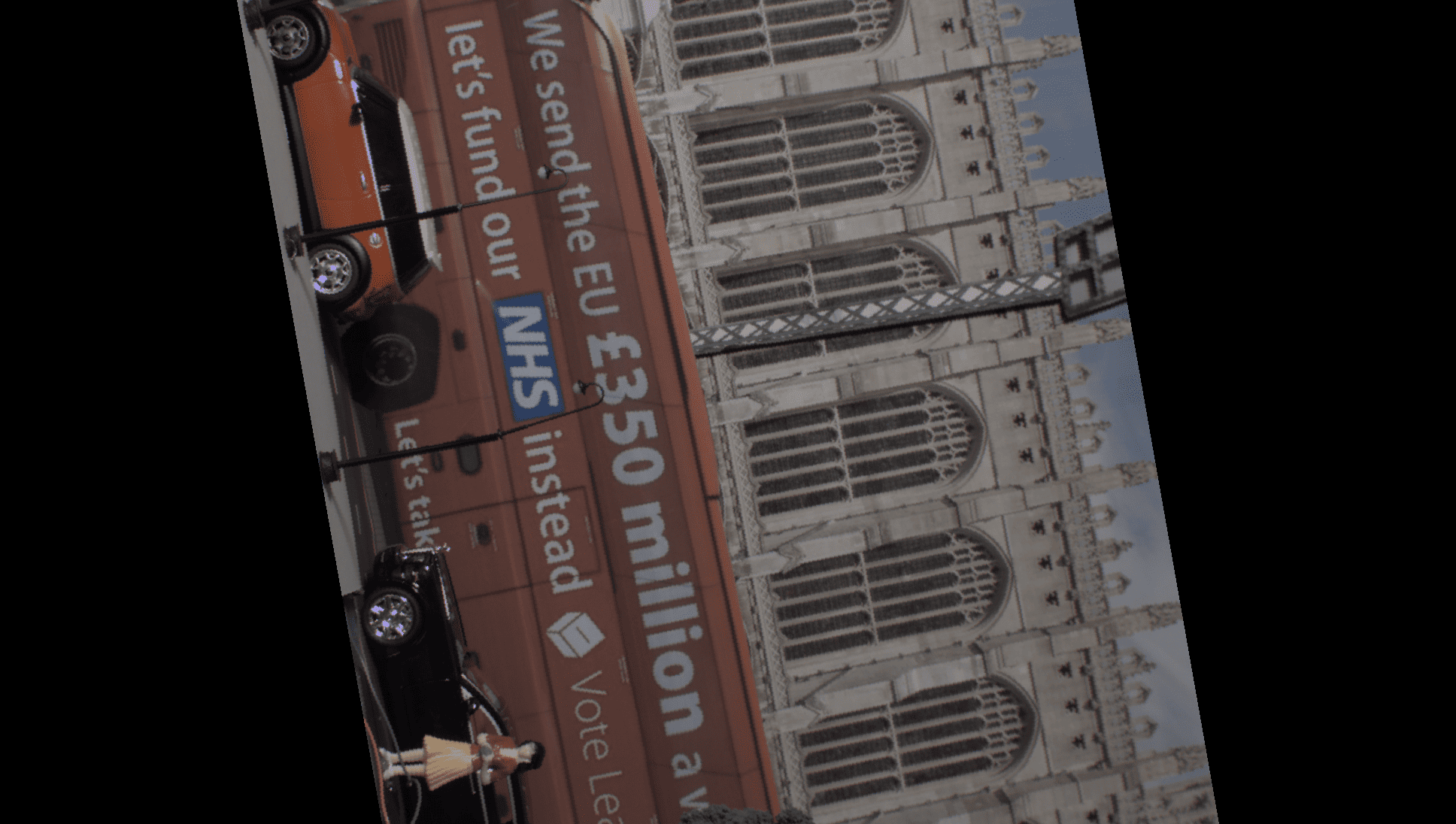}}
\end{subfigure}
\begin{subfigure}{1.5in}
{\includegraphics[width=1.5in,height=1.5in,keepaspectratio]{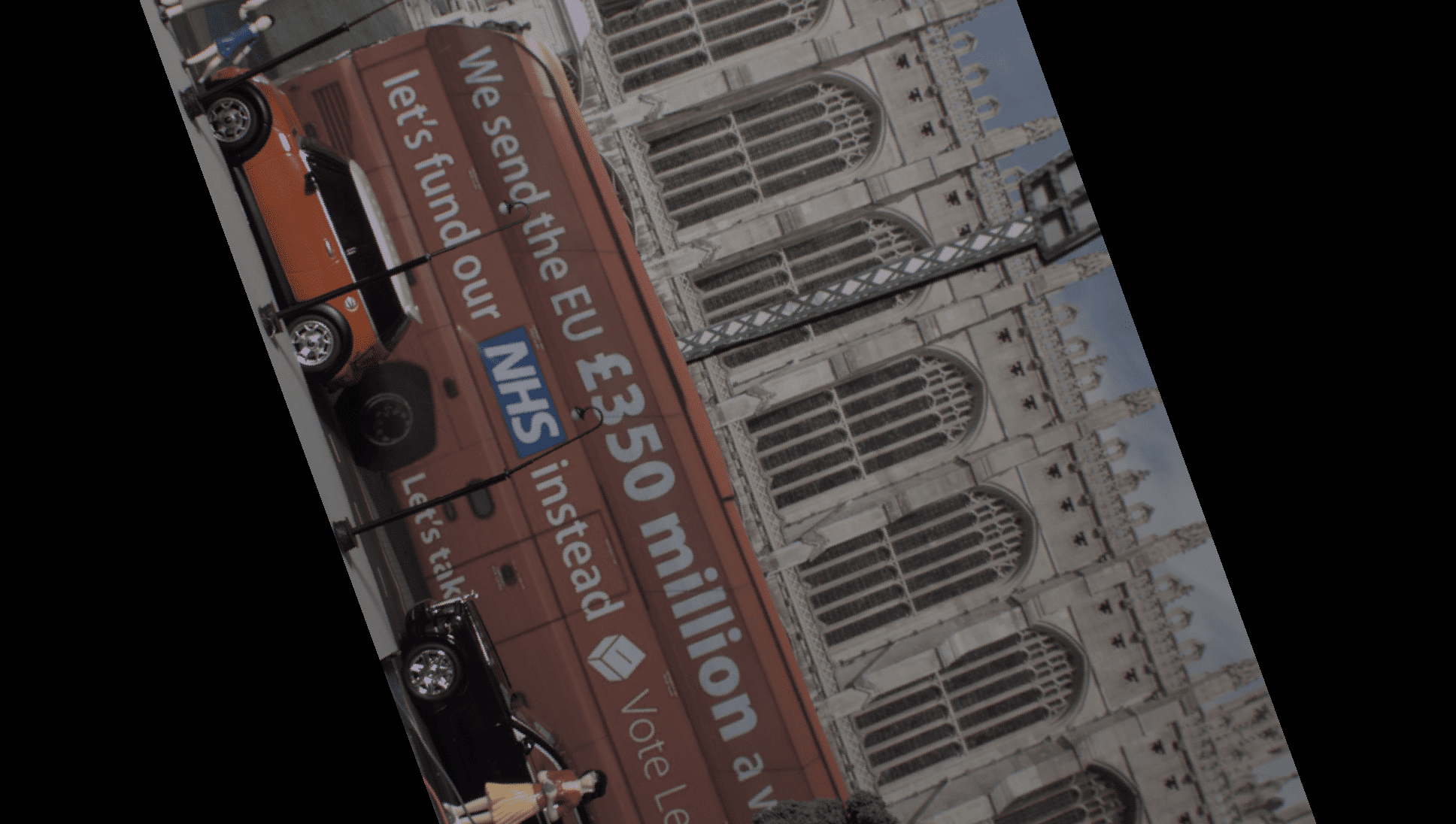}}
\end{subfigure}
\begin{subfigure}{1.5in}
{\includegraphics[width=1.5in,height=1.5in,keepaspectratio]{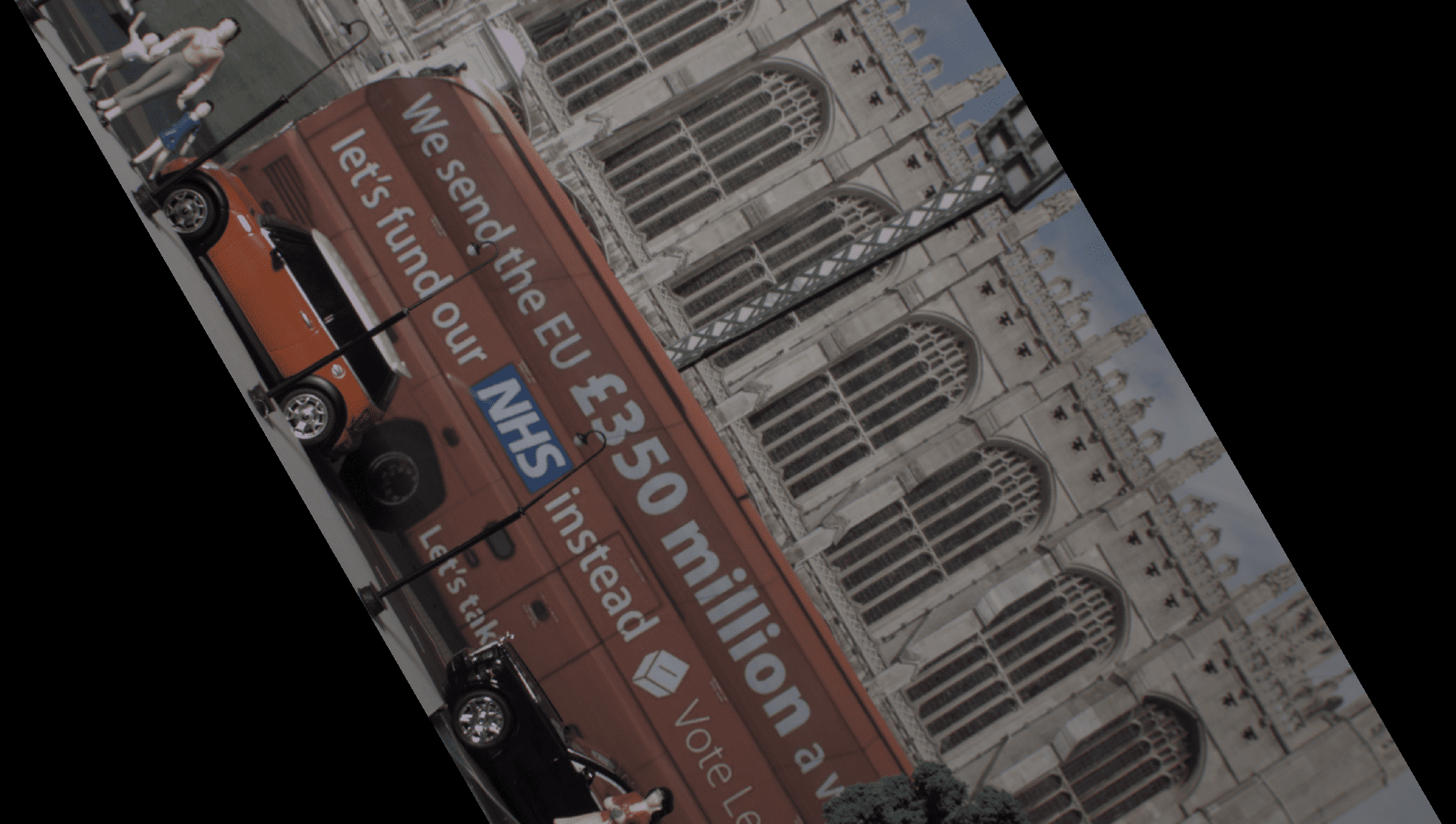}}
\end{subfigure}
\begin{subfigure}{1.5in}
{\includegraphics[width=1.5in,height=1.5in,keepaspectratio]{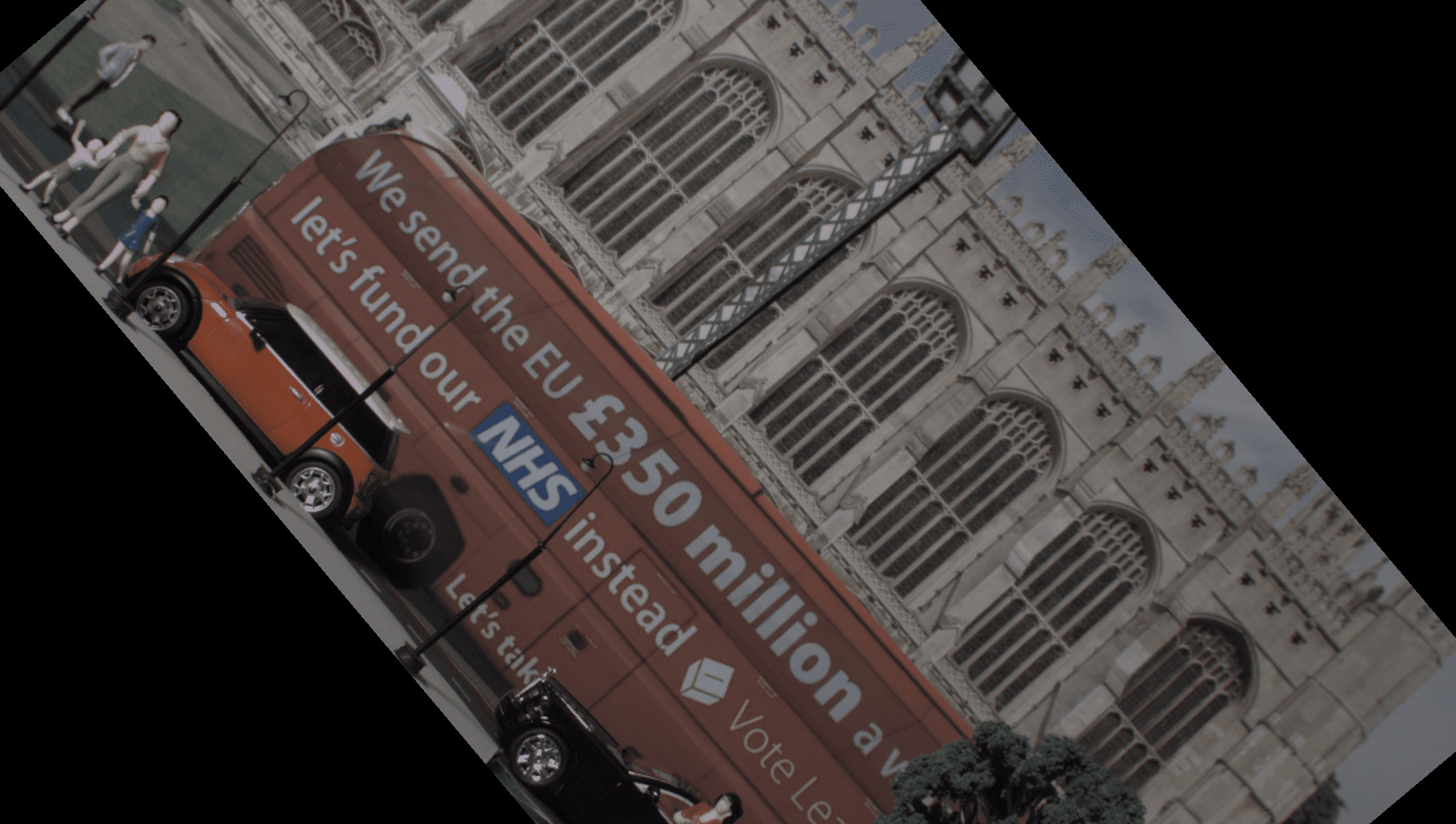}}
\end{subfigure}
\begin{subfigure}{1.5in}
{\includegraphics[width=1.5in,height=1.5in,keepaspectratio]{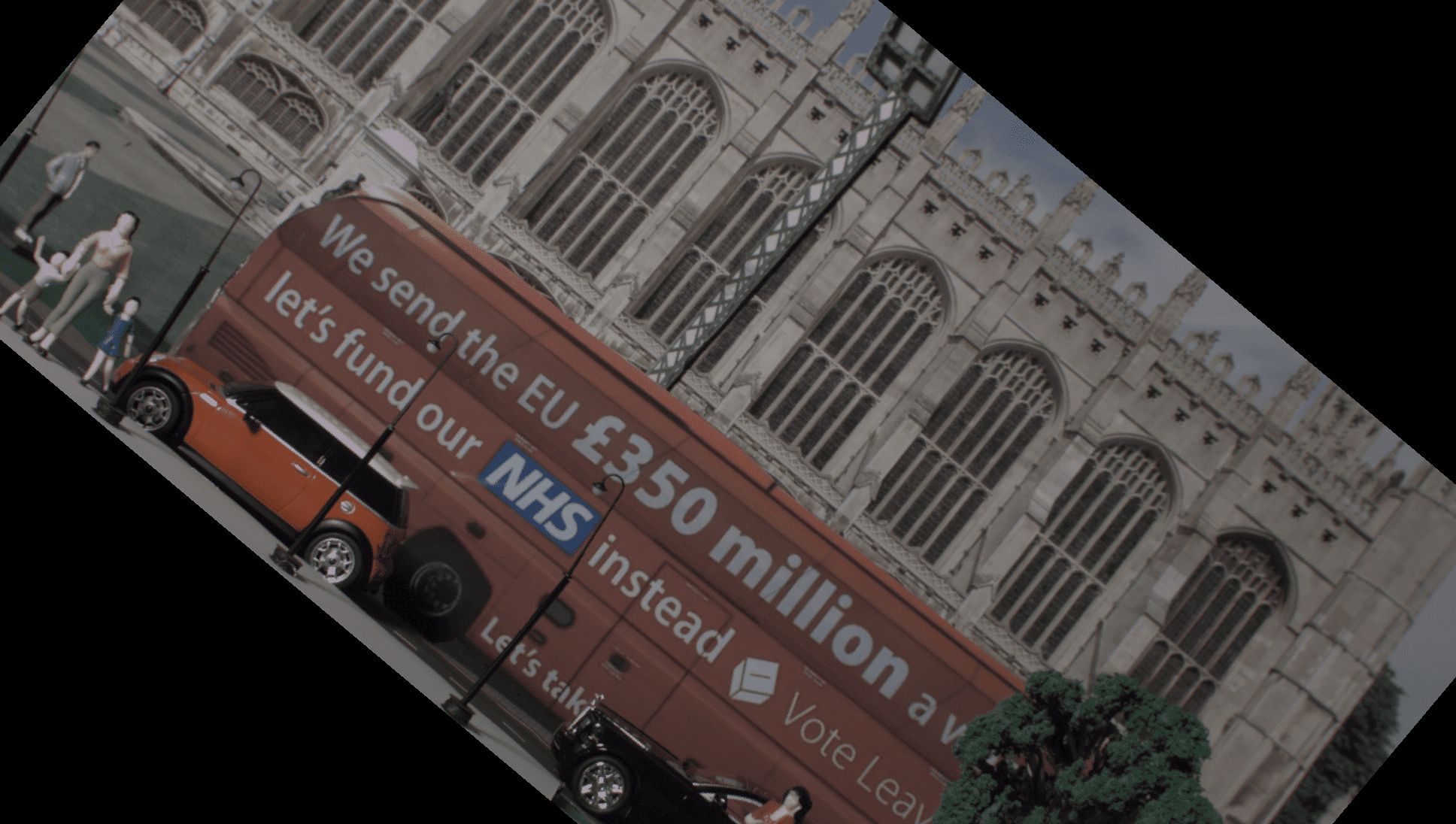}}
\end{subfigure}
\begin{subfigure}{1.5in}
{\includegraphics[width=1.5in,height=1.5in,keepaspectratio]{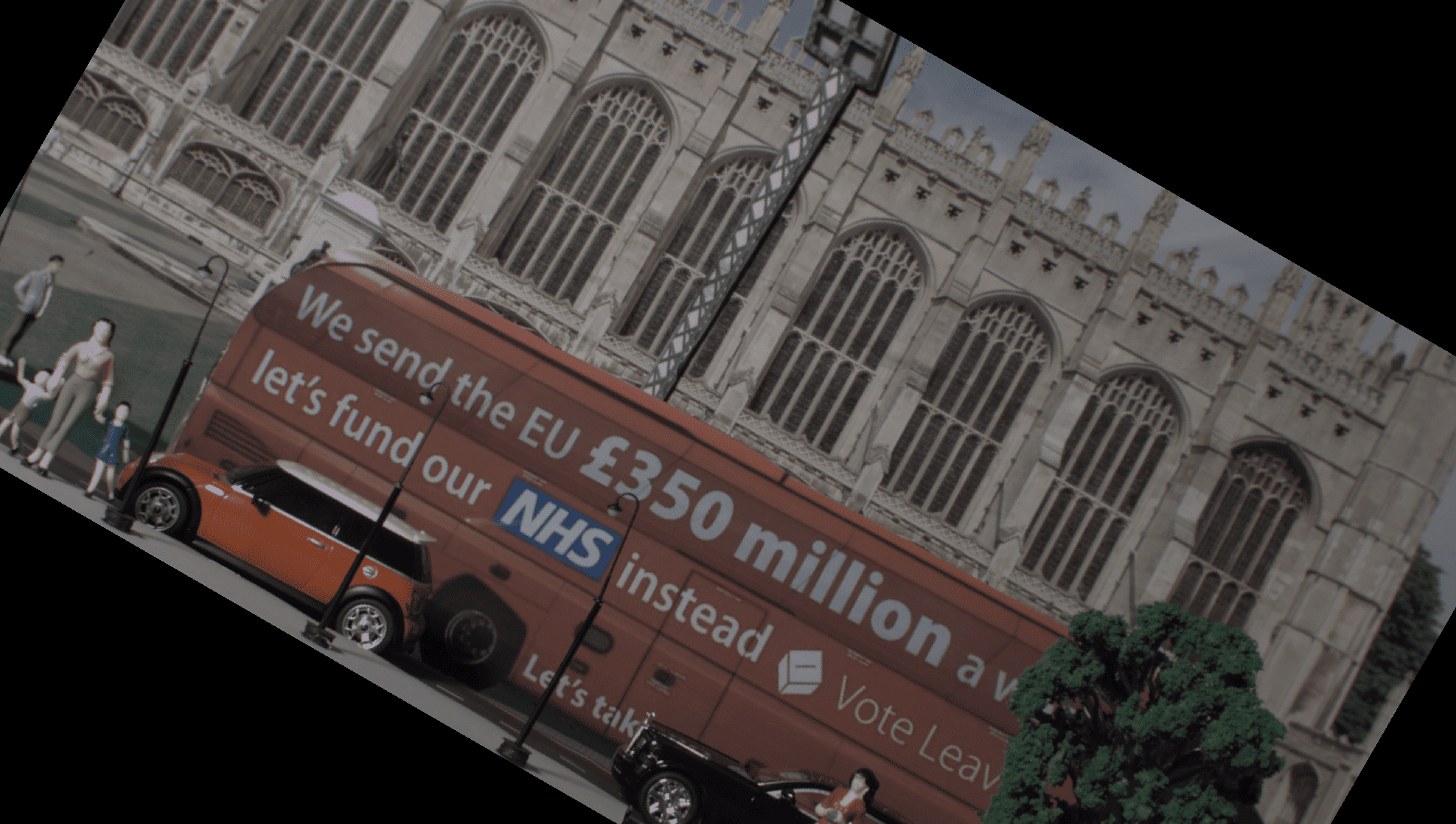}}
\end{subfigure}
\begin{subfigure}{1.5in}
{\includegraphics[width=1.5in,height=1.5in,keepaspectratio]{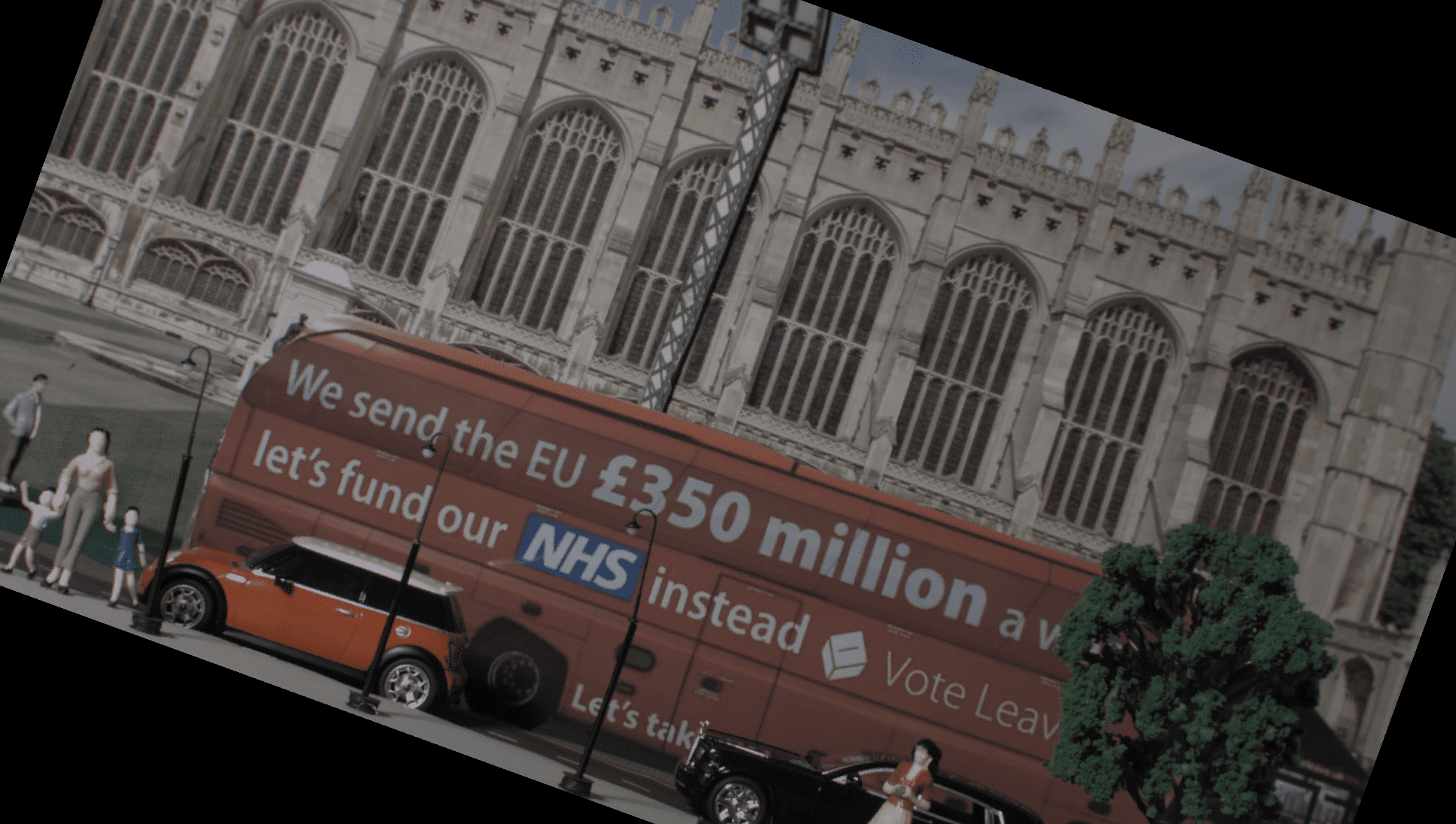}}
\end{subfigure}
\begin{subfigure}{1.5in}
{\includegraphics[width=1.5in,height=1.5in,keepaspectratio]{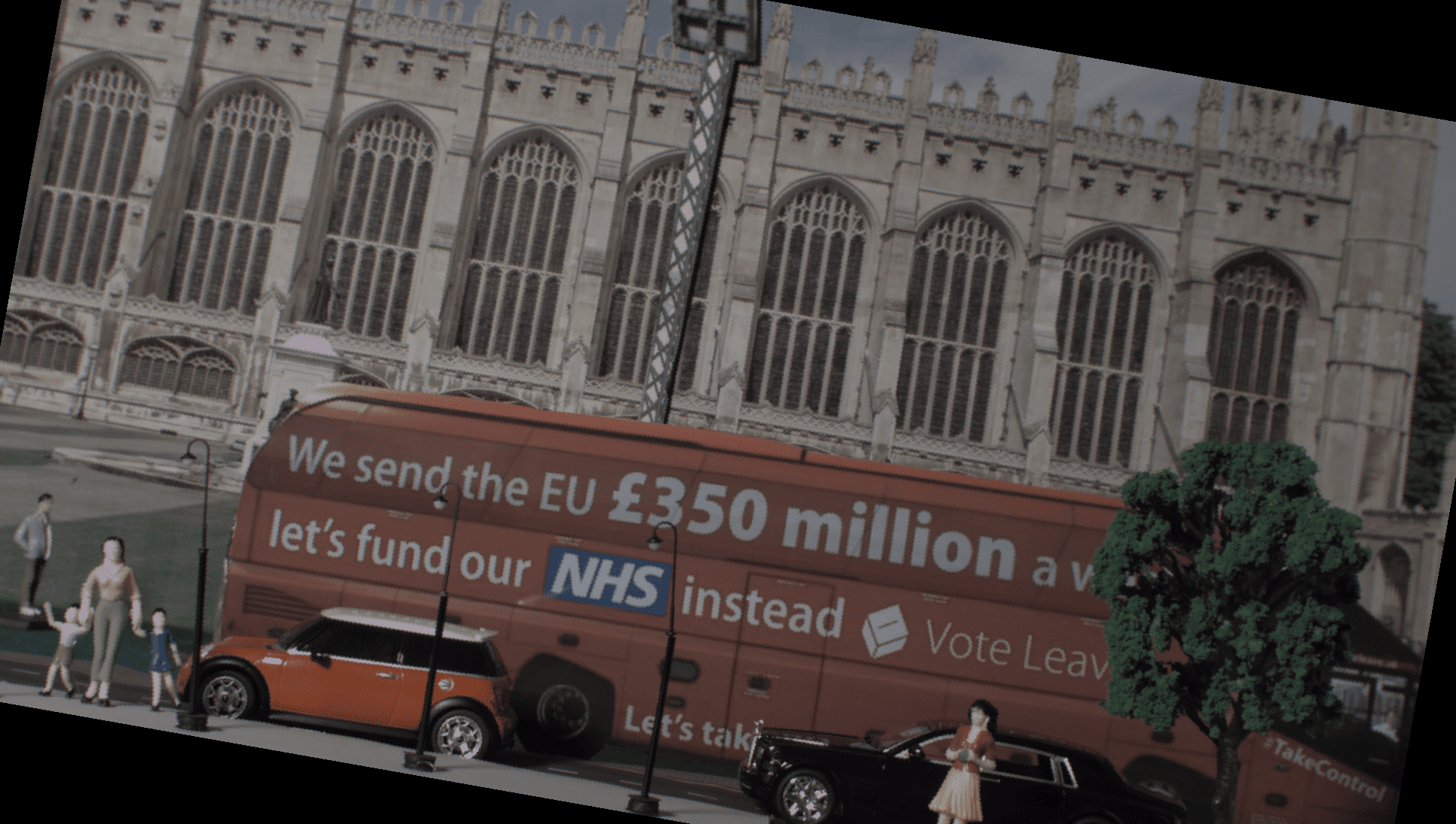}}
\end{subfigure}
\begin{subfigure}{1.5in}
{\includegraphics[width=1.5in,height=1.5in,keepaspectratio]{images/1_36.png}}
\end{subfigure}\\
\raisebox{.7\height}{ \rotatebox[origin=]{00}{Tunnel}}
\caption{Visualization of the image rotations tested for fine-scale image orientation to produce results in Fig. \ref{fig:fineQuant}  }
\label{fig:rot_finescale}
\end{figure*}

In this section, we present both quantitative and qualitative evaluations of our proposed technique for estimating image and video orientation. To ensure a thorough analysis, we employ two distinct datasets: \cite{hanji2021hdr4cv} for fine-grained image orientation estimation (in 10-degree increments) and \cite{xiao2010sun} for broader orientation predictions (at 0, 90, 180, and 270 degrees). This dual-dataset approach facilitates a comprehensive assessment of our method's performance and allows for a detailed comparison with established techniques.\\
{We have conducted a quantitative comparison between the proposed and existing methods, with the results detailed in Table \ref{Summary}.  We present outcomes using two different sizes of test sets, accommodating the various test/training splits employed by these existing methods. For the assessment of coarse image orientation, the SUN database \cite{xiao2010sun}, containing approximately 108,000 images across 397 categories, including a wide range of indoor, outdoor, and natural imaging scenarios is used. Indoor scenes include spaces like classrooms, bedrooms, and libraries, while outdoor scenes cover urban streets, highways, and parks. Additionally, the dataset features natural environments such as mountains, forests, beaches, and deserts. Its comprehensiveness makes it ideal for developing and testing algorithms for scene understanding, offering a challenging variety of scenes with complex spatial arrangements, lighting conditions, and object interactions.} \color{black}{The dataset is split into train/test according to \cite{joshi2017automatic} with 20\% of this dataset allocated for testing purposes. Similarly, for evaluations consistent with \cite{ciocca2015image}, the test set constituted  42,464 images which is approximately 40\% of the total dataset.}\\ 
In Fig. \ref{fig:badPredict}, we presented a subset of images used in the calculation of results presented in Table \ref{Summary}. We present the results generated using two different approaches for depth estimation. The first approach involves detailed depth recovery, requiring the explicit computation of a depth map. For this, we utilize MiDaS, as the dataset does not include explicit depth maps for the images. The second approach focuses on depth inference rather than producing a complete depth map, leveraging the principle of depth-from-defocus, as explained in Section \ref{DOD}. \\
\begin{table}[!h]
\caption{Performance comparison of the proposed method with existing image orientation estimation methods on two different test set splits on Sun397 \cite{xiao2010sun} dataset adopted in various state-of-the-art methods. }
\label{Summary}
\begin{tabular}{|l|l|l|}
\hline
Method                                                                 & \begin{tabular}[c]{@{}l@{}}Sun397 40 \% test\\ Accuracy\end{tabular} & \begin{tabular}[c]{@{}l@{}}Sun397 20\% test\\ Accuracy\end{tabular} \\ \hline
Vailaya et. al. \cite{vailaya2002automatic} & 80.1 \%                                                              & --                                                                  \\ \hline
Tolstaya \cite{tolstaya2007content}                   & 85.6\%                                                               & --                                                                  \\ \hline
Ciocca et. al \cite{ciocca2010image}                  & 89.3\%                                                               & --                                                                  \\ \hline
Appia and Narasimah \cite{appia2011low}               & 54.3\%                                                               & --                                                                  \\ \hline
Ciocca et. al \cite{ciocca2015image}                  & 92.4\%                                                               & --                                                                  \\ \hline
Joshi et. al \cite{joshi2017automatic}               & --                                                                   & 98.5\%                                                              \\ \hline
Subhadip et. al \cite{maji2020deep}                   & 96.2\%                                                               & 96.4\%                                                              \\ \hline
Proposed                                                               & 98.5\%                                                               & 99.0\%                                                              \\ \hline
\end{tabular}
\end{table}
It should be noted that in most scenarios where the algorithm fails, irrespective of the depth estimation methods, contains images of the planer scene with little to no depth variation, which goes against the principle of linear perspective adopted in the proposed algorithm. For planar scenes, such as the Cathedral and Canyon images, Fig. \ref{fig:badPredict} (top row), the depth-from-defocus approach outperforms Monoclour depth estimation \cite{Ranftl2020}. This success can be attributed to its reliance on localized defocus cues, which are well-suited for scenes with clear focal gradients. In contrast, MiDaS struggles in these scenarios due to its dependency on global priors, which may not generalize well to uniform or texture-less regions.\\
However, in Fig. \ref{fig:badPredict} (2nd row), the image featuring Nuns, depth-map based method captures the global structural and semantic depth relationships effectively, enabling accurate orientation prediction in scenes with intricate depth variations. Depth-from-defocus fails in this specific case, due to the absence of significant blur gradients, which are critical for its depth inference. \\
The accuracy of the proposed orientation prediction method is not heavily reliant on the specific method used for depth extraction or the fine-level details within the depth map. Neither high-resolution depth maps nor precise depth measurements are primary concerns for the success of the approach.  
The method leverages coarse depth estimates to infer relative depth differences in the scene, which are sufficient to achieve competitive orientation prediction results.\\
If the input image comes with a precomputed depth map, the method can directly utilize it for orientation estimation. However, in cases where the depth map is unavailable, any existing depth inference technique from a single image can be employed, as long as it provides a computationally efficient and approximate depth estimation. Techniques such as depth-from-shading, depth-from-defocus, and depth-from-perspective can generate depth estimates that are sufficient for the proposed algorithm's requirements. The findings suggest that the method is robust and adaptable to diverse depth extraction techniques, enabling its application in computationally constrained environments or scenarios where precise depth information is challenging to obtain. \\
In Table \ref{Summary}, The proposed method surpasses state-of-the-art deep learning models, such as Subhajit et al. \cite{maji2020deep} and Joshi et al. \cite{joshi2017automatic}, particularly due to its ability to generalize across diverse scenes in the Sun397 dataset. Deep learning models often rely on dataset-specific features and priors, which can limit their adaptability to unseen scenarios in image orientation estimation. This limitation arises because these models struggle to infer global structural relationships and relative depth cues effectively, especially in scenes with unconventional perspectives or limited training data coverage. The proposed method, by integrating depth-based reasoning, is inherently robust to such variations, ensuring reliable orientation estimation across diverse application scenarios.\\
The comparison between the 20 \% and 40\% test splits of the Sun397 dataset reveals a notable drop in performance for state-of-the-art deep learning-based methods, highlighting their dependence on larger training sets and their limited ability to generalize. For example, Subhajit et al. \cite{maji2020deep} demonstrate a drop in accuracy from 96.4\% on the 20\% split to 92.4\% on the 40\% split, a reduction of 4\%. Similarly, Joshi et al. \cite{joshi2017automatic} exhibits a drop of approximately 3.6\%. These declines reflect the data-hungry nature of deep learning models, which rely heavily on extensive training data to perform well across diverse scenarios when encounter a decrease in 21,751 images of various categories from the training dataset suffer significantly in performance. In contrast, the proposed method exhibits remarkable stability, with only a 0.5\% reduction in accuracy (from 99.0\% to 98.5\%) despite the significant increase in testing images.\\ 
\color{black}
In Fig.~\ref{fig:fineQuant}, we compare the performance of the proposed method against state-of-the-art approach for fine-scale image rotation angle estimation. The evaluation considers the accuracy of angle predictions across a complete range of 360\textdegree rotations, with results computed at increments of 10\textdegree. The state-of-the-art method by \cite{maji2020deep} reports angle predictions with a tolerance of up to 1\textdegree, making it necessary to define a delta value to fairly evaluate and represent tolerance-based accuracy. The results are presented for delta values of 5\textdegree and 10\textdegree, alongside a rounded evaluation for baseline comparisons.\\
For this test, a carefully controlled dataset was created by selecting 5 images each from 4 diverse categories, shown in Fig. \ref{fig:Subset}, resulting in 20 unique images. Each image was rotated by 10\textdegree increments, generating a total of 720 test images (20 images × 36 rotations).  Fig. \ref{fig:rot_finescale} visualize the rotation applied to one of the 5 images from one of the four categories to demonstrate a robust evaluation across varying image types and orientations. 
The results in Fig.~\ref{fig:fineQuant} demonstrate the superior performance of the proposed method across almost all tested angles, except for 340\textdegree and 350\textdegree, consistently achieving near-perfect accuracy. Unlike the state-of-the-art method (OAD-Round, OAD-$\Delta$5, and OAD-$\Delta$10), the proposed method exhibits minimal variation in accuracy across different rotation angles, highlighting its robustness and reliability in fine-scale orientation estimation. The delta-based evaluation of the state-of-the-art method shows significant sensitivity to tolerance thresholds, with a marked drop in accuracy as delta decreases. This performance gap underscores the challenges faced by the method in precisely estimating fine-scale rotations, particularly for challenging angles where small prediction errors can lead to substantial accuracy reductions.  
\\
\\

\begin{figure*}[!h]
\centering
\begin{subfigure}{1.7in}
{\includegraphics[width=1.7in,height=1.7in,keepaspectratio]{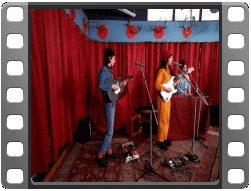}}
\end{subfigure}
\begin{subfigure}{1.7in}
{\includegraphics[width=1.7in,height=1.7in,keepaspectratio]{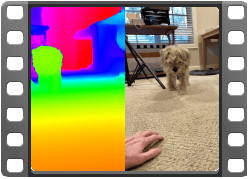}}
\end{subfigure}
\begin{subfigure}{1.7in}
{\includegraphics[width=1.7in,height=1.7in,keepaspectratio]{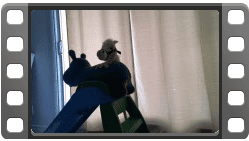}}
\end{subfigure}
\begin{subfigure}{1.7in}
{\includegraphics[width=1.7in,height=1.7in,keepaspectratio]{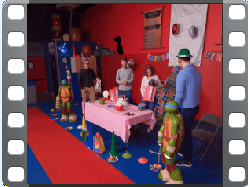}}
\end{subfigure}
\caption{Thumbnails of the selected subset of video sequences tested for the evaluation of the proposed orientation estimation technique.}
\label{fig:thumbs}
\end{figure*}

In addition to testing the proposed method on large-scale image datasets, we conducted a comprehensive evaluation on a diverse collection of 100 videos with varying lengths. These videos were captured using multiple camera sensors, including Kinect, multi-view machine vision cameras, and iPhones, each recorded with different orientations. For all video sources, the acquisition systems provided depth maps along with the corresponding RGB frames, including those captured with iPhones. Fig. \ref{fig:thumbs} present thumbnails of the subset of the tested videos.\\
\\
The proposed method demonstrated exceptional robustness, achieving 100\% accuracy in true video orientation estimation across all frames of the tested videos. These results highlight the effectiveness of the algorithm in maintaining consistent performance across diverse video sources and acquisition scenarios, further validating its applicability to real-world dynamic settings.
\color{black}
\section{Conclusion}
In this paper, we introduced a new method for estimating image and video orientation using depth distribution in natural images. By combining DGC and HSA, our technique accurately identifies the orientation of images at a fine scale. Our methodology offers a novel alternative to the prevailing deep learning methods. This approach stands out for its broader applicability and ease of generalization across various data types, obviating the need for extensive fine-tuning and simplifying the otherwise complex model training often seen in current state-of-the-art techniques, while providing a unique perspective that was previously unexplored. Our evaluations, conducted on two datasets, show that our method performs well against existing approaches in accurately predicting image orientation both broadly and in detail. This work advances computer vision's automation goals by developing an automated method for estimating image orientation, demonstrating promising outcomes and exploration potential.
\section{Future work}
{As discussed in Section III, the horizon line shifts with changes in camera angle (e.g., low, high, eye level). In the future, we aim to enhance the selection of image quadrant boundaries by dynamically adjusting them based on the movement of the horizon line. This adjustment would enable better alignment and more accurate estimation of depth magnitude. We believe that a dynamic selection of quadrant boundaries informed by the horizon line will further improve the accuracy of orientation prediction.}
\color{black}
Furthermore, future directions for this research will delve into computationally efficient alternatives for depth estimation in image orientation tasks. Building on the concept of defocus to determine depth displacement, we aim to refine sharpness measures across image quadrants as a proxy for depth, minimizing the need for intensive computation. This approach will explore the use of image processing algorithms to assess defocus levels, creating a depth profile to inform orientation estimation. Further, we will investigate heuristic methods that leverage defocus characteristics, simplifying the orientation process while maintaining accuracy. By advancing these alternatives, we seek to enhance the model's efficiency, making it more practical for real-time applications where computational resources are at a premium.\\
Additionally, we plan to selectively estimate the depth by concentrating on critical image regions that contribute most significantly to orientation. Gradient-based methods will also be harnessed to infer depth displacement from the rate of intensity changes, offering a lightweight alternative to traditional depth mapping. These methods aim to achieve a balance between computational efficiency and the accuracy required for reliable image orientation determination. \\
\bibliographystyle{IEEEbib}
\bibliography{strings.bib}
\clearpage
\begin{IEEEbiography}[{\includegraphics[width=1in,height=1.25in,clip,keepaspectratio]{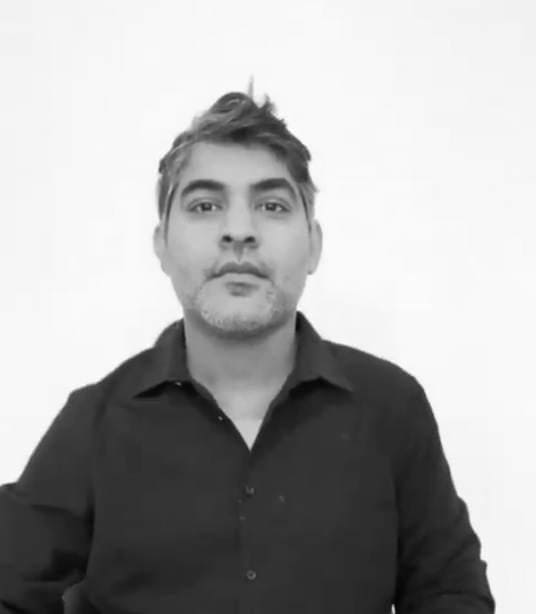}}]{Muhamad Zeshan Alam} received  his  B.S.  degree  in Computer  Engineering  from  COMSATS  University,  Pakistan,   M.S.  degree  in  Electrical  and Electronics  Engineering  from  the  University  of Bradford, UK, and Ph.D. In Electrical Engineering  and  Cyber-Systems  from  Istanbul  Medipol University, Turkey. He worked at the University of Cambridge as a post-doctoral fellow where his work focused on computer vision and machine learning models. He currently joined Brandon University, Canada, as an assistant professor while also working as a Computer Vision Consultant at Vimmerse INC. His research interests include Autonomous vehicle perception, immersive videos, computational imaging, computer vision, and machine learning modeling.
\end{IEEEbiography}
 \begin{IEEEbiography}[{\includegraphics[width=1in,height=1.25in,clip,keepaspectratio]{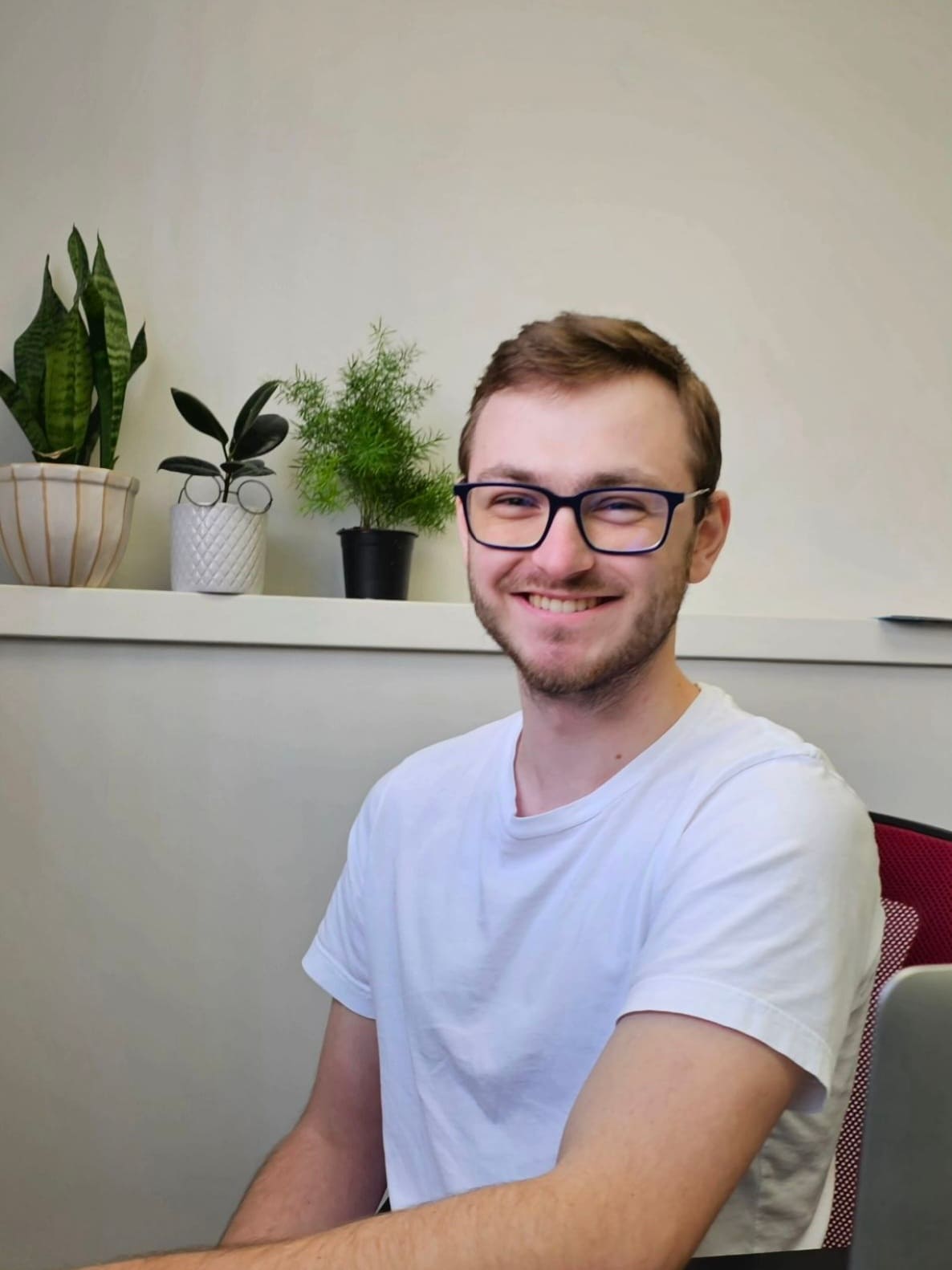}}]{Larry Stetsiuk}  is a 4th year student as Brandon University, majoring in Computer Science with a minor in mathematics. 
\end{IEEEbiography}
\begin{IEEEbiography}[{\includegraphics[width=1in,height=1.25in,clip,keepaspectratio]{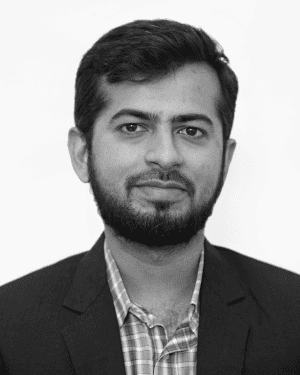}}]{M. Umair Mukati} earned his M.S. degree in Electronics Engineering from Istanbul Medipol University, Turkey, in 2017, followed by his Ph.D. degree in Electronics and Photonics Engineering in 2022. During his master's studies, he focused on enhancing the quality of light fields captured using affordable cameras. He is an alumnus of the EU MSCA ITN RealVision project, where he served as an Early Stage Researcher. Currently, he is serving as a post-doctoral researcher at Technical University of Denmark and carrying out research for the advancement of visual experience in video-conferencing. 
His research interests includes exploring unconventional methods for image compression, enhancement, and view rendering.
\end{IEEEbiography}
\begin{IEEEbiography}[{\includegraphics[width=1in,height=1.25in,clip,keepaspectratio]{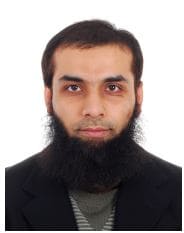}}]{Zeeshan Kaleem [Senior Member, IEEE] }  is serving as an Associate Professor at COMSATS University Islamabad, Wah Campus. He consecutively
received the National Research Productivity Award
(RPA) awards from the Pakistan Council of Science
and Technology (PSCT) in 2017 and 2018. He
won the Higher Education Commission (HEC) Best
Innovator Award for the year 2017, and there was
a single award all over Pakistan. He is the recipient
of the 2021 Top Reviewer Recognition Award for
IEEE TRANSACTIONS on VEHICULAR TECHNOLOGY and published over 100 technical papers and patents.
\end{IEEEbiography}

\end{document}